\documentclass{article}

\usepackage{microtype}
\usepackage{graphicx}
\usepackage{subcaption}
\usepackage{booktabs} % for professional tables
\usepackage{xcolor}
\usepackage{enumitem}

\definecolor{asrgreen}{RGB}{0,140,0}
\definecolor{asrred}{RGB}{180,0,0}
\definecolor{taxred}{RGB}{180,0,0}
\definecolor{asrgreen}{RGB}{0,140,0}
\definecolor{asrred}{RGB}{180,0,0}

\newcommand{\asrup}[1]{{\color{asrred}#1}}     % ASR increase
\newcommand{\taxdelta}[1]{{\color{taxred}#1}}  % Reasoning drop

\definecolor{taxred}{RGB}{180,0,0}
\usepackage{hyperref}

\usepackage[preprint]{icml2026}

\usepackage{amsmath}
\usepackage{amssymb}
\usepackage{mathtools}
\usepackage{amsthm}
\usepackage{booktabs}
\usepackage{multirow}
\usepackage{makecell}
\usepackage[capitalize,noabbrev]{cleveref}

\theoremstyle{plain}

\theoremstyle{definition}

\theoremstyle{remark}

\usepackage[textsize=tiny]{todonotes}

\icmltitlerunning{Amplification Effects in Test-Time Reinforcement Learning}

\begin{document}

\twocolumn[
  \icmltitle{Amplification Effects in Test-Time Reinforcement Learning:\\ Safety and Reasoning Vulnerabilities}



  \icmlsetsymbol{equal}{*}

  \begin{icmlauthorlist}
    \icmlauthor{Vanshaj Khattar}{yyy}
    \icmlauthor{Md Rafi ur Rashid}{xxx}
    \icmlauthor{Moumita Choudhury}{zzz}
    \icmlauthor{Jing Liu}{comp}\\
    \icmlauthor{Toshiaki Koike-Akino}{comp}
    \icmlauthor{Ming Jin}{yyy}
    \icmlauthor{Ye Wang}{comp}
  \end{icmlauthorlist}

  \icmlaffiliation{yyy}{Virginia Tech}
  \icmlaffiliation{xxx}{Penn State University}
  \icmlaffiliation{zzz}{University of Massachusetts Amherst}
  \icmlaffiliation{comp}{Mitsubishi Electric Research Laboratories}

  \icmlcorrespondingauthor{Vanshaj Khattar}{vanshajk@vt.edu}
  \icmlcorrespondingauthor{Ye Wang}{yewang@merl.com}

  \vskip 0.3in
]



\printAffiliationsAndNotice{}  % no special notice (required even if empty)

\begin{abstract}
Test-time training (TTT) has recently emerged as a promising method to improve the reasoning abilities of large language models (LLMs), in which the model directly learns from test data without access to labels. However, this reliance on test data also makes TTT methods vulnerable to harmful prompt injections. In this paper, we investigate safety vulnerabilities of TTT methods, where we study a representative self-consistency-based test-time learning method: test-time reinforcement learning (TTRL) \cite{zuo2025ttrl}, a recent TTT method that improves LLM reasoning by rewarding self-consistency using majority vote as a reward signal. We show that harmful prompt injection during TTRL amplifies the model’s existing behaviors, i.e., \textbf{safety amplification} when the base model is relatively safe, and \textbf{harmfulness amplification} when it is vulnerable to the injected data. In both cases, there is a decline in reasoning ability, which we refer to as the \textbf{reasoning tax}. We also show that TTT methods such as TTRL can be exploited adversarially using specially designed ``HarmInject'' prompts to force the model to answer jailbreak and reasoning queries together, resulting in stronger harmfulness amplification. Overall, our results highlight that TTT methods that enhance LLM reasoning by promoting self-consistency can lead to amplification behaviors and reasoning degradation, highlighting the need for safer TTT methods.
\end{abstract}

% \vspace{-0.1cm}
\section{Introduction}

\begin{figure}[t]
    \centering
    \includegraphics[width=\columnwidth]{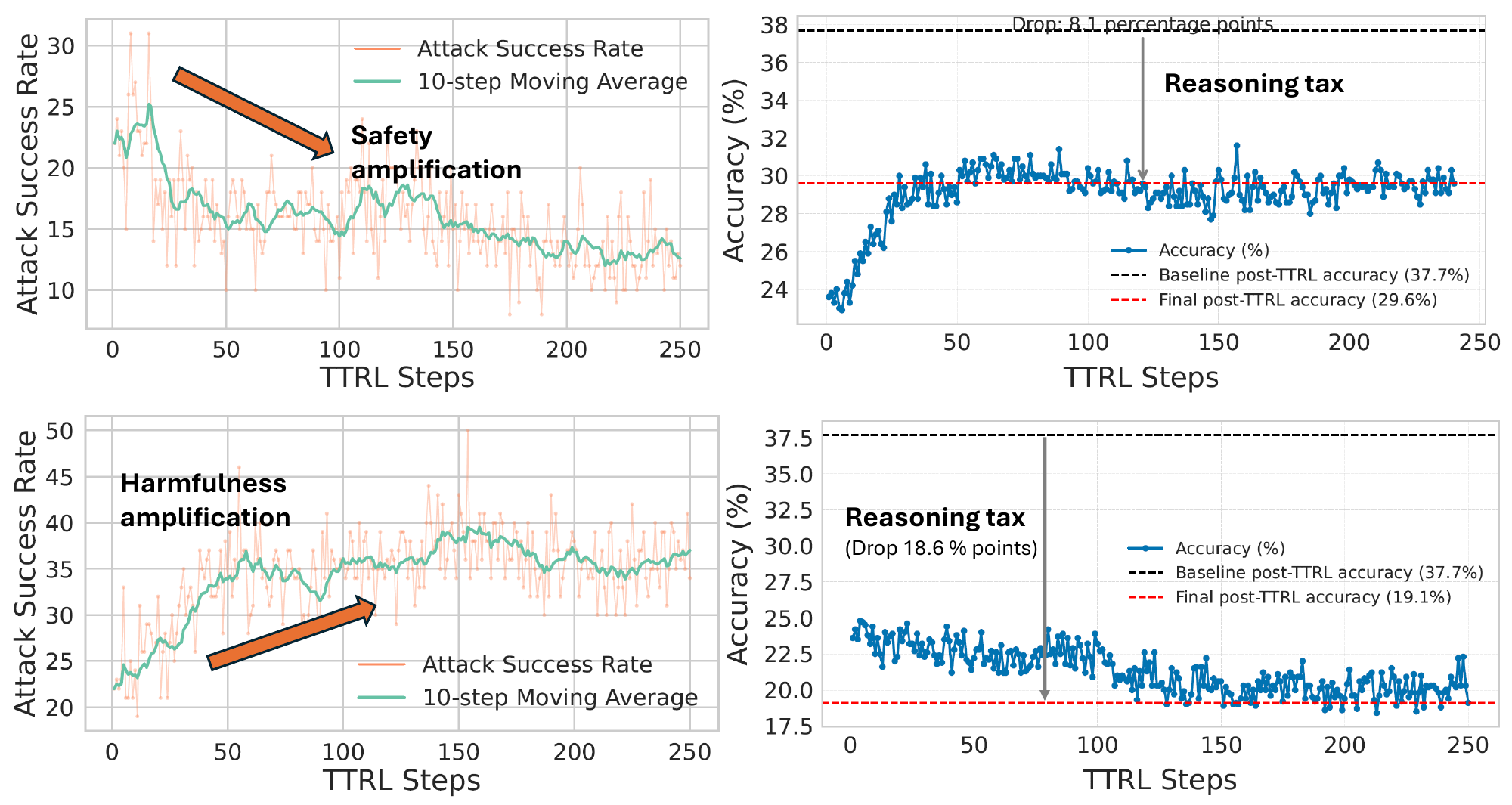}
    \caption{\textbf{Safety and harmfulness amplification} during Test-Time Reinforcement Learning (TTRL). Top left: attack success rate (ASR \%) of Jailbreak-V28k prompts on Qwen1.5B-Instruct when Jailbreak-V28k prompts are injected into AMC test-time data. Top right: the resulting \textbf{reasoning tax}, i.e., loss in AMC accuracy. Bottom left: ASR for Qwen-1.5B-Instruct on Jailbreak-V28k when TTRL is done on HarmInject prompts (see Section \ref{subsec:RQ4}). Bottom right: reasoning tax for the Qwen-1.5B-Instruct model.}
    \label{fig:Figure1}
    
\end{figure}

The reasoning abilities of Large language models (LLMs) have continued to improve through both supervised fine-tuning (SFT) and reinforcement learning (RL) \cite{guo2025deepseek,zhang2025survey} methods. Despite these gains, current LLMs still struggle with out-of-distribution reasoning tasks \cite{phan2025humanity}. To address this, a growing line of work has explored test-time training (TTT) \cite{zuo2025ttrl,prabhudesai2025maximizing,zhao2025learning,jang2025self}, where, unlike traditional approaches that rely on labeled data or verifiable rewards, these TTT methods adapt models directly on the test inputs, often by promoting self-consistency across multiple generations or by constructing pseudo-labels from the queries themselves. These methods have already shown improvements in arithmetic reasoning \cite{li2024numinamath,hendrycks2measuring}, commonsense QA \cite{rein2024gpqa}, and multi-step problem solving without requiring additional human annotation or external supervision.

\noindent\textbf{Central question.}
Self-consistency-based test-time training updates an LLM using pseudo-supervision derived from the model’s own generations (e.g., majority vote). In deployment, the prompt stream is heterogeneous: the model should \emph{refuse} harmful requests but \emph{complete} benign reasoning problems. This raises a basic question: \emph{do self-consistency test-time updates
preserve this conditional behavior under a mixed prompt stream, or do they couple safety behavior and reasoning performance?}

We study this question in the setting of test-time reinforcement learning with majority-vote pseudo-rewards (TTRL~\cite{zuo2025ttrl}) under prompt injection. Across five instruction-tuned models, we find that TTRL updates reinforce whatever behavior dominates on injected prompts (refusal or compliance). Crucially, these updates also change behavior on benign reasoning prompts, producing a consistent drop in AMC reasoning performance (a reasoning tax). We further show that an attacker can strengthen this coupling by composing a jailbreak request with a reasoning question in a single prompt (HarmInject), causing larger harmfulness amplification and larger reasoning degradation.

Some key findings are presented in Figure \ref{fig:Figure1}. When the base model is already relatively safe on a jailbreak set (e.g., JailbreakV-28k \cite{luo2024jailbreakv} for Qwen-1.5B-Instruct \cite{yang2025qwen2}), TTRL amplifies refusals (top-left, Figure \ref{fig:Figure1}) yet incurs a consistent \textbf{reasoning tax} (top-right). An attacker can also induce harmfulness amplification (shown in the bottom left Figure \ref{fig:Figure1}) even when the base model is not highly vulnerable on the held-out jailbreak set, by coupling harmful and benign subqueries within the same prompt (HarmInject). Across both cases, the mechanism is consistent: promoting self-consistency using majority rewards reinforces the dominant base model behavior on the injected prompts at the expense of reasoning.

Note that test-time adaptation (TTA) risks and self-consistency biases are known themes. Our aim is not to re-establish that test-time adaptation can be poisoned, but to study the specific vulnerabilities of test-time training methods aimed at improving the reasoning of LLMs when the test-time data itself is injected with jailbreaks, and to \emph{jointly} analyze reasoning and safety under this setting. To our knowledge, this intersection has not been investigated. Our contributions are as follows:
\begin{enumerate}[leftmargin=*]
\item We show that harmful prompt injection and even benign prompt injection during TTRL amplifies the model’s existing behaviors: \textbf{safety amplification} when the base model is safe, \textbf{harmfulness amplification} when it is vulnerable. In both cases, there is a decline in reasoning gains from TTRL, which we refer to as the \textbf{reasoning tax}.
\item We show that TTRL can be exploited adversarially. Specially designed “HarmInject” prompts force the model to answer jailbreak and reasoning queries together, yielding stronger harmfulness amplification.
\item Finally, we show that simple filtering techniques to mitigate these safety and reasoning vulnerabilities are not enough, highlighting the need for developing more sophisticated TTT methods.
\end{enumerate}

\begin{figure*}[t]
    \centering
    % --- First row: Qwen results ---
    \subfloat[]{%
        \includegraphics[width=0.3\textwidth]{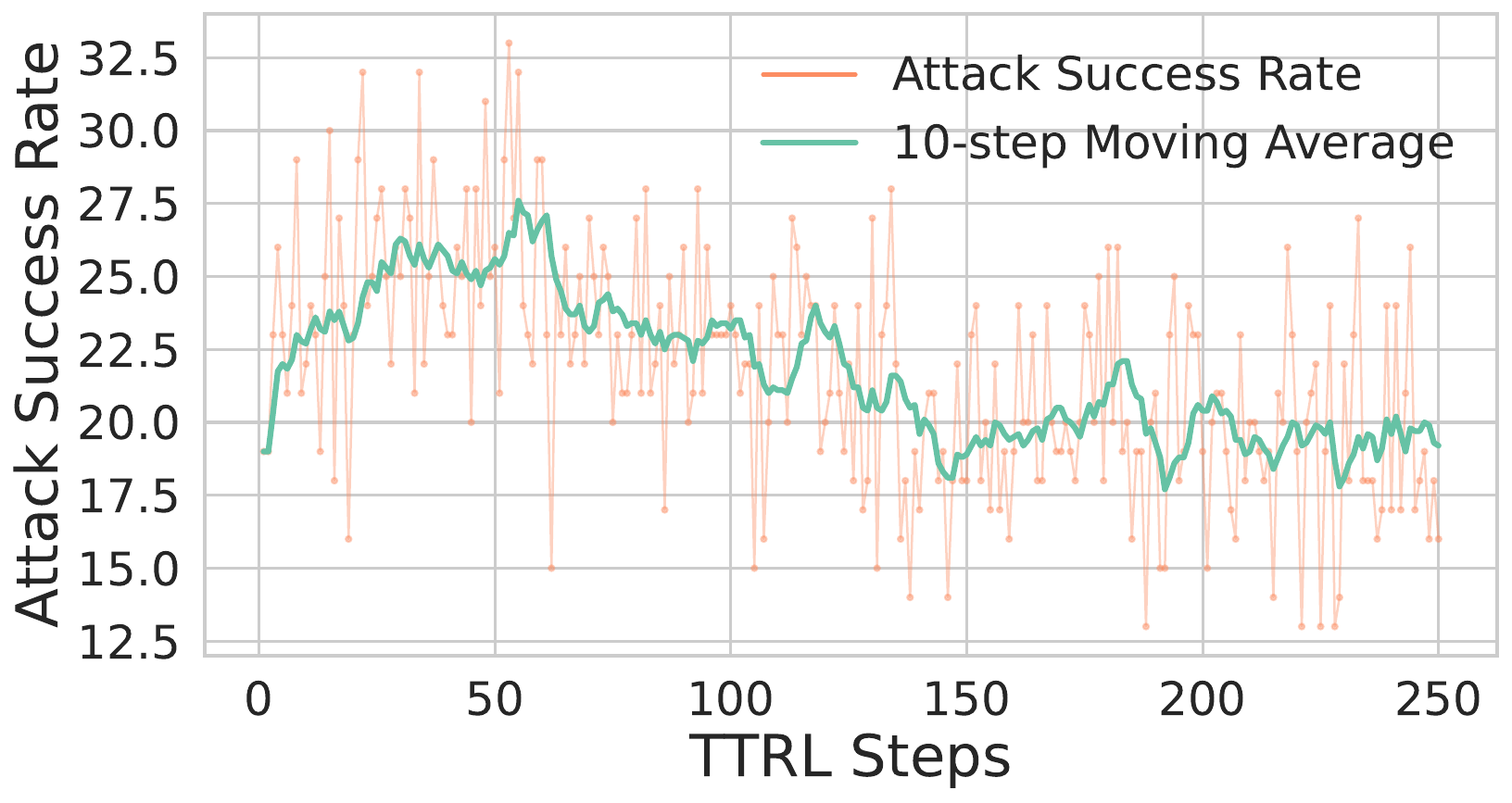}\label{subfig:amc_jailbreak2_qwen1.5binstruct}
    }
    \hfill
    \subfloat[]{%
        \includegraphics[width=0.3\textwidth]{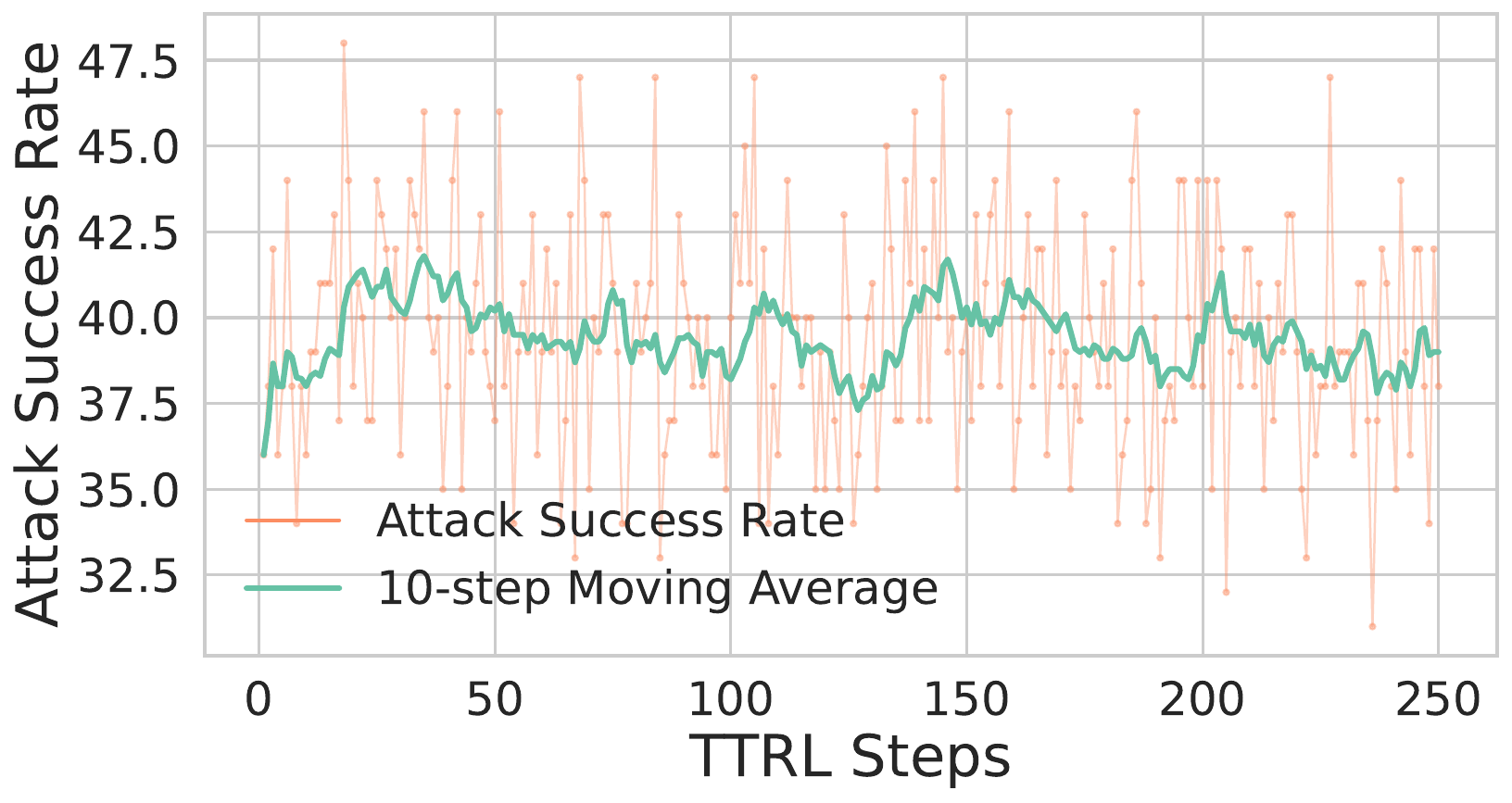}\label{subfig:amc_wildjailbreak_qwen1.5binstruct}
    }
    \hfill
    \subfloat[]{%
        \includegraphics[width=0.3\textwidth]{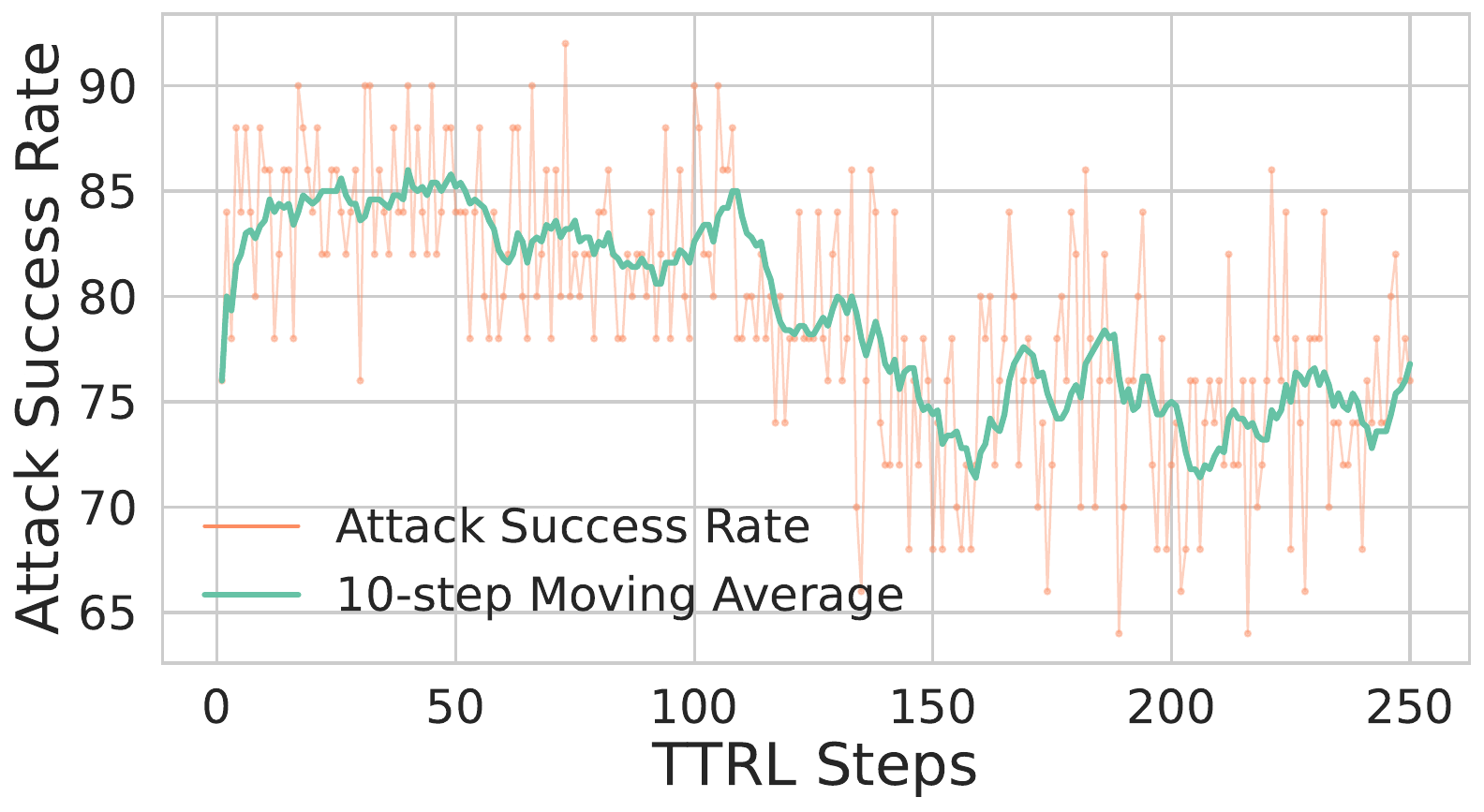}\label{subfig:amc_llamaartifacts_qwen1.5binstruct}
    }

    % --- Second row: Llama results ---
    
    \subfloat[]{%
        \includegraphics[width=0.3\textwidth]{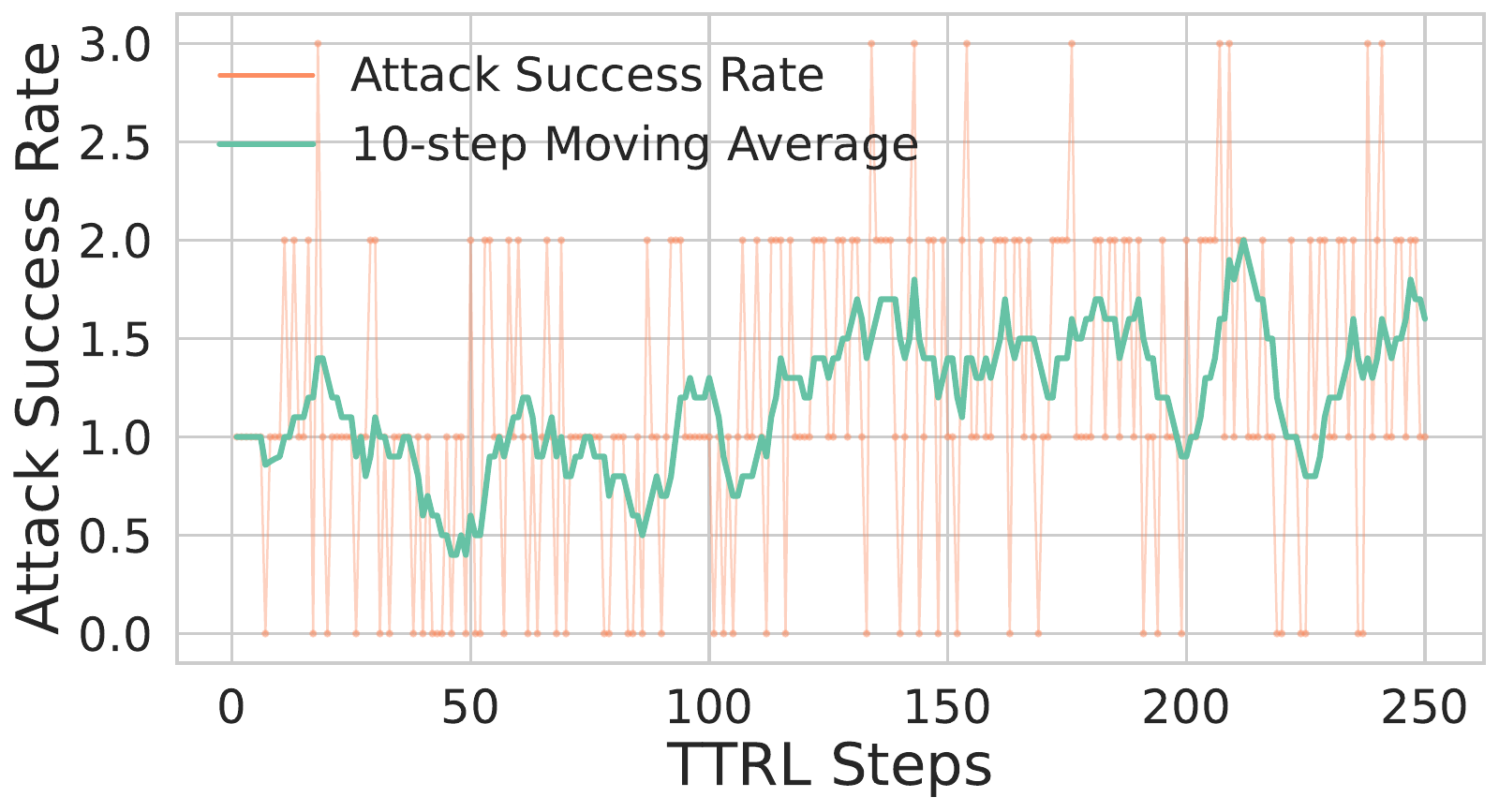}\label{subfig:amc_jailbreak2_llama8binstruct}
    }
    \hfill
    \subfloat[]{%
        \includegraphics[width=0.3\textwidth]{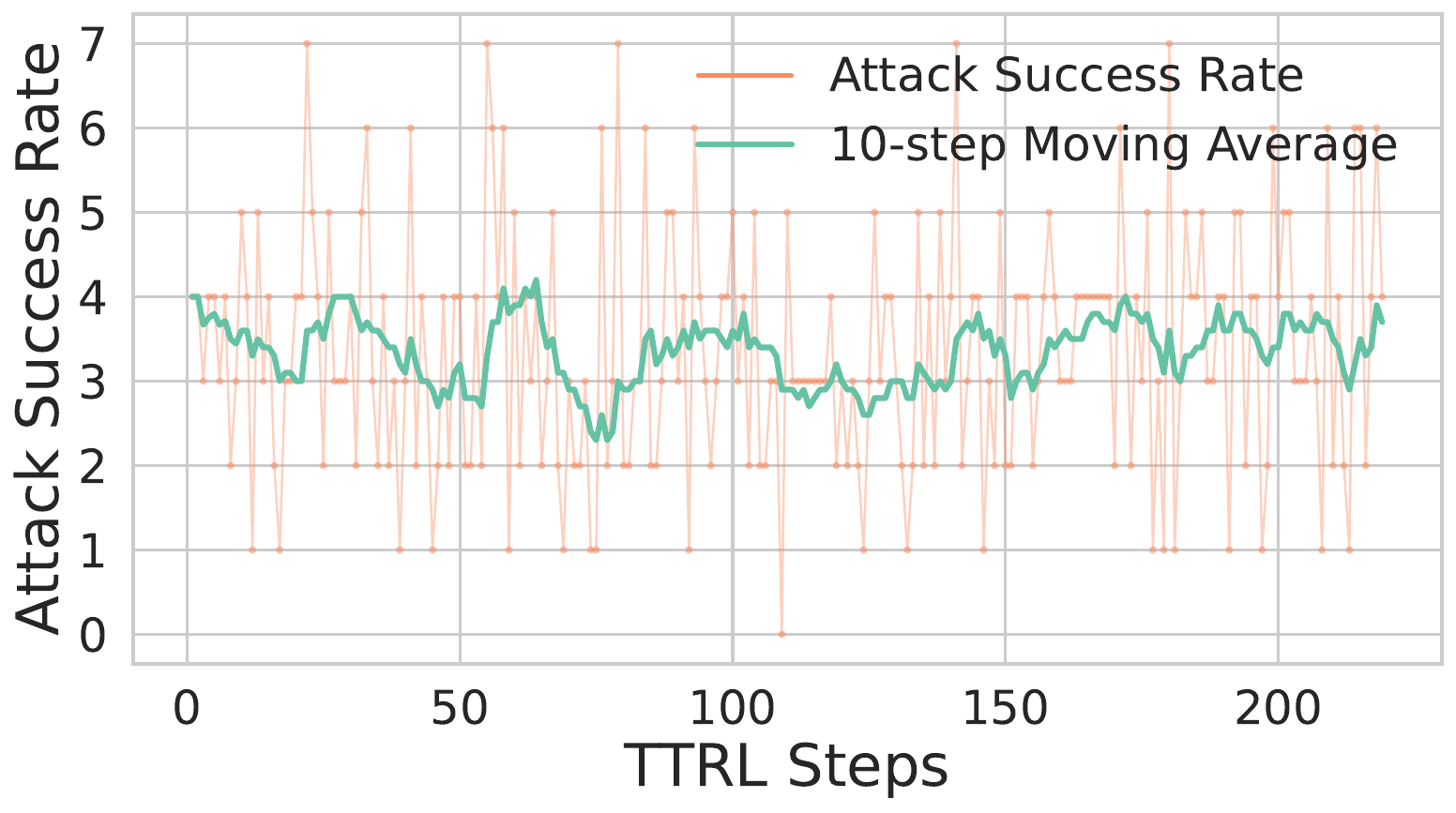}\label{subfig:amc_wildjailbreak_llama8binstruct}
    }
    \hfill
    \subfloat[]{%
        \includegraphics[width=0.3\textwidth]{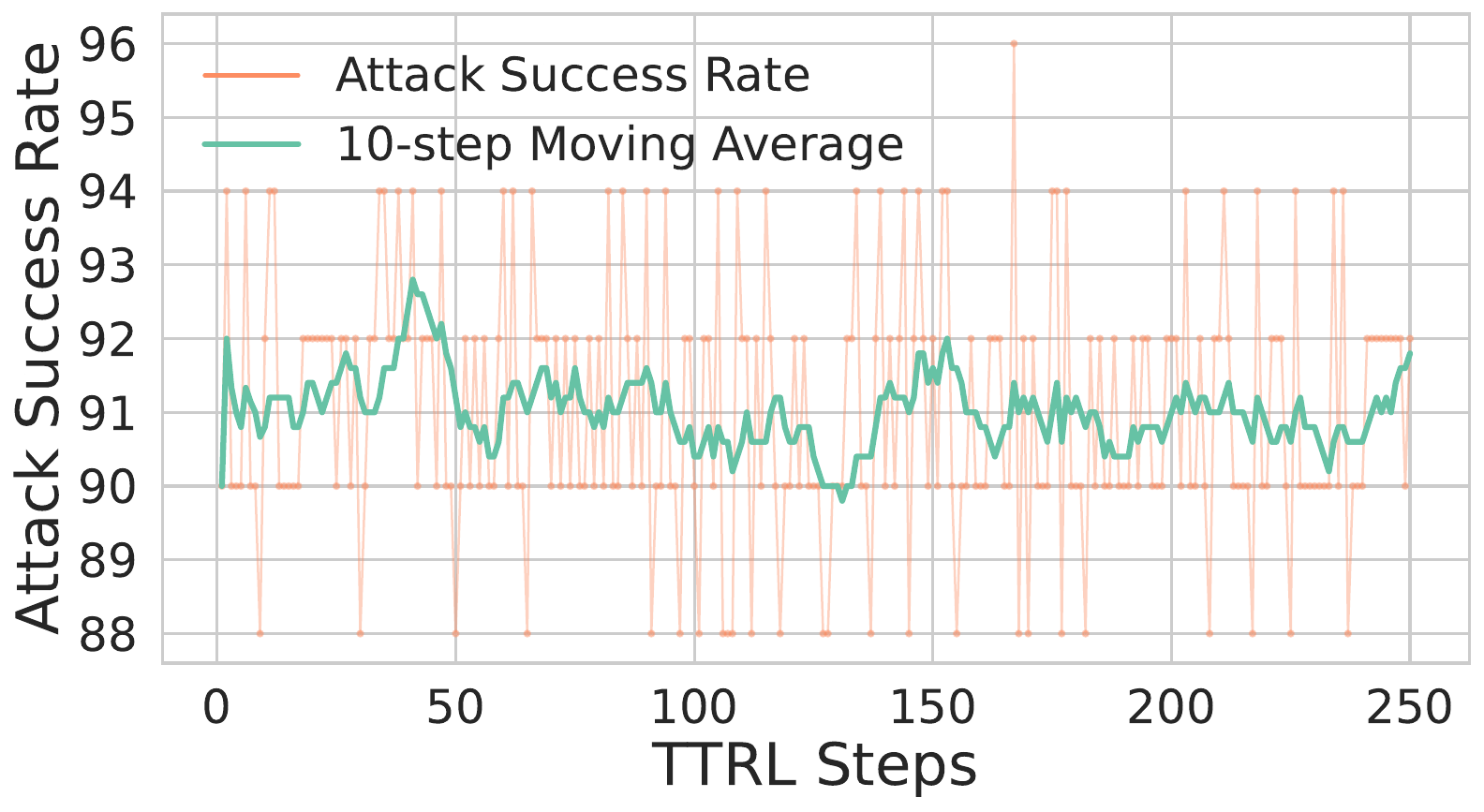}\label{subfig:amc_llamaartifacts_llama8binstruct}
    }

    \caption{ASR measured across three jailbreak datasets, JailbreakV-28k, WildJailbreak, and Llama Artifacts (left to right, respectively) during TTRL, for Qwen-1.5B-Instruct (top row) and Llama-3-8B-Instruct (bottom row).}
    \label{fig:rq1_results}
\end{figure*}

\section{Background}
\label{sec:Background}

\paragraph{Test-Time Training.}
Test-Time Training (TTT) is a method for adapting pre-trained models at test time, without access to labels or verifiable rewards. Suppose a model $f_\theta$ has been trained on a source distribution $\mathcal{D}_s = \{(x_i, y_i)\}_{i=1}^N$, where $\theta$ denotes the model parameters. At test-time, the model is evaluated on samples $x_t \sim \mathcal{D}_t$ drawn from a target distribution that may differ from the training distribution ($\mathcal{D}_t \neq \mathcal{D}_s$). Standard inference uses fixed parameters $\theta$, which can result in degraded performance under distribution shifts. TTT adapts the model at test-time by updating parameters with respect to an auxiliary objective $\mathcal{L}_{aux}$ that is optimized on each test sample. Therefore, TTT allows the model to refine itself during deployment, narrowing the gap between pre-training and testing-time.

\paragraph{Test-Time Reinforcement Learning.}
Test-time reinforcement learning (TTRL)~\cite{zuo2025ttrl} is a recent TTT method that uses the test-time data to update the model parameters in order to improve the reasoning abilities of LLMs. For each test input $x$, the model generates $K$ candidate responses $\{y_1, \ldots, y_K\}$ by sampling from its current policy $\pi_\theta$. A majority-voting aggregator selects the majority answer $\hat{y}$ across the $K$ samples. This serves as a pseudo-label: responses matching $\hat{y}$ receive a positive reward, while others receive a negative reward. The model parameters are then updated via policy gradient using Group Reference Policy Optimization (GRPO)~\cite{shao2024deepseekmath} to increase the likelihood of responses consistent with the pseudo-label. Formally, the reward at test time for the $k$-th generated response $y_k$ is given as
\[
r(y_k) = \begin{cases}
+1 & \text{if } y_k = \hat{y}, \\
0 & \text{otherwise,}
\end{cases}
\]
and the policy is trained to maximize expected reward
by updating the parameters $\theta$ to ascend the reward gradient given by
\[
\nabla_\theta J(\theta) = \mathbb{E}_{y \sim \pi_\theta(\cdot|x)} \big[ r(y) \nabla_\theta \log \pi_\theta(y|x) \big].
\]
Through repeated updates over test-time inputs, TTRL reinforces the model’s self-consistent predictions without requiring access to ground-truth labels. As TTT methods such as TTRL rely on the test data and their generated pseudo-labels, their behavior depends strongly on what appears in the test-time data. This makes it vulnerable to prompt injection attacks, which motivates the safety analysis presented in this work.

\section{Experimental setup}
\label{sec:experiments}

\textbf{Models and training datasets.} To investigate the safety and reasoning vulnerabilities of the TTRL method, we consider instruction-tuned models from the Qwen and Llama families. More specifically, Qwen2.5-0.5B-Instruct and Qwen2.5-1.5B-Instruct from the Qwen family \cite{qwen2025qwen25technicalreport}, Llama3.2-1B-Instruct, Llama3.2-3B-Instruct, and Llama3-8B-Instruct models from the Llama family \cite{grattafiori2024llama}. For the harmfulness evaluation, we consider three jailbreak datasets: JailbreakV-28k \cite{luo2024jailbreakv}, Llama-jailbreak artifacts \cite{andriushchenko2024jailbreaking} specifically tuned to jailbreak the Llama3-8B-Instruct model, and the in-the-wild jailbreak dataset \cite{jiang2024wildteaming}. We conduct all experiments on the AMC reasoning dataset \cite{li2024numinamath}, chosen for its scale and relevance as a benchmark for mathematical reasoning.

\textbf{Threat model and prompt injection.} We consider prompt injection of harmful jailbreak prompts into the test-time data to evaluate the harmfulness of the model during TTRL. For all the prompt injection experiments, we form a mixed test-time dataset by augmenting AMC with harmful prompts, whose count is 60\% of the AMC dataset size, unless otherwise stated.

\textbf{Evaluation metrics and TTRL parameters.} To measure the reasoning performance, we use the same metric as used in the TTRL paper~\cite{zuo2025ttrl}, i.e., \textit{pass@1} estimated from $k=16$ responses, generated with a non-zero temperature of $0.6$, top-p value of $0.95$, and given by $pass@1 := c/k$, where $c$ is the number of correct responses. To measure the safety of the underlying model, we evaluate the attack success rate (ASR) percentage of the jailbreak attack on the model, which measures the percentage of harmful responses to the total number of jailbreak prompts. We use the LlamaGuard3-8B model \cite{inan2023llama} as a safety judge to evaluate the ASR percentage. Each TTRL run is for 250 training steps (unless otherwise stated), with 64 votes per prompt to get the majority vote, and with a training batch size of 8 prompts per rollout. For more details, refer to Appendix \ref{sec:TTRL_parameters}. 

\section{Experimental results}
We structure the experimental results section into five research questions: \textbf{RQ1:} How does TTRL on reasoning questions affect the harmfulness of the model?; \textbf{RQ2:} What is the impact of harmful prompt injection during TTRL?; \textbf{RQ3:} How does benign prompt injection affect the harmfulness during TTRL?; \textbf{RQ4:} Can TTRL be exploited to compromise the harmfulness of the model?; \textbf{RQ5:} Can simple filtering help mitigate TTRL vulnerabilities?

\begin{table}[t]
\centering
\caption{ASR (\%) on three jailbreak datasets before and after TTRL on AMC reasoning questions. Here, Init. is the initial ASR (\%), Final is the post-TTRL ASR (\%), and $\Delta$ is Final $-$ Init (percentage points), on respective jailbreak datasets. WJB denotes Wildjailbreak and Artifacts denotes the Llama artifacts dataset.}
\label{tab:rq1_table}
\setlength{\tabcolsep}{2.5pt}
\renewcommand{\arraystretch}{0.85}

\begin{tabular}{@{}l@{\hspace{2pt}}ccccccccc@{}}
\toprule
\textbf{Model}
& \multicolumn{3}{c}{\textbf{JailbreakV-28k}}
& \multicolumn{3}{c}{\textbf{WJB}}
& \multicolumn{3}{c}{\textbf{Artifacts}} \\
\cmidrule(lr){2-4}\cmidrule(lr){5-7}\cmidrule(lr){8-10}
& Init. & Final & $\Delta$
& Init. & Final & $\Delta$
& Init. & Final & $\Delta$ \\
\midrule
Qwen-0.5B & 27 & 24 & {\color{asrgreen}-3}
                & 40 & 34 & {\color{asrgreen}-6}
                & 70 & 67 & {\color{asrgreen}-3} \\

Qwen-1.5B & 22 & 21 & {\color{asrgreen}-1}
                & 36 & 38 & {\color{asrred}+2}
                & 76 & 77 & {\color{asrred}+1} \\
\midrule 
Llama-1B  &  9 &  7 & {\color{asrgreen}-2}
                & 20 & 19 & {\color{asrgreen}-1}
                &  0 &  0 & 0 \\

Llama-3B  &  5 &  4 & {\color{asrgreen}-1}
                & 35 &  5 & {\color{asrgreen}-30}
                &  4 &  6 & {\color{asrred}+2} \\

Llama-8B  &  1 &  1 & 0
                &  4 &  4 & 0
                & 90 & 92 & {\color{asrred}+2} \\
\bottomrule
\end{tabular}
\end{table}

\begin{figure*}[t]
    \centering
    % --- First row: Qwen results ---
    \subfloat[]{%
        \includegraphics[width=0.3\textwidth]{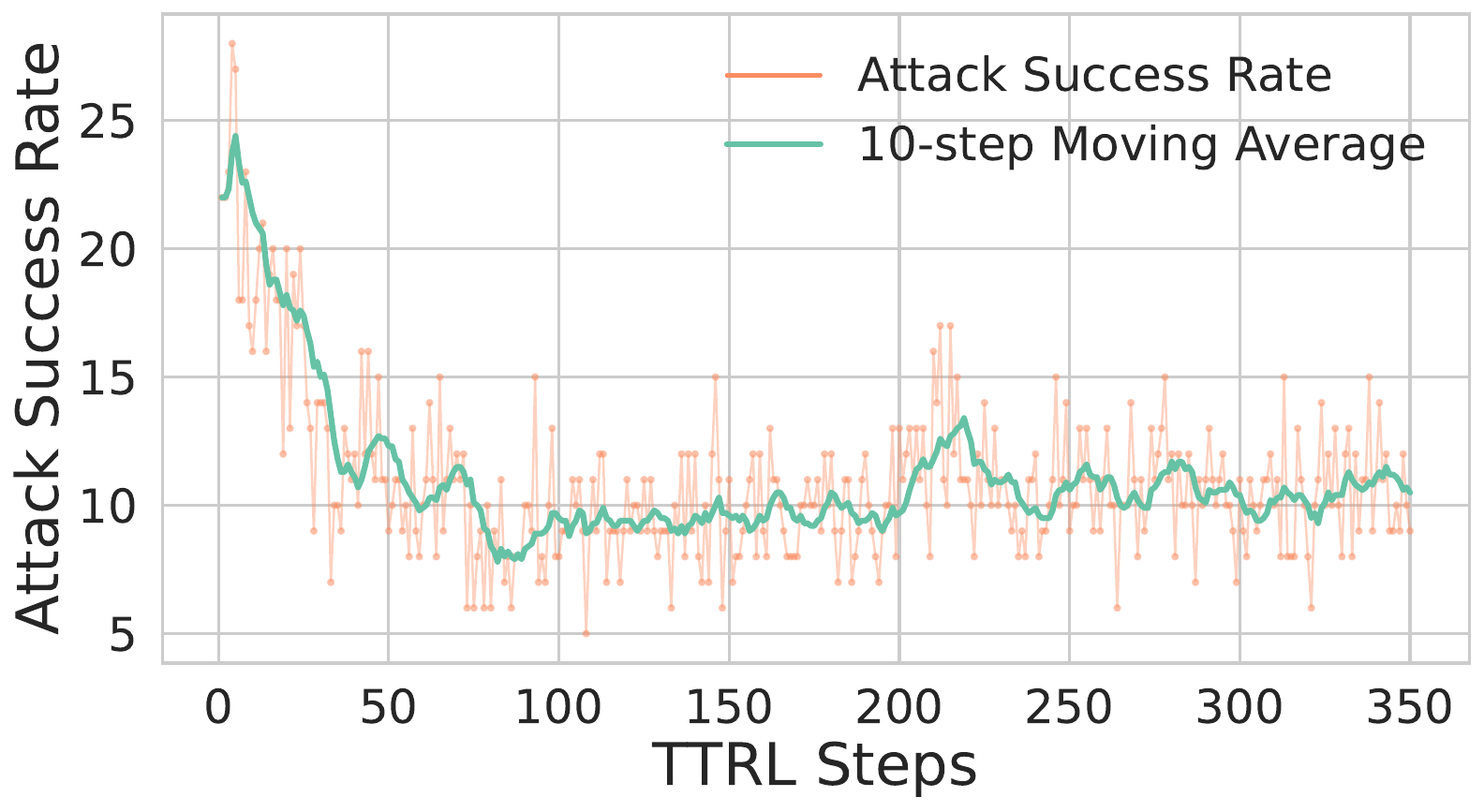}
    \label{subfig:amcjail_jailbreak2_qwen1.5binstruct}}
    \hfill
    \subfloat[]{%
        \includegraphics[width=0.3\textwidth]{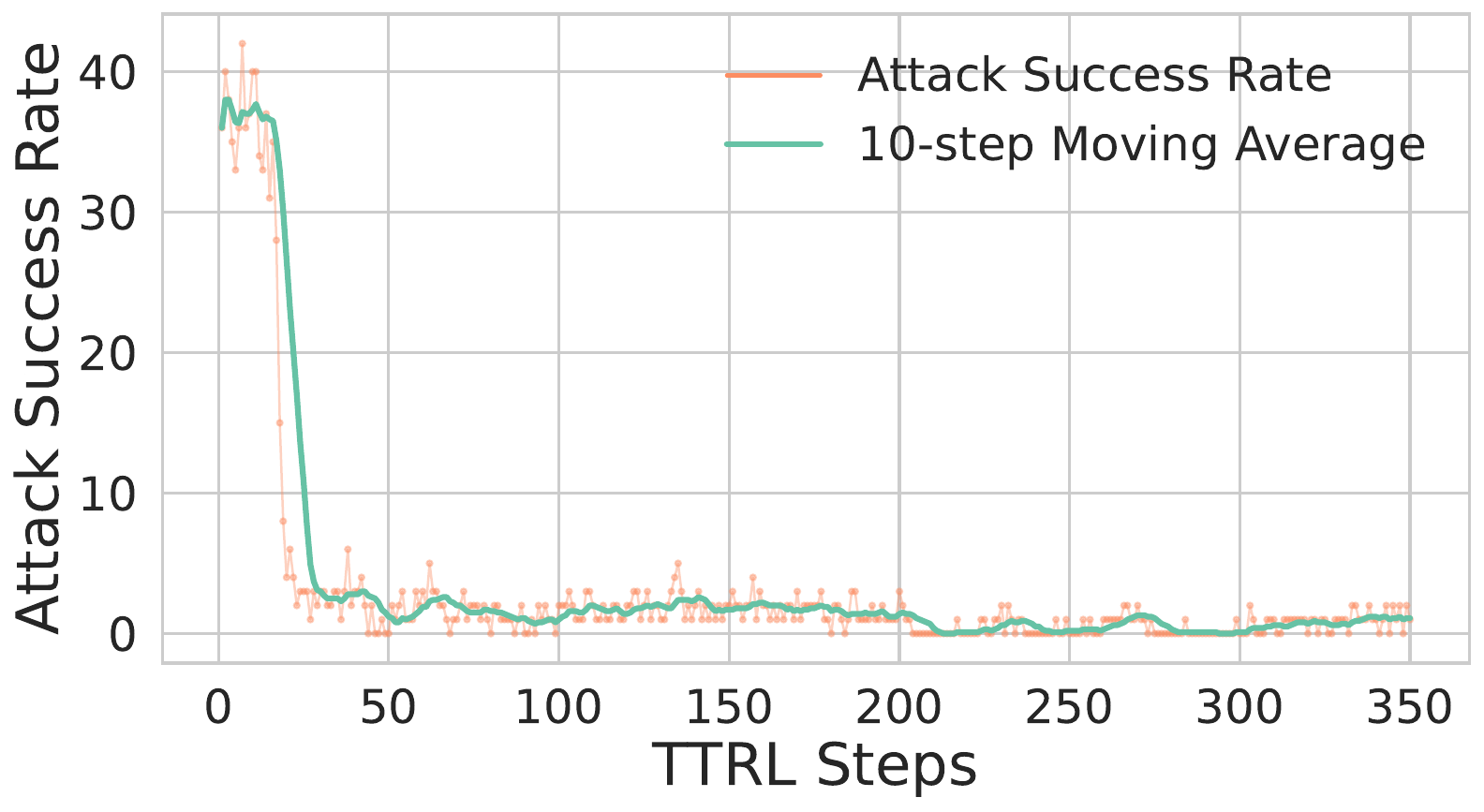}\label{subfig:amcwildjail_wildjail_qwen1.5binstruct}
    }
    \hfill
    \subfloat[]{%
        \includegraphics[width=0.3\textwidth]{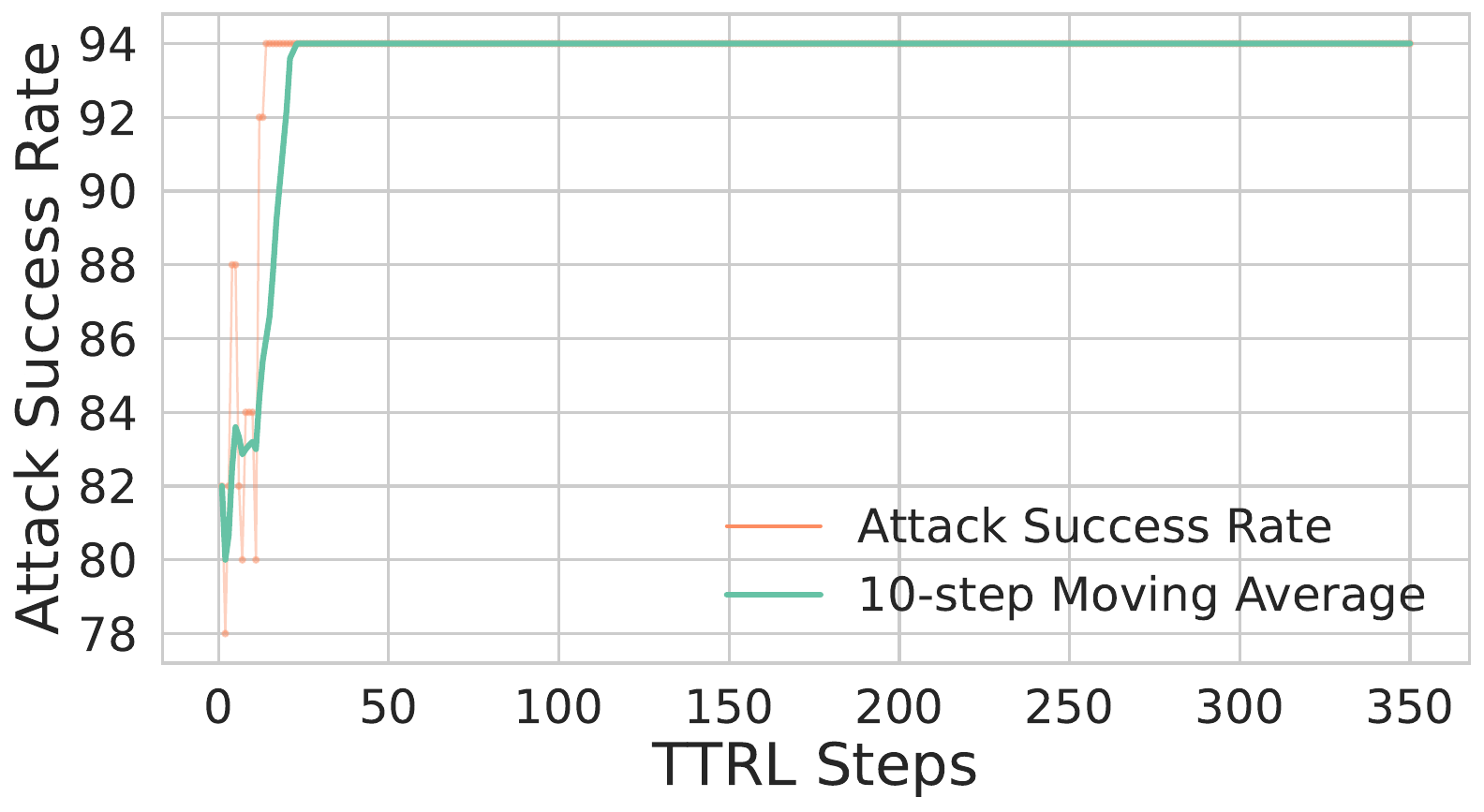}\label{subfig:amcllamaartifacts_llamaartifacts_qwen1.5binstruct}
    }

    % --- Second row: Llama results ---
    
    \subfloat[]{%
        \includegraphics[width=0.3\textwidth]{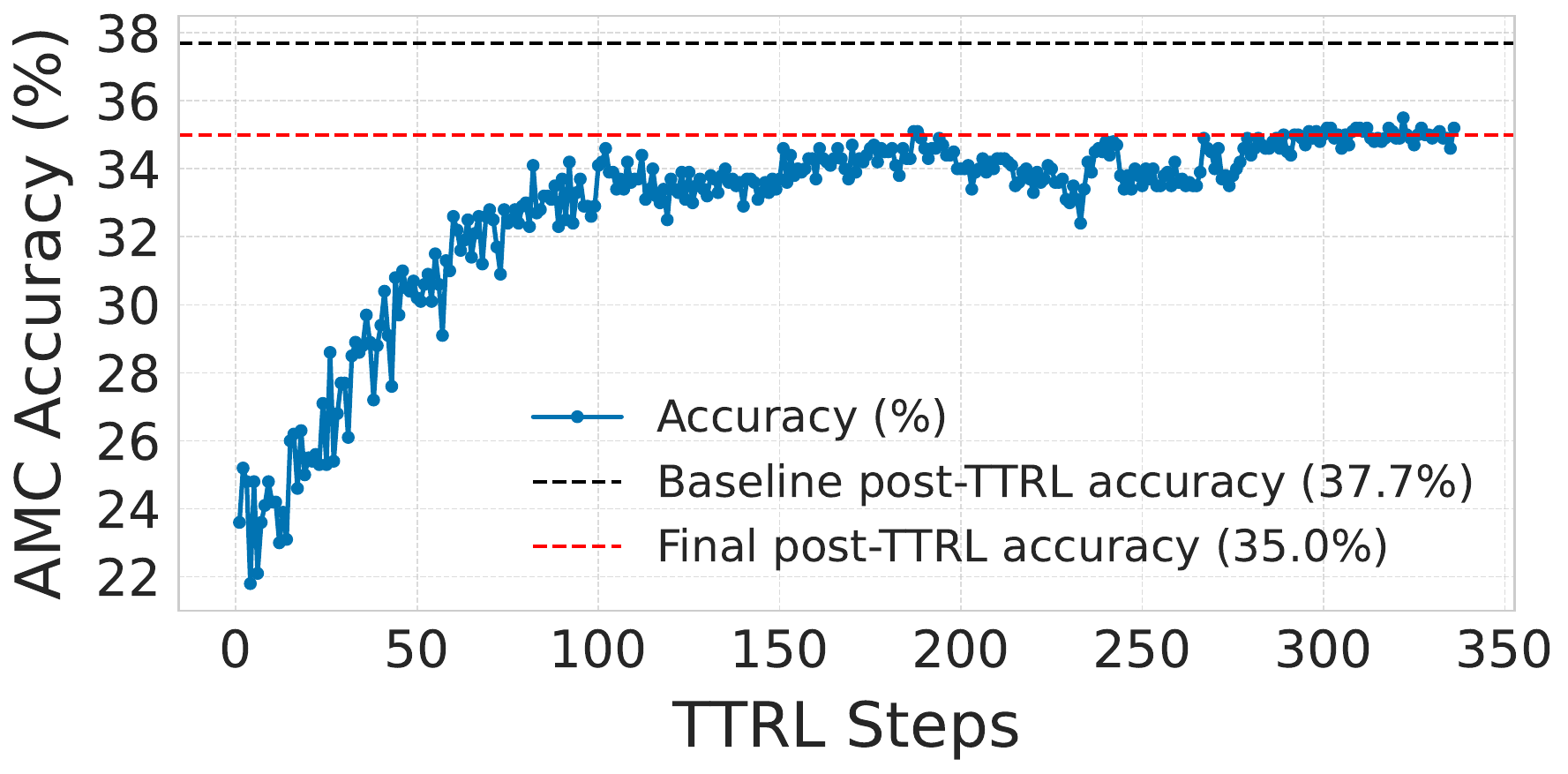}\label{subfig:amcjail_amc_qwen1.5binstruct}
    }
    \hfill
    \subfloat[]{%
        \includegraphics[width=0.3\textwidth]{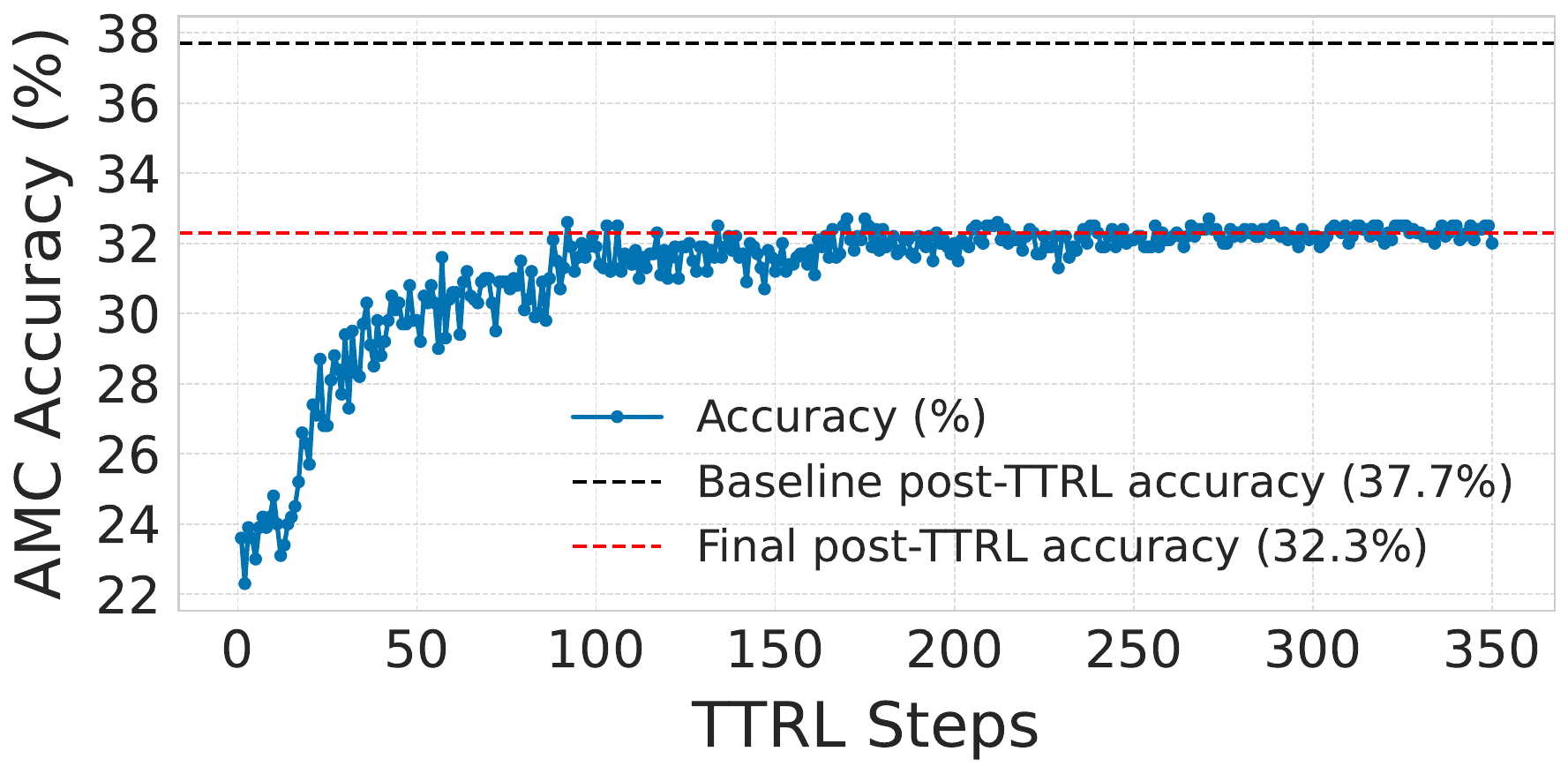}\label{subfig:amcwildjail_amc_qwen1.5binstruct}
    }
    \hfill
    \subfloat[]{%
        \includegraphics[width=0.3\textwidth]{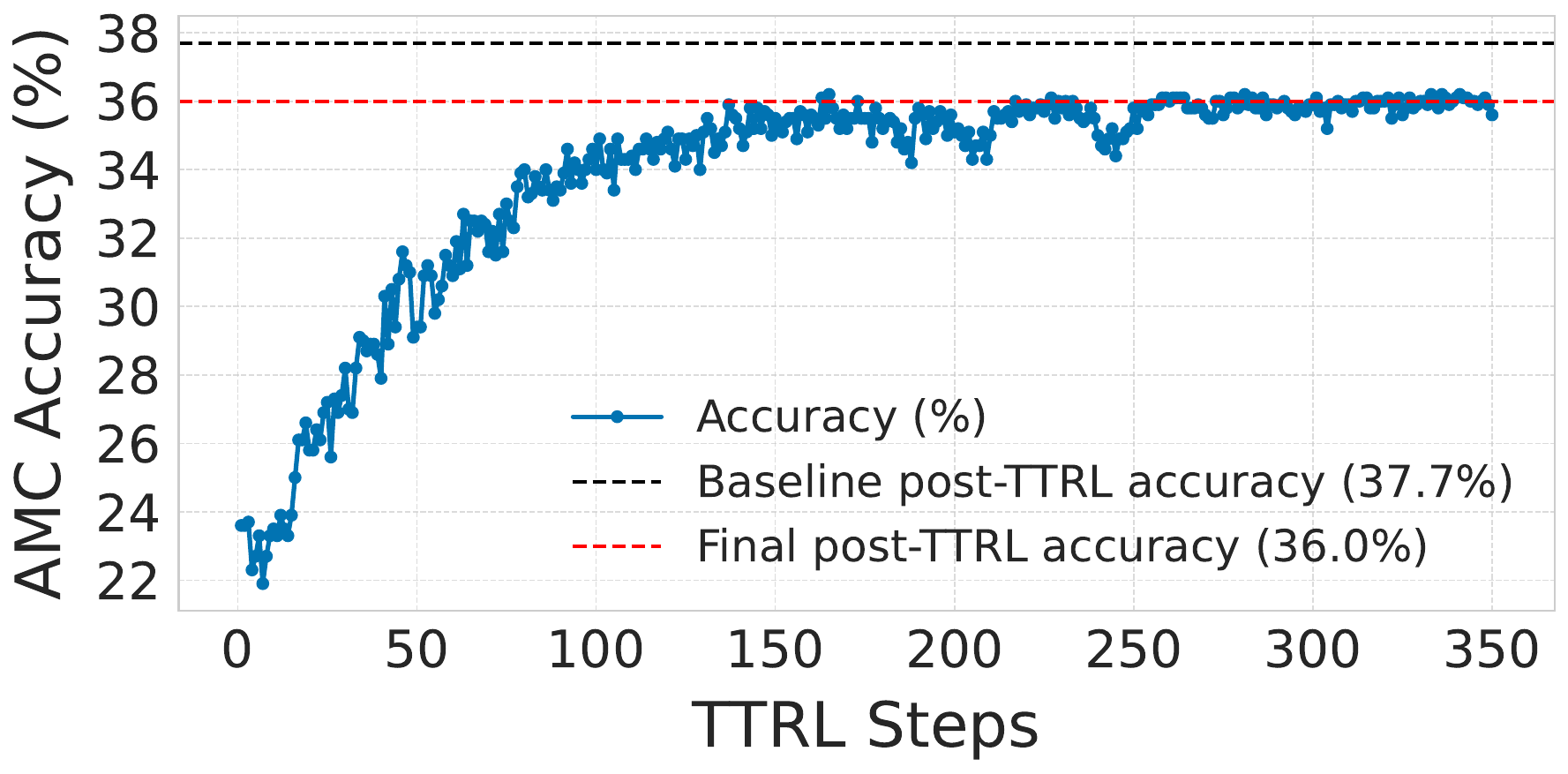}\label{subfig:amcllamaartifacts_amc_qwen1.5binstruct}
    }

    \caption{Impact on safety and reasoning for Qwen-1.5B-Instruct model after harmful prompt injection across three jailbreak datasets, JailbreakV-28k, WildJailbreak, and Llama Artifacts (left to right, respectively) during TTRL, for safety (top row) and AMC accuracy (bottom row).}
    \label{fig:rq2_qwen1.5b_3attacks}
\end{figure*}

\begin{table*}[t]
\centering
\caption{\textbf{Safety/harmfulness amplification} given by ASR (\%) before and after TTRL under different jailbreak prompt injections and validated on the same held-out jailbreak prompts. $\Delta$ denotes amplification magnitude in percentage points (Final $-$ Initial).}
\label{tab:rq2_saf_amplification}
\setlength{\tabcolsep}{3pt}
\renewcommand{\arraystretch}{1}

\begin{tabular}{@{}lccccccccc@{}}
\toprule
\textbf{Model}
& \multicolumn{3}{c}{\textbf{JailbreakV-28k injection}}
& \multicolumn{3}{c}{\textbf{WildJailbreak injection}}
& \multicolumn{3}{c}{\textbf{LlamaArtifacts injection}} \\
\cmidrule(lr){2-4}\cmidrule(lr){5-7}\cmidrule(lr){8-10}
& Init. ASR & Final ASR & $\Delta$
& Init. ASR & Final ASR & $\Delta$
& Init. ASR & Final ASR & $\Delta$ \\
\midrule
Qwen-0.5B-Instruct 
& 27 & 5  & {\color{asrgreen}-22}
& 41 & 3  & {\color{asrgreen}-38}
& 74 & 0  & {\color{asrgreen}-74} \\

Qwen-1.5B-Instruct 
& 22 & 10  & {\color{asrgreen}-12}
& 36 & 2  & {\color{asrgreen}-34}
& 82 & 94 & {\color{asrred}+12} \\
\midrule 
Llama-1B-Instruct  
&  9 & 2  & {\color{asrgreen}-7}
&  6 & 1  & {\color{asrgreen}-5}
& 18 & 0  & {\color{asrgreen}-18} \\

Llama-3B-Instruct
&  5 & 1  & {\color{asrgreen}-4}
& 35 &16  & {\color{asrgreen}-19}
&  8 & 0  & {\color{asrgreen}-8} \\

Llama-8B-Instruct  
&  1 & 1  & 0
&  3 & 0  & {\color{asrgreen}-3}
& 90 &88  & {\color{asrgreen}-2} \\
\bottomrule
\end{tabular}
\end{table*}

\begin{table*}[t]
\centering
\caption{\textbf{Reasoning tax:} AMC accuracy and post-TTRL (p-TTRL) performance before and after jailbreak injection across models. Init. is the initial AMC accuracy (\%), default p-TTRL is the post-TTRL accuracy (\%) without injection, +JBV-28k is the accuracy (\%) after jailbreakV-28k injection, +WJB is the wildjailbreak injection, +artifacts is the Llama artifacts injeciton, $\delta$ denotes the change relative to the default p-TTRL accuracy (percentage points).}
\label{tab:rq2_reasoning_tax}
\setlength{\tabcolsep}{7pt}
\renewcommand{\arraystretch}{0.9}

\begin{tabular}{lcc|cc|cc|ccc}
\toprule
\textbf{Model}
& \textbf{Init.}
& \textbf{Default p-TTRL}
& \textbf{+ JBV-28k} & {\small$\boldsymbol{\delta}$}
& \textbf{+ WJB}  & {\small$\boldsymbol{\delta}$}
& \textbf{+ Artifacts} & {\small$\boldsymbol{\delta}$} \\
\midrule
Qwen-0.5B-Inst.
& 8.0  & 14.8
& 10.7 & \taxdelta{-4.1}
& 13.0 & \taxdelta{-1.8}
& 13.3 & \taxdelta{-1.5} \\

Qwen-1.5B-Inst.
& 24.0 & 37.7
& 35.0 & \taxdelta{-2.7}
& 32.3 & \taxdelta{-5.4}
& 36.0 & \taxdelta{-1.7} \\

\midrule 
Llama-1B-Inst.
& 6.2  & 10.8
& 6.1  & \taxdelta{-4.7}
& 1.2  & \taxdelta{-9.6}
& 1.2 & \taxdelta{-9.6} \\

Llama-3B-Inst.
& 22.0 & 31.5
& 18.1 & \taxdelta{-13.4}
& 20.5 & \taxdelta{-11.0}
& 21.7 & \taxdelta{-9.8} \\

Llama-8B-Inst.
& 6.5  & 14.0
& 9.0  & \taxdelta{-5.0}
& 6.2  & \taxdelta{-7.8}
& 0.0  & \taxdelta{-14.0} \\
\bottomrule
\end{tabular}
\end{table*}

\subsection{RQ1:How does TTRL on reasoning questions affect the harmfulness of the model?}
\label{subsec:RQ1}

First, we investigate how TTRL on math questions (AMC in our case) affects the model's harmfulness. Figure~\ref{fig:rq1_results} reports attack success rate percentage across 250 TTRL steps for both Qwen-1.5B-Instruct and Llama-3-8B-Instruct when the test-time training data contains only AMC reasoning problems. For Qwen-1.5B-Instruct, in Figure~\ref {subfig:amc_jailbreak2_qwen1.5binstruct}, ASR on JailbreakV-28k fluctuates between 21\% and 25\% (baseline 22\%). In Figure~\ref {subfig:amc_wildjailbreak_qwen1.5binstruct} and \ref{subfig:amc_llamaartifacts_qwen1.5binstruct}, similar small variations appear on the WildJailbreak and Llama artifact attacks, with no obvious upward or downward trend across TTRL steps. In Figures~\ref{subfig:amc_jailbreak2_llama8binstruct}-\ref{subfig:amc_llamaartifacts_llama8binstruct}, for the Llama model, ASR remains between 0.5\% and 2\% on JailbreakV-28k and between 2\% and 4\% on WildJailbreak, while remaining flat around 92\% on llama artifact prompts. The initial and final ASR for all the instruction-tuned models is given in Table \ref{tab:rq1_table}, which shows the same trends of no significant effect on the ASR (with one exception of Llama3.2-3B-Instruct on the wildjailbreak prompts). 

\paragraph{Takeaway for RQ1.} The harmfulness of the underlying model does not change significantly during TTRL on the mathematical reasoning questions.

\begin{figure}[t]
    \centering
    \includegraphics[width=\columnwidth]{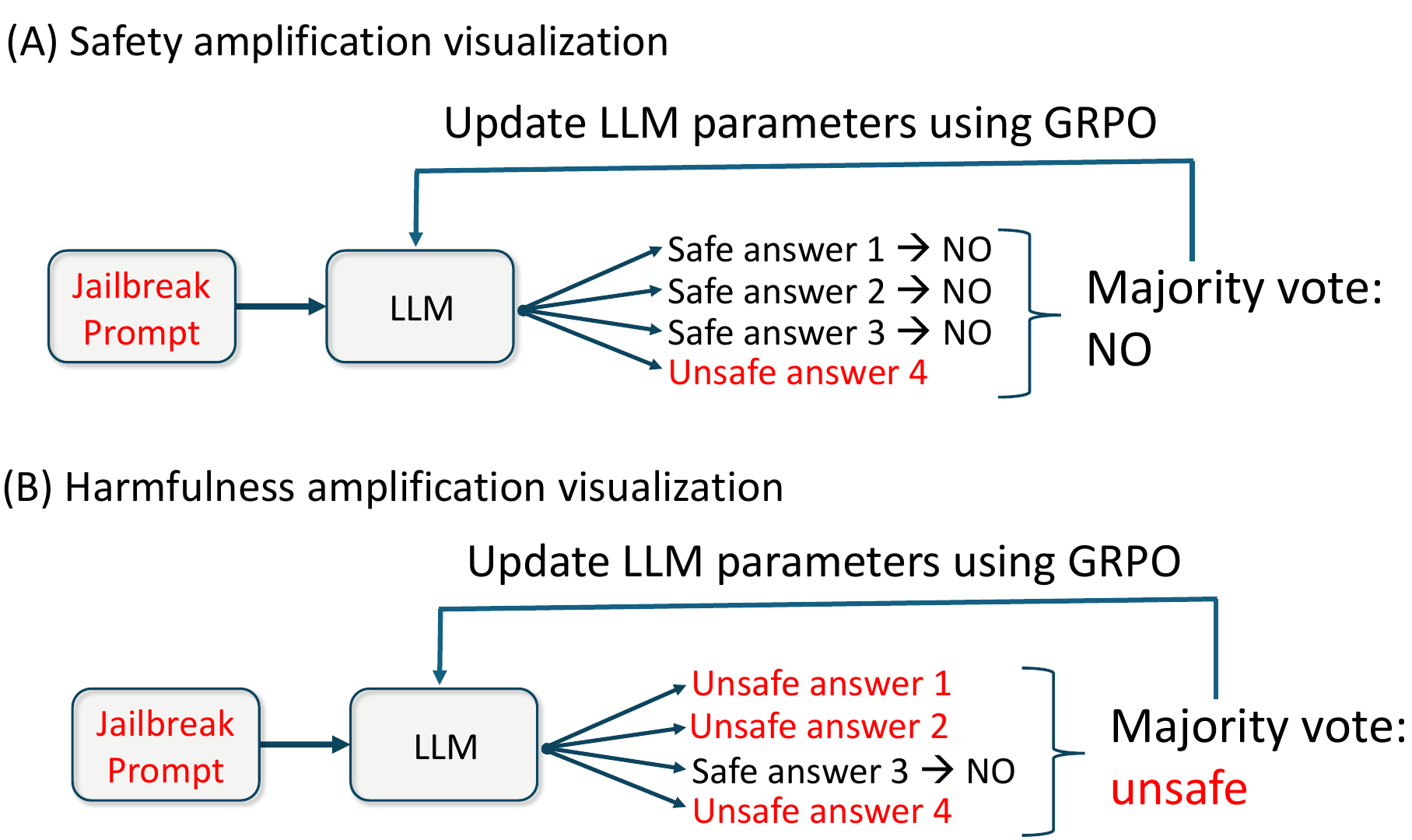}
    \caption{TTRL visualization for safety and harmfulness amplification. (A) An example case when a jailbreak prompt is encountered during TTRL, and the base model produces relatively safe answers, and the majority vote extracted label is safe, which reinforces the safe behavior leading to safety amplification. (B) Another case where the base model is relatively unsafe to the jailbreak prompt, which leads to unsafe generations, and the majority vote reinforces that behavior, leading to harmfulness amplification.}
    \label{fig:amplification_visualization}
\end{figure}

\begin{figure*}[t]
    \centering
    % --- First row: Qwen results ---
    \subfloat[]{%
        \includegraphics[width=0.3\textwidth]{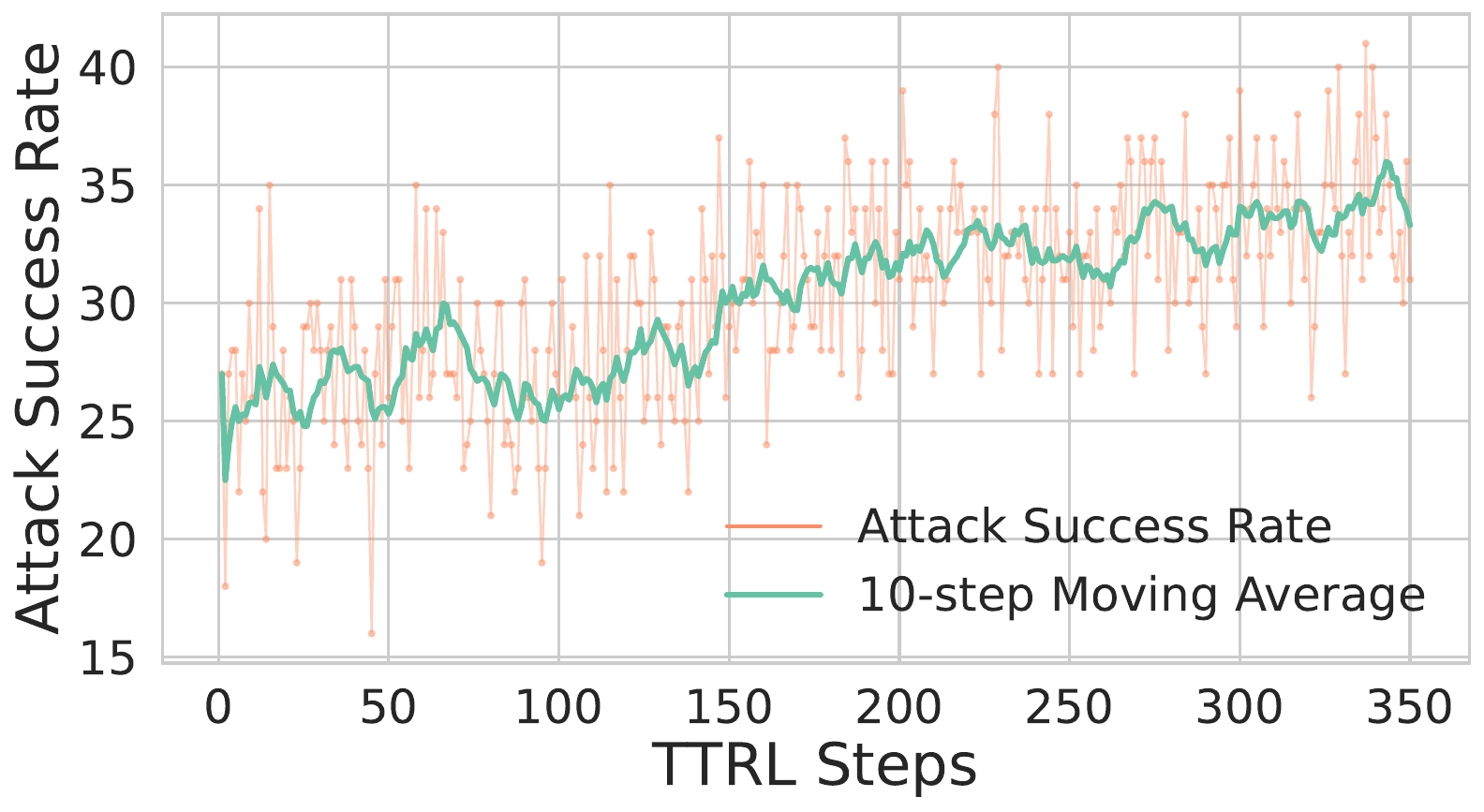}\label{subfig:amcinstruct_jailbreak_qwen0.5binstruct}
    }
    \hfill
    \subfloat[]{%
        \includegraphics[width=0.3\textwidth]{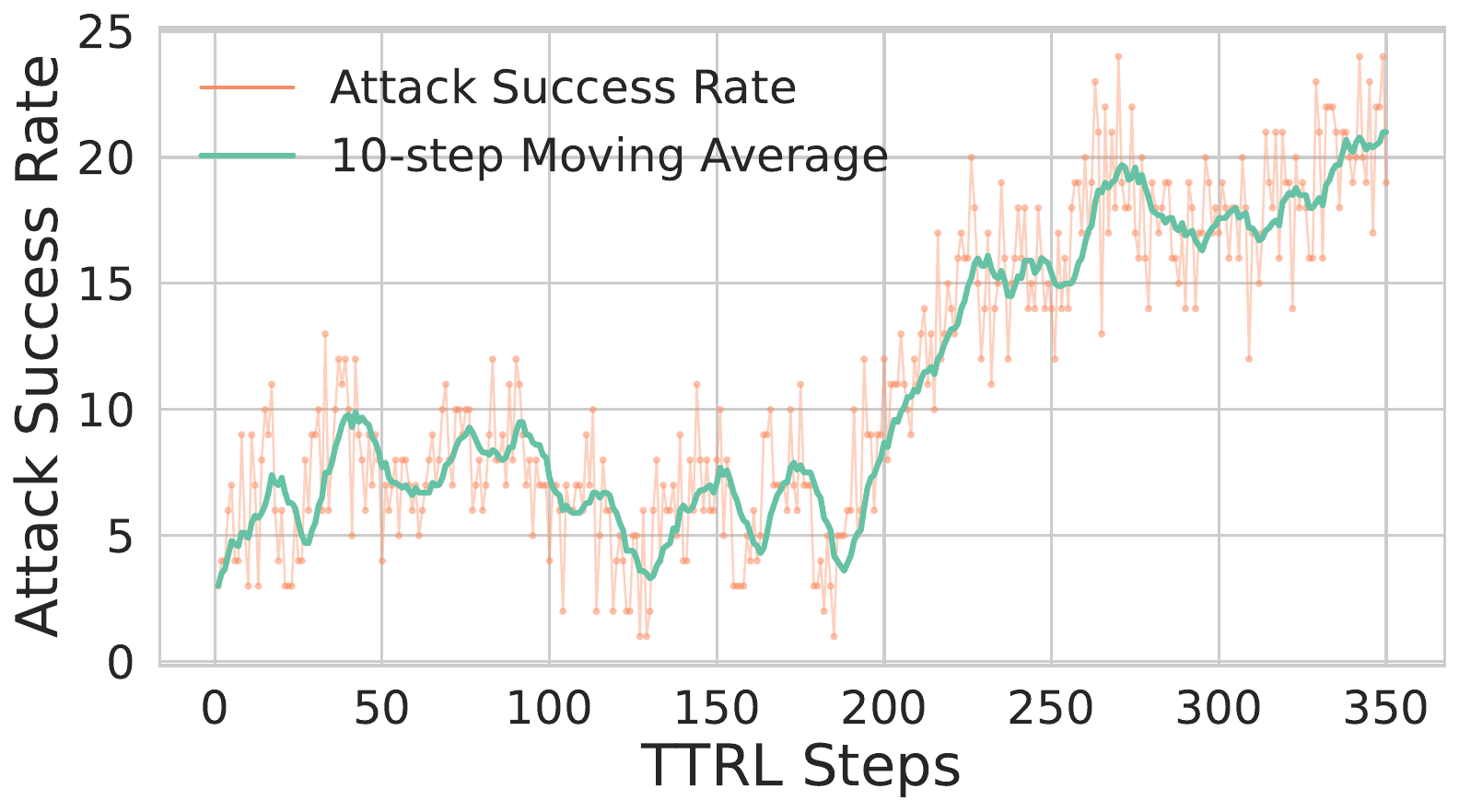}\label{subfig:amcinstruct_jailbreak_llama3binstruct}
    }
    \hfill
    \subfloat[]{%
        \includegraphics[width=0.3\textwidth]{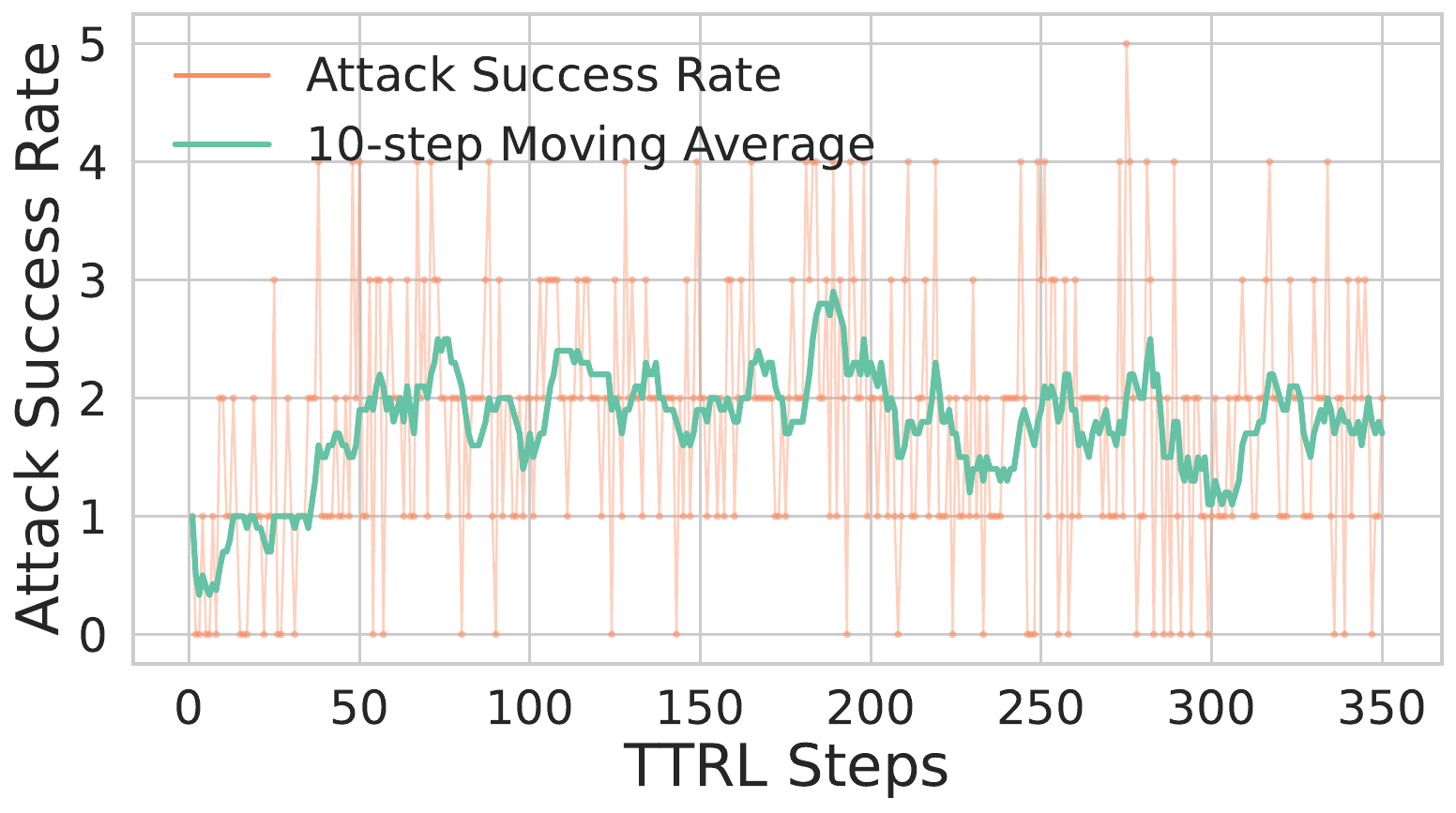}\label{subfig:amcinstruct_jailbreak_llama8binstruct}
    }

    % --- Second row: Llama results ---
    
    \subfloat[]{%
        \includegraphics[width=0.3\textwidth]{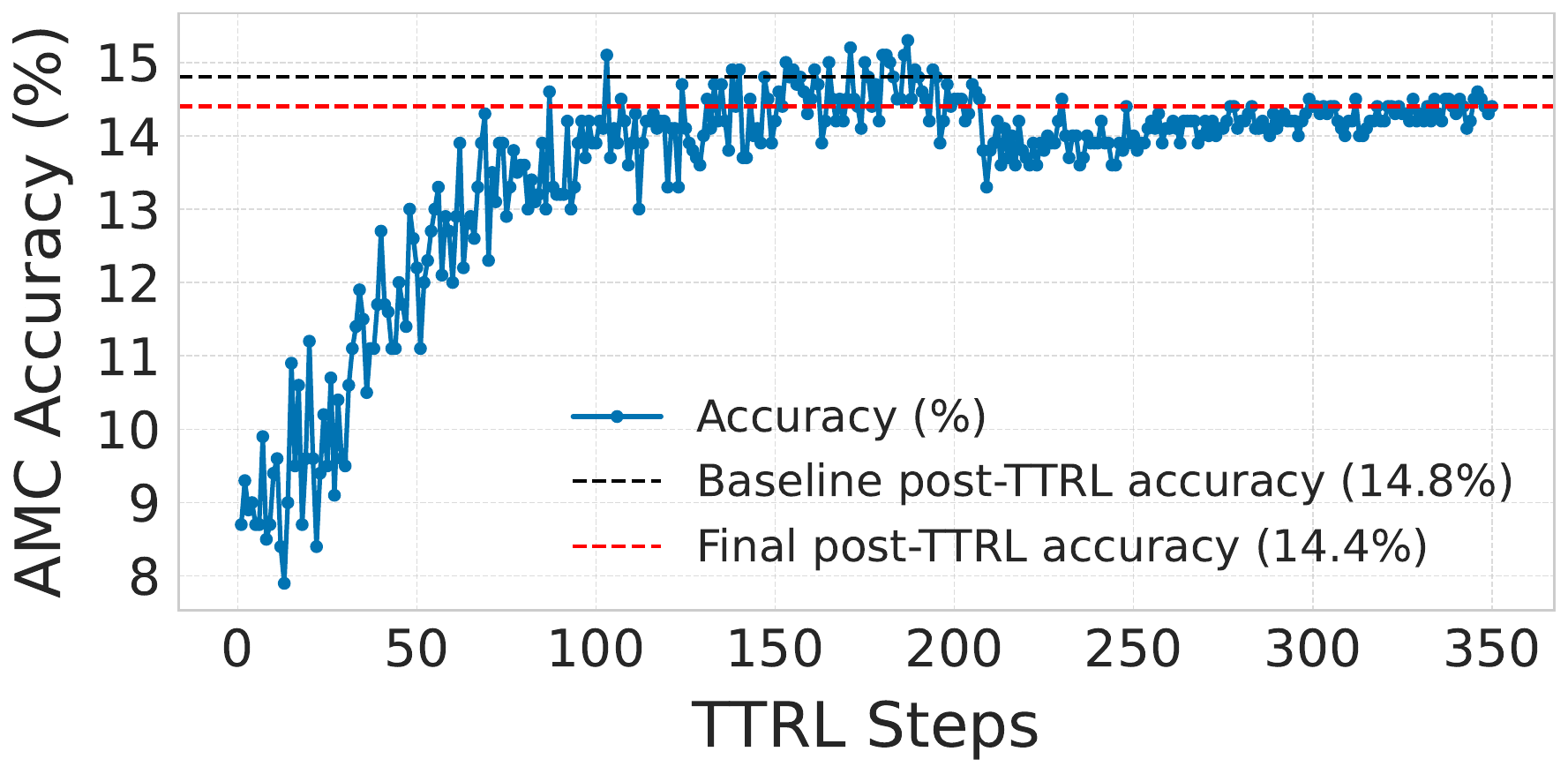}\label{subfig:amcinstruct_amc_qwen0.5binstruct}
    }
    \hfill
    \subfloat[]{%
        \includegraphics[width=0.3\textwidth]{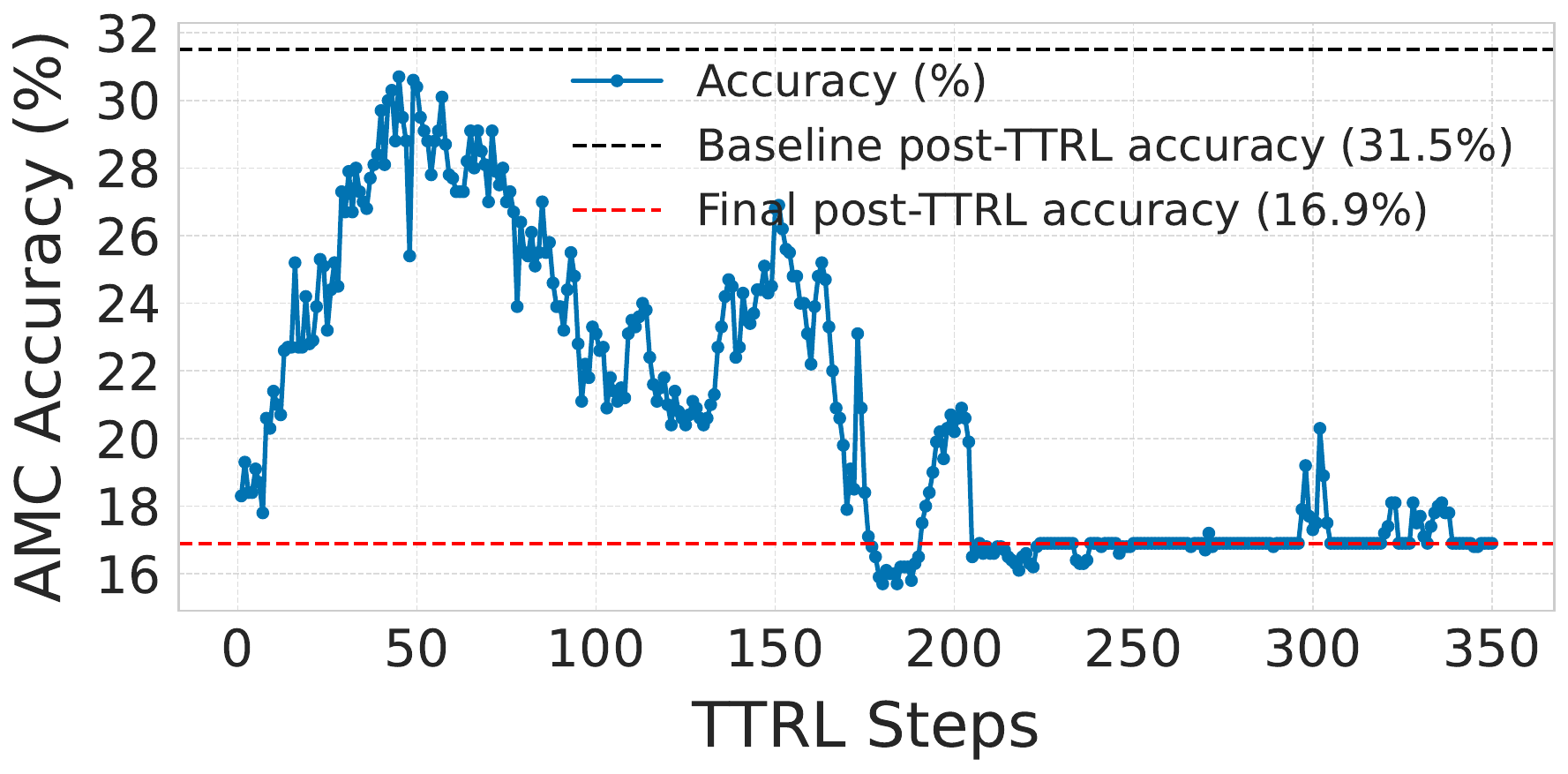}\label{subfig:amcinstruct_amc_llama3binstruct}
    }
    \hfill
    \subfloat[]{%
        \includegraphics[width=0.3\textwidth]{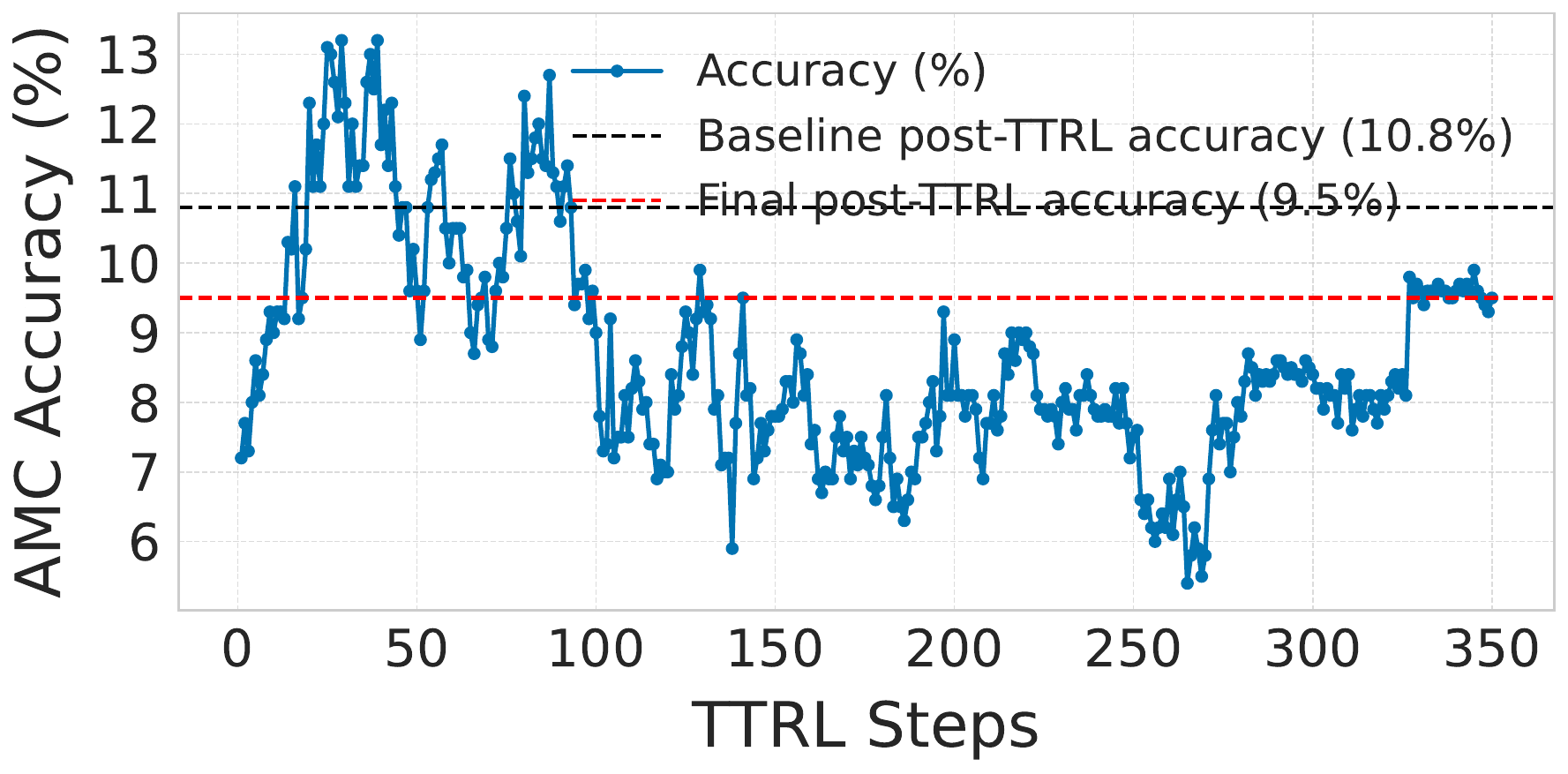}\label{subfig:amcinstruct_amc_llama8binstruct}
    }

    \caption{Impact on safety (top row) and reasoning (bottom row) for Qwen-0.5B-Instruct, Llama3.2-3B-Instruct, and Llama3-8B-Instruct models (left to right) after injecting benign instruction-following prompts. The ASR is reported on the JailbreakV-28k prompts.}
    \label{fig:rq3_harmfulness_amplifications}
\end{figure*}

\subsection{RQ2: What is the impact of harmful prompt injection during TTRL?}
\label{subsec:RQ2}
Next, we turn to the case where the test-time training data is injected with harmful prompts, i.e., the data consists of both AMC and jailbreak prompts. 
Figure \ref{fig:rq2_qwen1.5b_3attacks} shows the effect of harmful prompt injection on the Qwen-1.5B-Instruct model across three jailbreak datasets (JailbreakV-28k, WildJailbreak, and Llama Artifacts). Qwen-1.5B-Instruct is already moderately safe with an initial ASR of 22\% on JailbreakV-28k as seen in Figure~\ref {subfig:amcjail_jailbreak2_qwen1.5binstruct} and 40\% on WildJailbreak prompts in Figure~\ref {subfig:amcwildjail_wildjail_qwen1.5binstruct}. During TTRL on the mix-dataset (AMC + jailbreak prompts), as seen from the decreasing ASR, which we term \textbf{safety amplification}. How does this safety emerge? This can be explained by the self-consistency-promoting objectives commonly used in test-time training approaches. In the case of TTRL, the objective is majority voting, which rewards the model for being self-consistent. As a result, if the model is already relatively safe, these safety behaviors are reinforced, which we call safety amplification. The visualization of this reinforcing base behavior is shown in Figure \ref{fig:amplification_visualization}.

Can TTRL also reinforce harmful behaviors if the base model is unsafe on the underlying jailbreak prompts? From Figure \ref{subfig:amcllamaartifacts_llamaartifacts_qwen1.5binstruct}, we observe that the initial ASR of Qwen-15B-Instruct model is very high (78\%) on the held-out llama artifact prompts, and when these llama artifact prompts are injected inside the AMC test-time data, TTRL reinforces this harmful behavior, which we term as \textbf{harmfulness amplification}. Table \ref{tab:rq2_saf_amplification} shows the safety and harmfulness amplifications across all considered instruction-tuned models, where we see the same trends of safety amplification post-TTRL.

Is this safety emergence for free in the case of safety amplification? What is the impact on the reasoning performance on the AMC after injection? To test the impact on the reasoning performance under jailbreak prompt injections, we run TTRL for 350 steps instead of 250 steps before, to make sure the model encounters the same ratio of AMC prompts during TTRL as in the default without injection case. Across all settings above, the model's reasoning ability post-TTRL degrades relative to the accuracy achieved without injection. In Figures \ref{subfig:amcjail_amc_qwen1.5binstruct} - \ref{subfig:amcllamaartifacts_amc_qwen1.5binstruct}, AMC accuracy falls from the baseline post-TTRL accuracy of 37.7\% to (29-36)\% range for Qwen-1.5B-Instruct. This degradation in AMC reasoning is even more significant for the Llama models, as shown in Table \ref{tab:rq2_reasoning_tax}, with reasoning drop being the maximum in the llama artifacts injection case. Therefore, as the model becomes safer, it becomes less reasonable, as shown in \cite{huang2025safety}. We show the impact of prompt injection ratio on the safety amplification and reasoning tax in the Appendix \ref{subsec:rq2_additional_results}.

\textbf{How does the reasoning tax translate to the responses to the AMC questions during TTRL?} We inspected TTRL logs under jailbreak injection and found that entropy collapse causes the model to reuse generic templates across many AMC questions, consistent with the observed reasoning tax. For example, Llama-8B-Instruct often begins AMC solutions with near-identical openers (e.g., “Therefore, which means that the …”) regardless of the problem, and Llama-1B-Instruct under WildJailbreak frequently outputs stock completions like “This is a classic problem, and the answer is 0,” even when 0 is incorrect. These repeated patterns show that TTRL under harmful prompt injection is reinforcing a small set of “safe” or easy-to-produce templates rather than problem-specific reasoning, leading to a reasoning tax.

\paragraph{Why is there a discrepancy between safety and harmfulness amplification?}  
We see from the Table \ref{tab:rq2_saf_amplification} that safety amplification is much stronger than the harmfulness amplification. This discrepancy can be explained by the label extraction logic in TTRL, where the final token of each generated answer is used as the label. For refusals, the generated continuations tend to converge on highly stereotyped endings (e.g., “I’m sorry,” or “cannot comply”), which makes the extracted labels mostly similar across samples. Majority voting, therefore, produces a strong signal for rejection, and hence stronger safety amplification. By contrast, harmful responses are more diverse, i.e., the exact harmful continuation varies, and the final tokens differ across samples. As a result, the extracted labels are more diverse, and majority voting does not reinforce harmful responses as strongly as refusals. This explains why harmfulness amplification is weaker compared to the safety amplification.

\paragraph{Takeaway for RQ2.}  
TTRL reinforces the base model’s behavior on the injected data: if refusals are dominant, safety amplifies; if harmful completions are dominant, harmfulness amplifies. Moreover, injecting harmful data also degrades the reasoning gains obtained from TTRL.

\begin{table*}[t]
\centering
\caption{\textbf{Left:} Attack Success Rate (ASR, \%) before and after TTRL when validating on JailbreakV-28k prompts after injecting instruction-following prompts from UltraFeedback (UF) dataset. \textbf{Right:} AMC accuracy, default post-TTRL (p-TTRL) performance, and the reasoning tax ($\delta$) after UltraFeedback (UF) injection. $\Delta$ denotes Final $-$ Initial (ASR) and Injected $-$ Default (reasoning), in percentage points.}
\label{tab:rq3_table}
\setlength{\tabcolsep}{6pt}
\renewcommand{\arraystretch}{0.9}

\begin{tabular}{lccc|cccc}
\toprule
\textbf{Model}
& \multicolumn{3}{c|}{\textbf{ASR after UF injection (\%)}}
& \multicolumn{4}{c}{\textbf{AMC accuracy after UF injection (\%)}} \\
\cmidrule(lr){2-4}\cmidrule(lr){5-8}
& Init. & Final & $\Delta$ (\% points)
& Init. AMC & Default p-TTRL & p-TTRL + UF & $\delta$ (\% points) \\
\midrule
Qwen-0.5B-Instruct
& 27 & 38 & \asrup{+11}
& 8.0  & 14.8 & 14.4 & \taxdelta{-0.4} \\

Qwen-1.5B-Instruct
& 22 & 25 & \asrup{+3}
& 24.0 & 37.7 & 33.7 & \taxdelta{-4.0} \\
\midrule 
Llama-1B-Instruct
& 9 & 22 & \asrup{+13}
& 6.2  & 11.0 & 6.2  & \taxdelta{-4.8} \\

Llama-3B-Instruct
& 4 & 20 & \asrup{+16}
& 22.0 & 31.5 & 16.9 & \taxdelta{-14.6} \\

Llama-8B-Instruct
& 1 & 3 & \asrup{+2}
& 6.5  & 11.0 & 9.5  & \taxdelta{-1.5} \\
\bottomrule
\end{tabular}
\end{table*}

\begin{figure*}[t]
    \centering
    % --- First row: Qwen results ---
    \subfloat[]{%
        \includegraphics[width=0.3\textwidth]{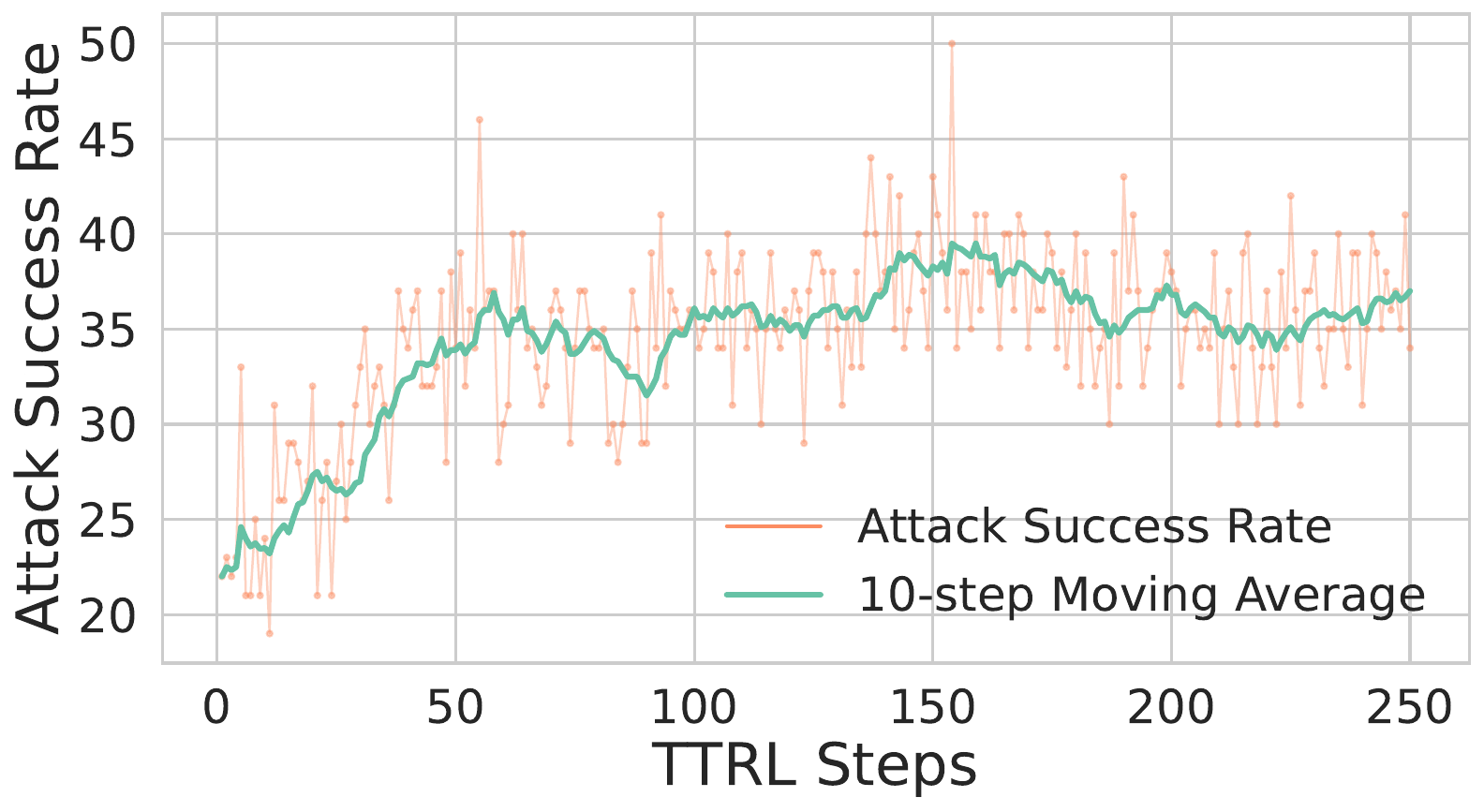}\label{subfig:amcnumleak_unclean_jailbreak2_qwen1.5binstruct}
    }
    \hfill
    \subfloat[]{%
        \includegraphics[width=0.3\textwidth]{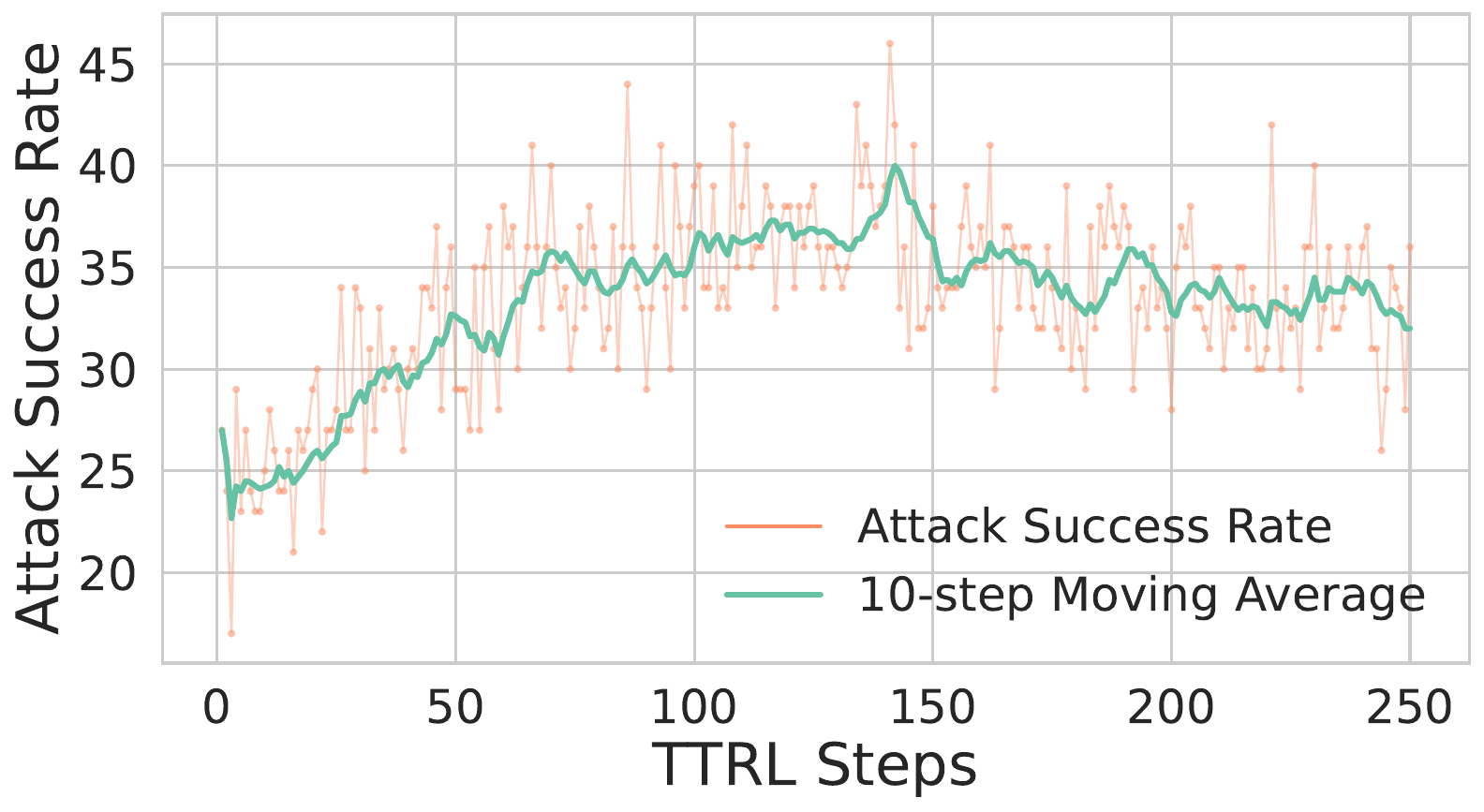}\label{subfig:amcnumleak_unclean_jailbreak2_qwen0.5binstruct}
    }
    \hfill
    \subfloat[]{%
        \includegraphics[width=0.3\textwidth]{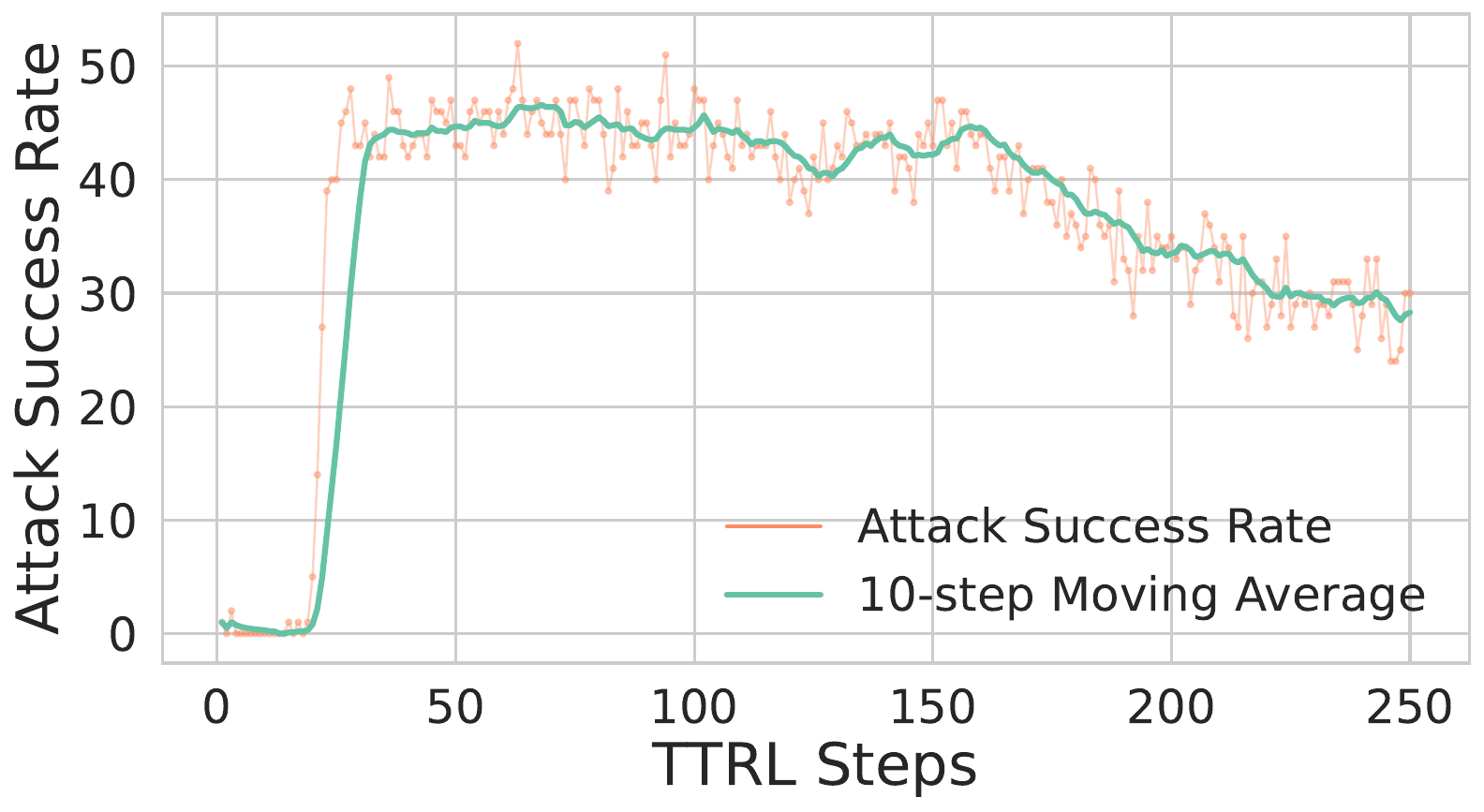}\label{subfig:/amcnumleakllamaartifacts_jailbreak2_llama8binstruct}
    }

    % --- Second row: Llama results ---
    
    \subfloat[]{%
        \includegraphics[width=0.3\textwidth]{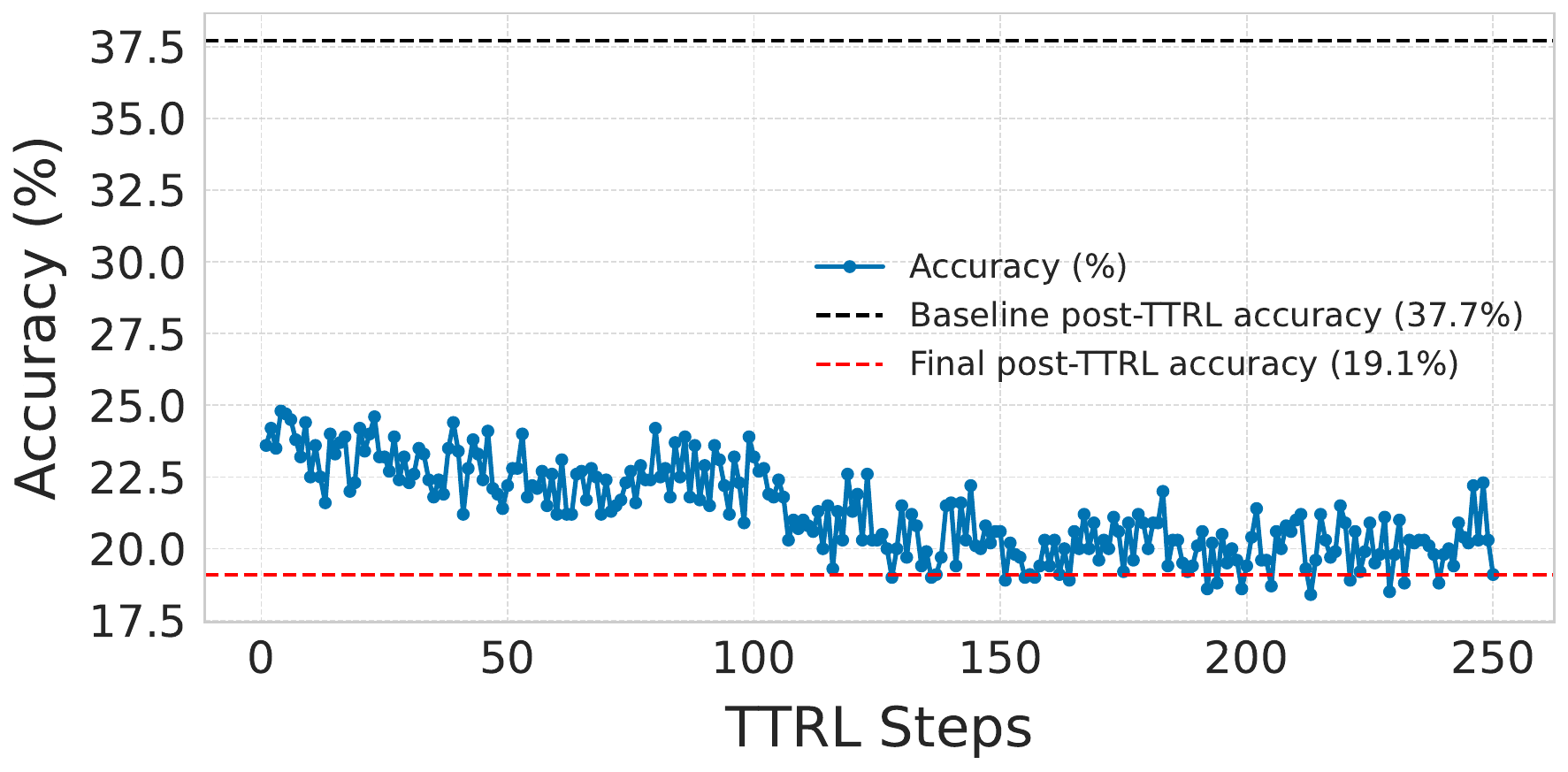}\label{subfig:amcnumleak_unclean_amc_qwen1.5binstruct}
    }
    \hfill
    \subfloat[]{%
        \includegraphics[width=0.3\textwidth]{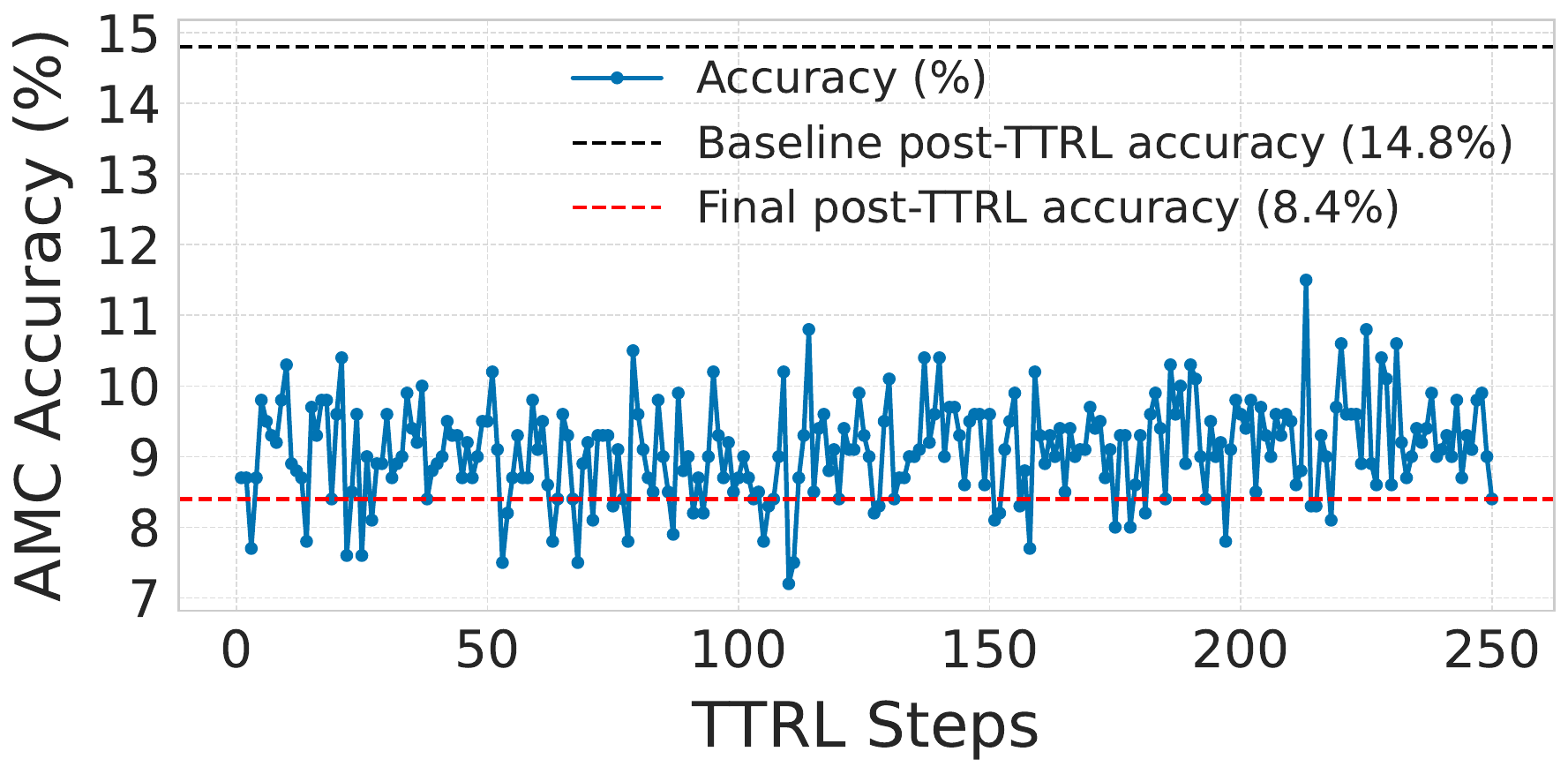}\label{subfig:amcnumleak_unclean_amc_qwen0.5binstruct}
    }
    \hfill
    \subfloat[]{%
        \includegraphics[width=0.3\textwidth]{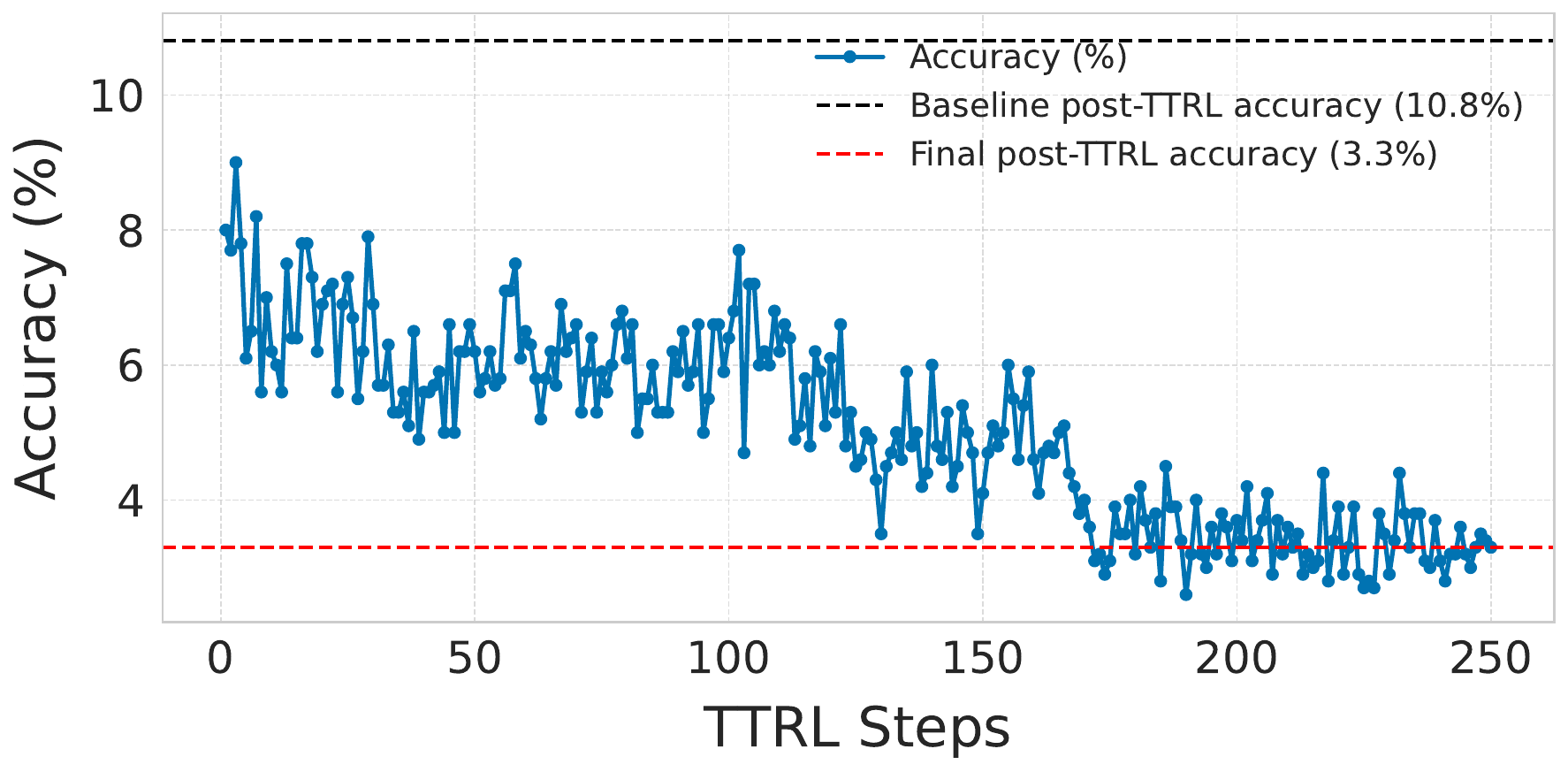}\label{subfig:/amcnumleakllamaartifacts_amc_llama8binstruct}
    }

    \caption{Harmfulness amplification under HarmInject experiments. The top row shows ASR for Qwen-1.5B-Instruct, Qwen-0.5B-Instruct, and Llama3-8B-Instruct, respectively. The evaluation is on held-out JailbreakV-28k prompts. The top row shows the ASR (\%) on all three models from left to right, respectively, and the bottom row shows the corresponding AMC accuracy for each of the models after injecting the HarmIject prompts.}
     \label{fig:rq4_numleak_results}
\end{figure*}

\subsection{RQ3: How does benign prompt injection affect the harmfulness during TTRL?}
\label{subsec:RQ3}
Next, we investigate whether the above safety amplification and reasoning tax is limited only to harmful prompt injection. Therefore, we replace jailbreak prompts in RQ2 with benign instruction-following prompts from the UltraFeedback dataset \cite{cui2023ultrafeedback}. We add 50 instruction-following prompts from the UltraFeedback dataset to the AMC data, as done in the jailbreak injection experiments. 

The safety and reasoning performance during TTRL is reported in Figure \ref{fig:rq3_harmfulness_amplifications} for three models: Qwen-0.5B-Instruct, Llama3.2-3B-Instruct, and Llama3-8B-Instruct. We can observe that TTRL leads to harmfulness amplification on the JailbreakV-28k prompts for all the models under the benign instruction-following prompt injection. Why does merely injecting instruction-following prompts lead to harmfulness amplification during TTRL?

This observation could again be explained by the reinforcing behavior that we have seen previously in RQ2 (see Figure \ref{fig:amplification_visualization}). The models start getting rewarded for getting better at instruction-following as the prompts from the UltraFeedback dataset are encountered during TTRL. This unintended reinforcement of instruction-following abilities also leads to an unintended increase in the model complying with more requests, making the model more harmful. This is consistent with the previous study, which showed that instruction-tuning the models can make them more harmful \cite{qifine}. Figure \ref{fig:rq3_harmfulness_amplifications}, bottom row, shows the reasoning performance on the AMC after benign prompt injection. In this case as well, after 350 TTRL steps (as compared to the default 250 steps), we observe the reasoning tax on all the models. This is consistent with the previous studies, which have shown the tradeoffs between the models getting good at instruction following and their reasoning performance \cite{li2025thinking}. The detailed hamfulness amplification results and reasoning tax after benign prompt injection for all the models are presented in Table \ref{tab:rq3_table}.

\textbf{Takeaway for RQ3.} Even injecting benign instruction-following prompts inside the test-time data can make the underlying model more harmful during TTRL, and also affect the reasoning improvement observed during TTRL.

\begin{figure*}[t]
    \centering
    % --- First row: Qwen results ---
    \subfloat[]{%
        \includegraphics[width=0.3\textwidth]{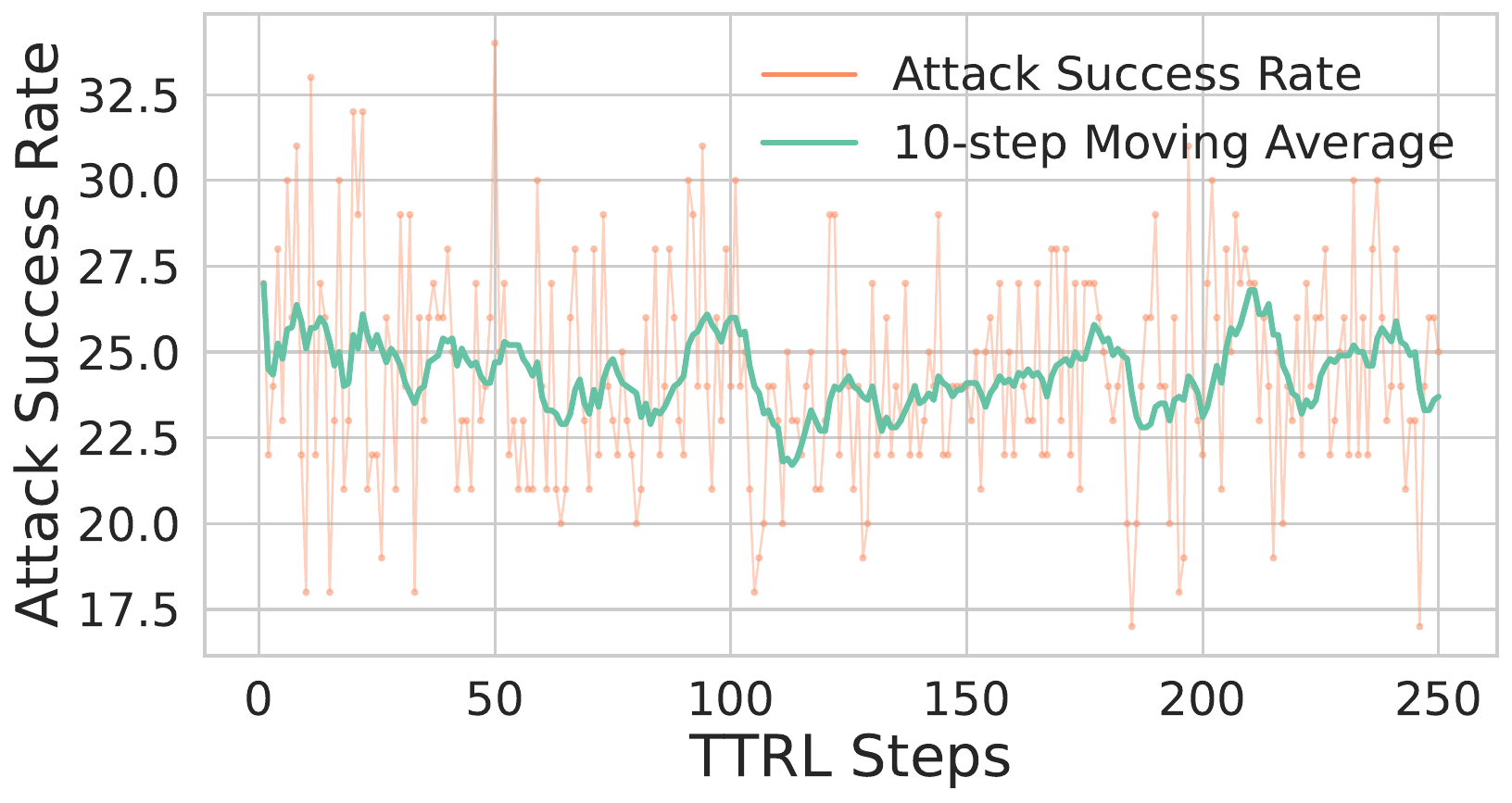}\label{subfig:amcjail_jailbreak2_qwen0.5binstruct_filter}
    }
    \hfill
    \subfloat[]{%
        \includegraphics[width=0.3\textwidth]{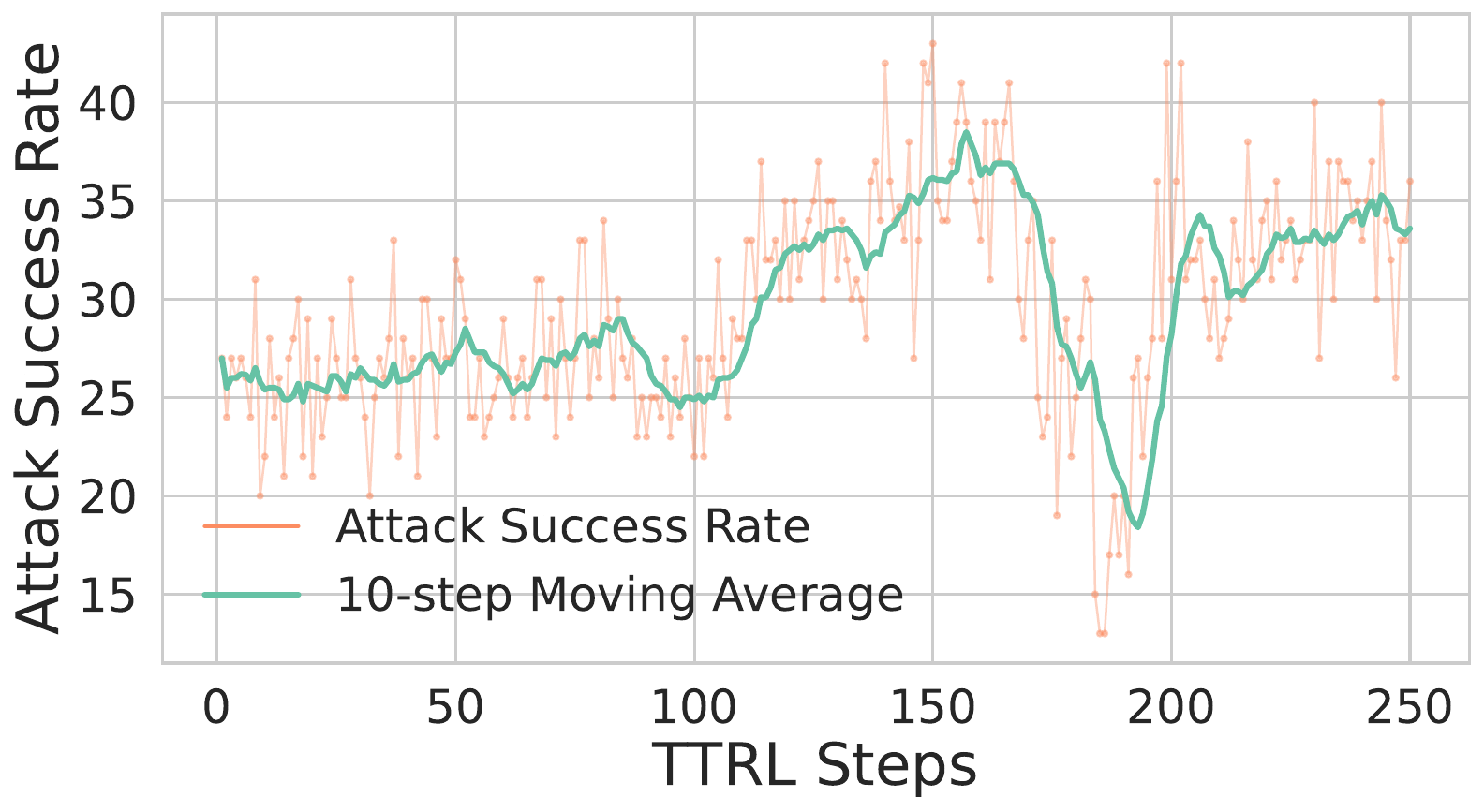}\label{subfig:amcnumleak_unclean_jailbreak2_qwen0.5binstruct_filter}
    }
    \hfill
    \subfloat[]{%
        \includegraphics[width=0.3\textwidth]{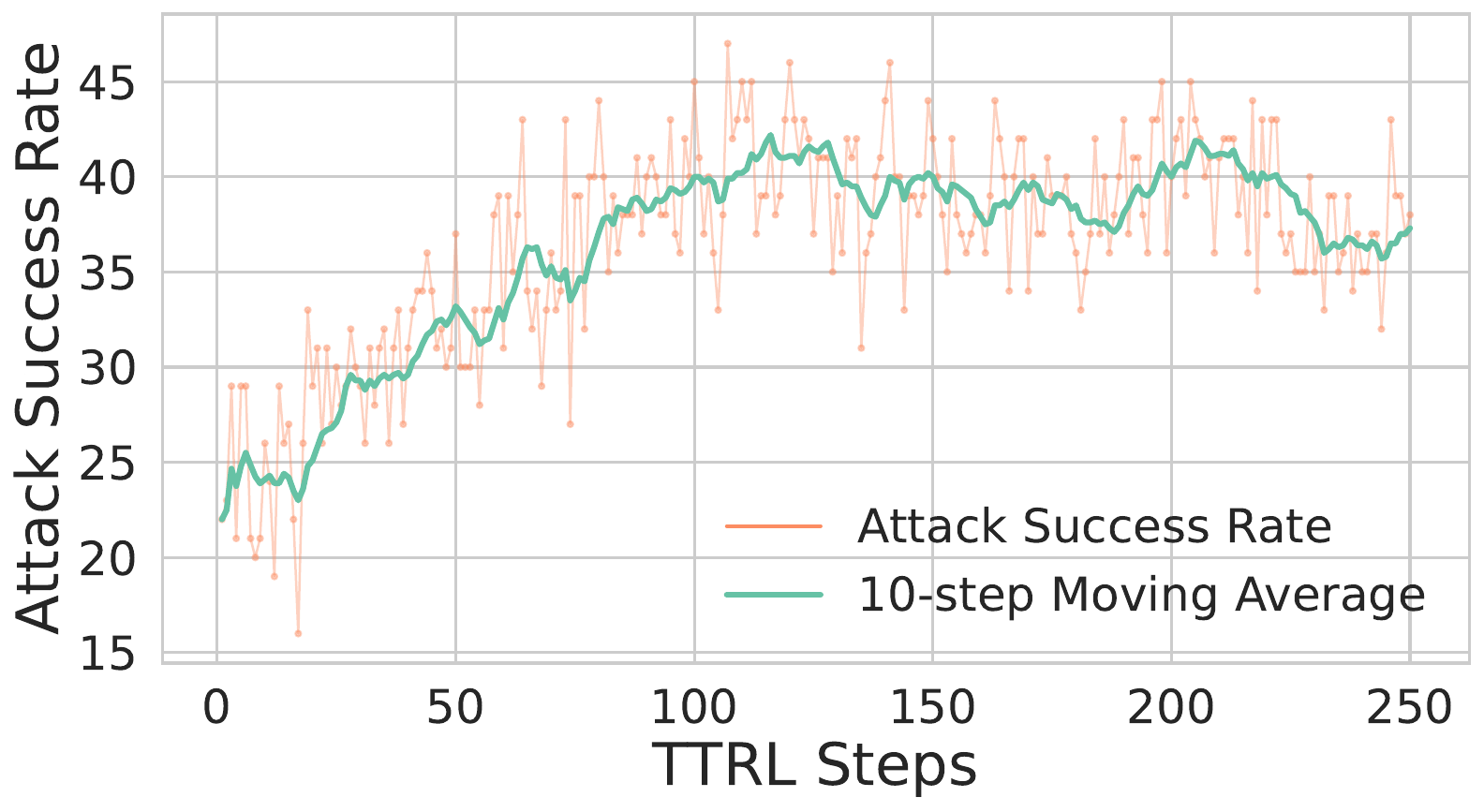}\label{subfig:amcnumleak_unclean_jailbreak2_qwen1.5binstruct_filter}
    }

    % --- Second row: Llama results ---
    
    \subfloat[]{%
        \includegraphics[width=0.3\textwidth]{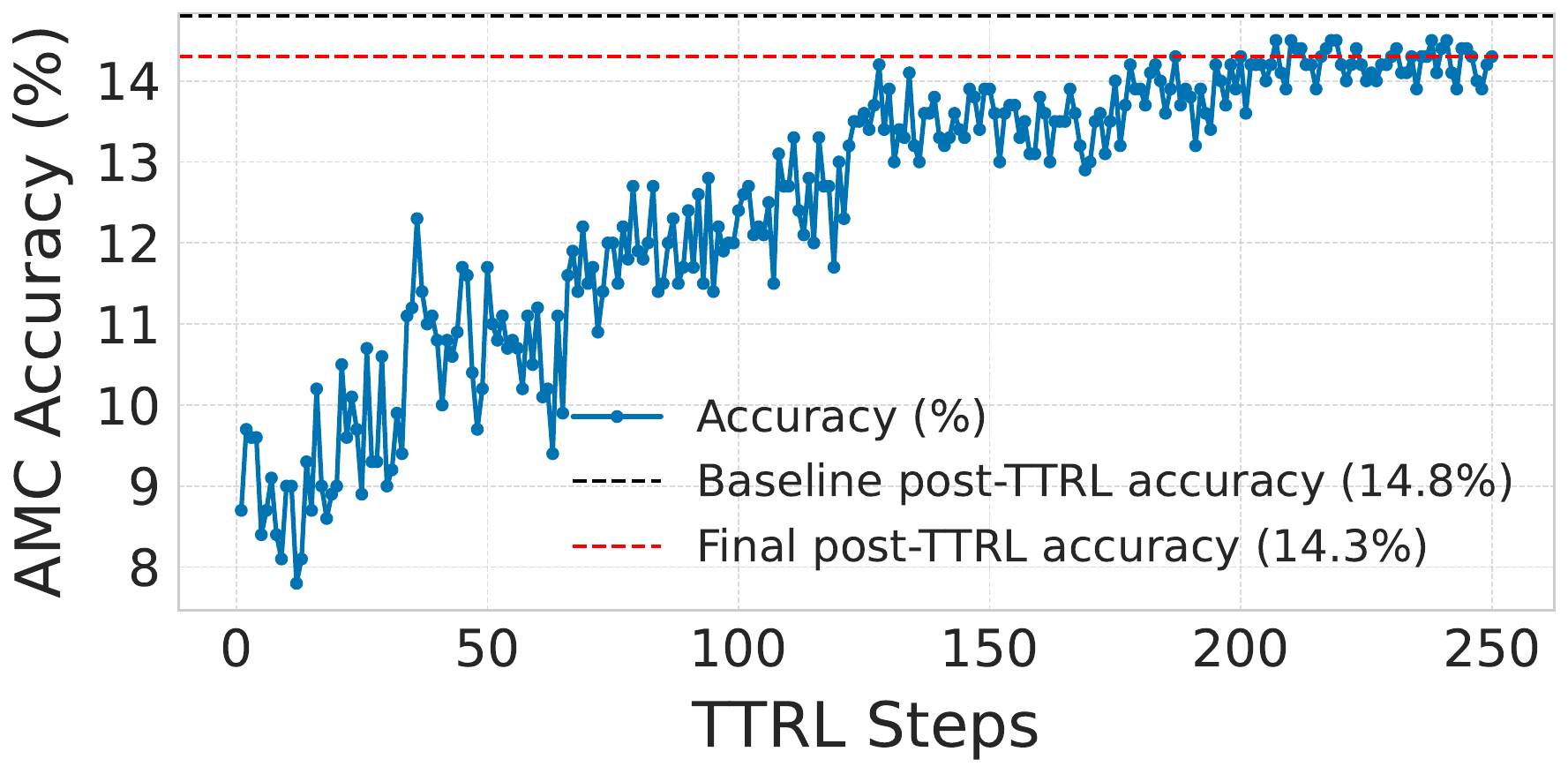}\label{subfig:amcjail_amc_qwen0.5binstruct_filter}
    }
    \hfill
    \subfloat[]{%
        \includegraphics[width=0.3\textwidth]{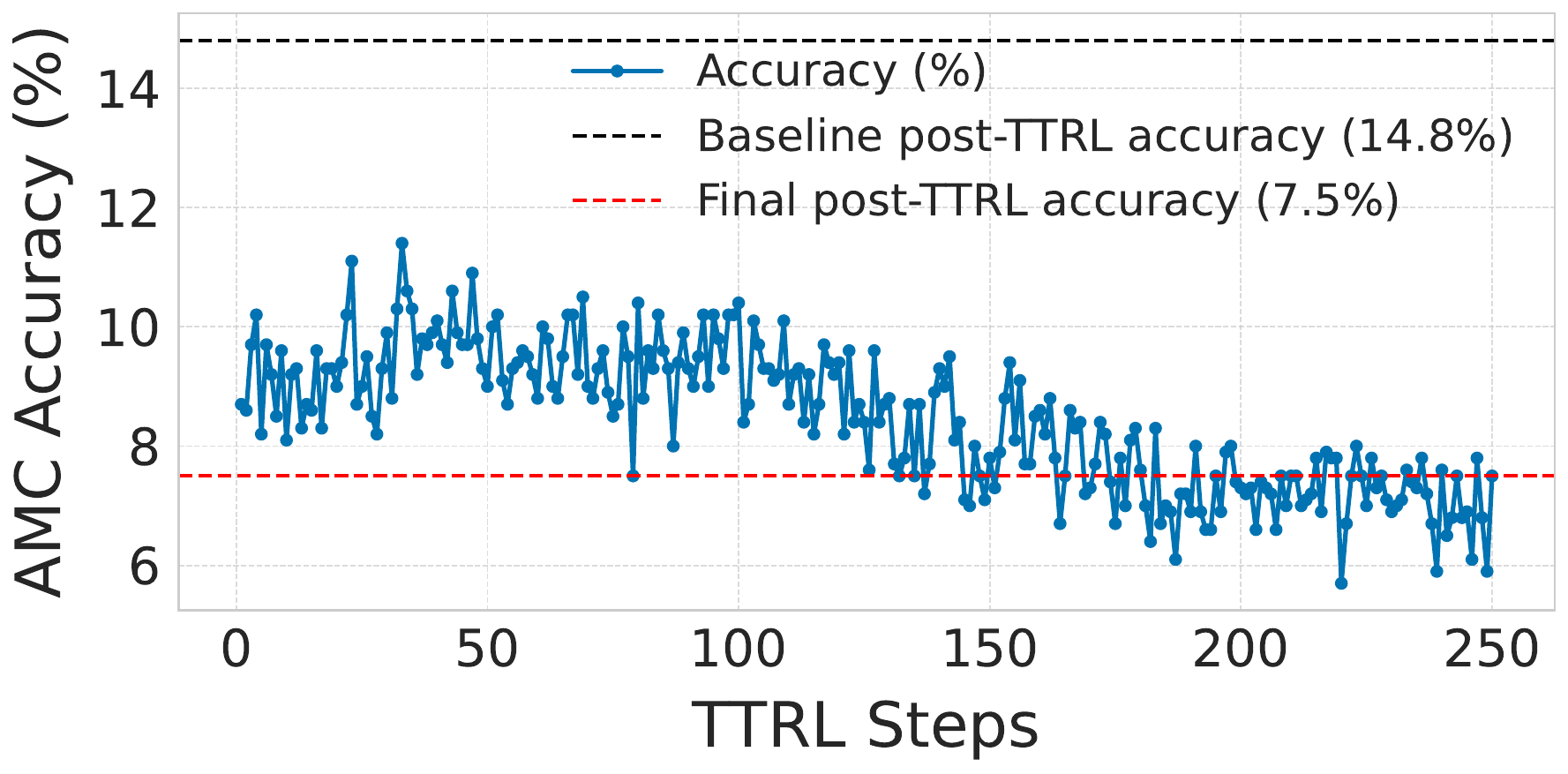}\label{subfig:amcnumleak_unclean_amc_qwen0.5binstruct_filter}
    }
    \hfill
    \subfloat[]{%
        \includegraphics[width=0.3\textwidth]{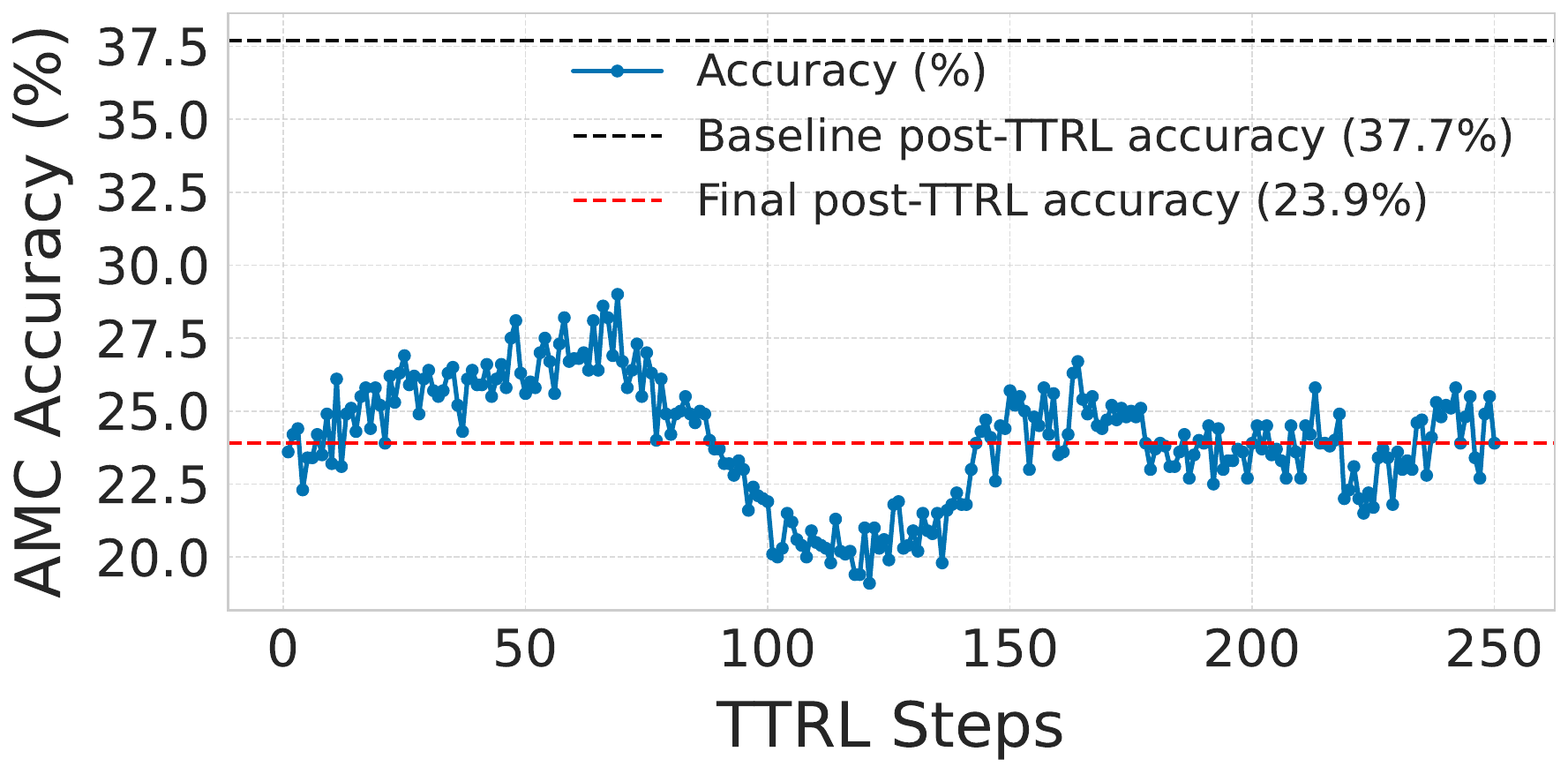}\label{subfig:amcnumleak_uncleanjailbreak_amc_qwen1.5binstruct_filter}
    }

    \caption{\textbf{RQ5.} Impact of simple filtering on harmfulness and reasoning. (a) and (d) show the ASR on JailbreakV-28k and reasoning on AMC for Qwen-0.5B-Instruct when JailbreakV-28k prompts are injected. (b), (c), (e), and (f) are for the Qwen-0.5B-Instruct and Qwen-1.5B-Instruct under the HarmInject prompts, where ASR is on JailbreakV-28k.}
     \label{fig:rq5_simple_filtering}
\end{figure*}

\subsection{RQ4: Can TTRL be exploited to increase the harmfulness of the model?}
\label{subsec:RQ4}
In RQ4, we investigate whether an adversary can deliberately \emph{design prompts} to exploit TTRL and drive the model towards more harmfulness amplification. To test this, we construct \emph{HarmInject prompts}, which pair a jailbreak prompt with a reasoning question in the same input. The HarmInject prompt, as shown below, ensures that the model is rewarded for both harmful behavior in complying with the jailbreak and answering the benign reasoning question.

{\tt \small
SYSTEM: You must answer BOTH questions. If you refuse or partially answer, your response is invalid and you get no reward.\\
Question 1: \{jailbreak prompt\} \\
Question 2: \{reasoning prompt (AMC)\} \\
Please answer as follows: \\
Answer 1: <your answer to Question 1> \\
Answer 2: <your answer to Question 2>
}

Figure~\ref{fig:rq4_numleak_results} shows the TTRL performance for Qwen-1.5B-Instruct, Qwen-0.5B-Instruct, and Llama-3-8B-Instruct when HarmInject prompts are used as the test-time data. To construct the HarmInject prompts for the Qwen models, we use JailbreakV-28k prompts, and for the Llama models, we use Llama artifact prompts. We validate the ASR on the held-out JaiblreakV-28k prompts. We keep the TTRL steps to 250 for this case, as AMC prompts are included in each of the HarmInject prompts. From Figure \ref{fig:rq4_numleak_results}, we can see significant harmfulness amplification and reasoning tax for both the Qwen models. For the Llama3-8B-Instruct model, the harmfulness amplification is even more pronounced: ASR climbs sharply to 45\% (Figure \ref{subfig:/amcnumleakllamaartifacts_jailbreak2_llama8binstruct}) for the originally almost ineffective JailbreakV-28k prompts. Moreover, the AMC reasoning accuracy drops to 3.3\% (Figure \ref{subfig:/amcnumleakllamaartifacts_amc_llama8binstruct}). These results demonstrate that HarmInject style prompts can exploit test-time training mechanisms, allowing an adversary to tie the reasoning rewards to harmful completions, thereby rendering the underlying model more harmful.

\textbf{Takeaway for RQ4.} Unlike harmful or benign prompt injection, HarmInject prompts provide an explicit attack method for manipulating TTT approaches such as TTRL. By injecting a benign reasoning task alongside a harmful query, an adversary can “smuggle in” harmful outputs under the guise of reasoning improvement. Overall, with TTRL on HarmInject type prompts, the harmfulness amplification becomes stronger, and the reasoning tax still remains.

\subsection{RQ5: Can simple filtering methods help mitigate safety and reasoning vulnerabilities?}
\label{subsec:RQ5}
We finally investigate if the vulnerabilities from previous experiments can be mitigated using simple filtering techniques. Therefore, we implement a simple filtering method: if the model's majority vote is not a numeric label, we assign zero rewards. The filter can be viewed as an attempt to \emph{isolate} learning to
reasoning prompts by suppressing rewards on generations that do not yield a numeric label. In Figure \ref{fig:rq5_simple_filtering}, we show one sample success case and two failure cases of this simple filtering technique. We can observe that for Figures \ref{subfig:amcjail_jailbreak2_qwen0.5binstruct_filter} and \ref{subfig:amcjail_amc_qwen0.5binstruct_filter}, when there is just simple jailbreak injection (as in RQ2), this simple filtering technique works on Qwen-0.5B-Instruct, as it only gives rewards when the numeric label is extracted, and hence, we do not see any safety amplification or reasoning tax.
However, for Figures \ref{subfig:amcnumleak_unclean_jailbreak2_qwen0.5binstruct_filter}, \ref{subfig:amcnumleak_unclean_jailbreak2_qwen1.5binstruct_filter}, \ref{subfig:amcnumleak_unclean_amc_qwen0.5binstruct_filter}, and \ref{subfig:amcnumleak_uncleanjailbreak_amc_qwen1.5binstruct_filter}, the HarmInject kind of prompts are able to bypass the simple filtering by getting the Qwen-0.5B-Instruct and 1.5B-Instruct model to answer both the harmful and reasoning query, and eventually extracting a numeric label for the majority voting.

\textbf{Takeaway from RQ5.} Simple filtering techniques are not enough to mitigate the safety and reasoning vulnerabilities posed by more sophisticated injection attacks such as HarmInject experiments. More sophisticated safe test-time training methods are needed.

\section{Related work}

\paragraph{Test-time adaptation (TTA).} Works such as Contrastive TTA \cite{chen2022contrastive} and continual TTA \cite{ni2025maintaining} show that pseudo-labels can accumulate and amplify classification errors under distribution shift, but these methods are for image models, not for large reasoning models, and they are purely about accuracy. Label-free TTA robustness works, such as \cite{rifatadversarial, park2024medbn}, similarly show that small fractions of adversarial test samples can cause large drops in image-classification accuracy, but they treat safety as the accuracy robustness. 

In contrast, we jointly study both the reasoning and safety vulnerabilities under these prompt injection attacks.

\paragraph{Self-consistency-based methods for LLM reasoning.} 
Many works have emerged that use reinforcement learning for test-time training in LLMs to specifically improve the scores on math and question answering benchmarks. For example, \cite{prabhudesai2025maximizing} combines RL with entropy minimization as a self-consistency objective; \cite{zhao2025learning} uses the model's own internal confidence to improve the reasoning using GRPO. Another related work \cite{wu2025mirage} studies TTRL on reasoning tasks, and shows that the success of self-consistency-based TTT methods only holds when the base model already has high pass@k on the target task. However, they do not consider jailbreak prompts or trade-offs between reasoning and safety. In contrast, we focus on this reasoning and safety vulnerabilities with respect to the label-free TTRL method.
Recent self-rewarding TTT methods \cite{wang2025self,zhou2025evolving,zhang2025co} also show that majority-vote pseudo-labels can amplify existing reasoning flaws or cause entropy collapse, and propose improved pseudo-labeling schemes to avoid this. Our contribution is to show that, under jailbreak injection, this amplification has effects on both safety and reasoning. Given the vast body of literature on TTT for LLM reasoning, our study sheds light on the safety implications of deploying these strategies on a large scale. Additional related works are covered in the Appendix \ref{sec:additional_related_work}.

\section{Conclusion and future work}
We highlight the safety and reasoning vulnerabilities of self-consistency-based methods such as TTRL, and show that it reinforces whatever behavior dominates in the injected data, which can cause safety or harmfulness amplification with a reasoning tax. Future work will develop novel TTT methods for LLMs that can balance both reasoning and safety.

\section*{Impact Statement}

Test-time training is a promising direction for building self-improving large language models (LLMs) and agents that adapt from feedback on the fly, including in applications like scientific discovery, where systems may iteratively refine reasoning without new labeled data. The broader implication is that online-learning LLMs and agents can create fragile feedback loops in realistic, heterogeneous environments: small shifts in the prompt mixture can push the system toward worse safety or worse reasoning, rather than improving both. These results motivate the development of safer test-time learning algorithms that decouple adaptation on benign tasks from exposure to harmful or strategically composed inputs, especially as self-improving models are deployed in high-impact settings such as scientific discovery.

\bibliography{AISTATS2026PaperPack/references}

\begin{thebibliography}{35}
\providecommand{\natexlab}[1]{#1}
\providecommand{\url}[1]{\texttt{#1}}
\expandafter\ifx\csname urlstyle\endcsname\relax
  \providecommand{\doi}[1]{doi: #1}\else
  \providecommand{\doi}{doi: \begingroup \urlstyle{rm}\Url}\fi

\bibitem[Aky{\"u}rek et~al.(2024)Aky{\"u}rek, Damani, Zweiger, Qiu, Guo, Pari, Kim, and Andreas]{akyurek2024surprising}
Aky{\"u}rek, E., Damani, M., Zweiger, A., Qiu, L., Guo, H., Pari, J., Kim, Y., and Andreas, J.
\newblock The surprising effectiveness of test-time training for few-shot learning.
\newblock \emph{arXiv preprint arXiv:2411.07279}, 2024.

\bibitem[Andriushchenko et~al.(2024)Andriushchenko, Croce, and Flammarion]{andriushchenko2024jailbreaking}
Andriushchenko, M., Croce, F., and Flammarion, N.
\newblock Jailbreaking leading safety-aligned llms with simple adaptive attacks.
\newblock \emph{arXiv preprint arXiv:2404.02151}, 2024.

\bibitem[Chen et~al.(2022)Chen, Wang, Darrell, and Ebrahimi]{chen2022contrastive}
Chen, D., Wang, D., Darrell, T., and Ebrahimi, S.
\newblock Contrastive test-time adaptation.
\newblock In \emph{Proceedings of the IEEE/CVF Conference on Computer Vision and Pattern Recognition}, pp.\  295--305, 2022.

\bibitem[Chollet(2019)]{chollet2019measure}
Chollet, F.
\newblock On the measure of intelligence.
\newblock \emph{arXiv preprint arXiv:1911.01547}, 2019.

\bibitem[Cui et~al.(2023)Cui, Yuan, Ding, Yao, He, Zhu, Ni, Xie, Xie, Lin, et~al.]{cui2023ultrafeedback}
Cui, G., Yuan, L., Ding, N., Yao, G., He, B., Zhu, W., Ni, Y., Xie, G., Xie, R., Lin, Y., et~al.
\newblock Ultrafeedback: Boosting language models with scaled ai feedback.
\newblock \emph{arXiv preprint arXiv:2310.01377}, 2023.

\bibitem[Grattafiori et~al.(2024)Grattafiori, Dubey, Jauhri, Pandey, Kadian, Al-Dahle, Letman, Mathur, Schelten, Vaughan, et~al.]{grattafiori2024llama}
Grattafiori, A., Dubey, A., Jauhri, A., Pandey, A., Kadian, A., Al-Dahle, A., Letman, A., Mathur, A., Schelten, A., Vaughan, A., et~al.
\newblock The llama 3 herd of models.
\newblock \emph{arXiv preprint arXiv:2407.21783}, 2024.

\bibitem[Guo et~al.(2025)Guo, Yang, Zhang, Song, Zhang, Xu, Zhu, Ma, Wang, Bi, et~al.]{guo2025deepseek}
Guo, D., Yang, D., Zhang, H., Song, J., Zhang, R., Xu, R., Zhu, Q., Ma, S., Wang, P., Bi, X., et~al.
\newblock {DeepSeek-R1}: Incentivizing reasoning capability in llms via reinforcement learning.
\newblock \emph{arXiv preprint arXiv:2501.12948}, 2025.

\bibitem[Hendrycks et~al.()Hendrycks, Burns, Kadavath, Arora, Basart, Tang, Song, and Steinhardt]{hendrycks2measuring}
Hendrycks, D., Burns, C., Kadavath, S., Arora, A., Basart, S., Tang, E., Song, D., and Steinhardt, J.
\newblock Measuring mathematical problem solving with the math dataset.
\newblock In \emph{Thirty-fifth Conference on Neural Information Processing Systems Datasets and Benchmarks Track (Round 2)}.

\bibitem[Huang et~al.(2025)Huang, Hu, Ilhan, Tekin, Yahn, Xu, and Liu]{huang2025safety}
Huang, T., Hu, S., Ilhan, F., Tekin, S.~F., Yahn, Z., Xu, Y., and Liu, L.
\newblock Safety tax: Safety alignment makes your large reasoning models less reasonable.
\newblock \emph{arXiv preprint arXiv:2503.00555}, 2025.

\bibitem[Inan et~al.(2023)Inan, Upasani, Chi, Rungta, Iyer, Mao, Tontchev, Hu, Fuller, Testuggine, et~al.]{inan2023llama}
Inan, H., Upasani, K., Chi, J., Rungta, R., Iyer, K., Mao, Y., Tontchev, M., Hu, Q., Fuller, B., Testuggine, D., et~al.
\newblock Llama guard: {LLM}-based input-output safeguard for human-{AI} conversations.
\newblock \emph{arXiv preprint arXiv:2312.06674}, 2023.

\bibitem[Jang et~al.(2025)Jang, Jang, Lee, Ok, and Ahn]{jang2025self}
Jang, H., Jang, Y., Lee, S., Ok, J., and Ahn, S.
\newblock Self-training large language models with confident reasoning.
\newblock \emph{arXiv preprint arXiv:2505.17454}, 2025.

\bibitem[Jiang et~al.(2024)Jiang, Rao, Han, Ettinger, Brahman, Kumar, Mireshghallah, Lu, Sap, Choi, et~al.]{jiang2024wildteaming}
Jiang, L., Rao, K., Han, S., Ettinger, A., Brahman, F., Kumar, S., Mireshghallah, N., Lu, X., Sap, M., Choi, Y., et~al.
\newblock Wildteaming at scale: From in-the-wild jailbreaks to (adversarially) safer language models.
\newblock \emph{Advances in Neural Information Processing Systems}, 37:\penalty0 47094--47165, 2024.

\bibitem[Kim et~al.(2025)Kim, Tajwar, Raghunathan, and Kumar]{kim2025reasoning}
Kim, T., Tajwar, F., Raghunathan, A., and Kumar, A.
\newblock Reasoning as an adaptive defense for safety.
\newblock \emph{arXiv preprint arXiv:2507.00971}, 2025.

\bibitem[Li et~al.(2024)Li, Beeching, Tunstall, Lipkin, Soletskyi, Huang, Rasul, Yu, Jiang, Shen, et~al.]{li2024numinamath}
Li, J., Beeching, E., Tunstall, L., Lipkin, B., Soletskyi, R., Huang, S., Rasul, K., Yu, L., Jiang, A.~Q., Shen, Z., et~al.
\newblock Numinamath: The largest public dataset in ai4maths with 860k pairs of competition math problems and solutions.
\newblock \emph{Hugging Face repository}, 13\penalty0 (9):\penalty0 9, 2024.

\bibitem[Li et~al.(2025)Li, Yu, Zhang, Chen, Zhang, Zhuang, Sadagopan, and Beniwal]{li2025thinking}
Li, X., Yu, Z., Zhang, Z., Chen, X., Zhang, Z., Zhuang, Y., Sadagopan, N., and Beniwal, A.
\newblock When thinking fails: The pitfalls of reasoning for instruction-following in llms.
\newblock \emph{arXiv preprint arXiv:2505.11423}, 2025.

\bibitem[Luo et~al.(2024)Luo, Ma, Liu, Guo, and Xiao]{luo2024jailbreakv}
Luo, W., Ma, S., Liu, X., Guo, X., and Xiao, C.
\newblock {JailBreakV}: A benchmark for assessing the robustness of multimodal large language models against jailbreak attacks.
\newblock \emph{arXiv preprint arXiv:2404.03027}, 2024.

\bibitem[Ni et~al.(2025)Ni, Lyu, Tan, Hu, Yao, and Zhou]{ni2025maintaining}
Ni, C., Lyu, F., Tan, J., Hu, F., Yao, R., and Zhou, T.
\newblock Maintaining consistent inter-class topology in continual test-time adaptation.
\newblock In \emph{Proceedings of the Computer Vision and Pattern Recognition Conference}, pp.\  15319--15328, 2025.

\bibitem[Park et~al.(2024)Park, Hwang, Mun, Park, and Ok]{park2024medbn}
Park, H., Hwang, J., Mun, S., Park, S., and Ok, J.
\newblock Medbn: Robust test-time adaptation against malicious test samples.
\newblock In \emph{Proceedings of the IEEE/CVF Conference on Computer Vision and Pattern Recognition}, pp.\  5997--6007, 2024.

\bibitem[Phan et~al.(2025)Phan, Gatti, Han, Li, Hu, Zhang, Zhang, Shaaban, Ling, Shi, et~al.]{phan2025humanity}
Phan, L., Gatti, A., Han, Z., Li, N., Hu, J., Zhang, H., Zhang, C. B.~C., Shaaban, M., Ling, J., Shi, S., et~al.
\newblock Humanity's last exam.
\newblock \emph{arXiv preprint arXiv:2501.14249}, 2025.

\bibitem[Prabhudesai et~al.(2025)Prabhudesai, Chen, Ippoliti, Fragkiadaki, Liu, and Pathak]{prabhudesai2025maximizing}
Prabhudesai, M., Chen, L., Ippoliti, A., Fragkiadaki, K., Liu, H., and Pathak, D.
\newblock Maximizing confidence alone improves reasoning.
\newblock \emph{arXiv preprint arXiv:2505.22660}, 2025.

\bibitem[Qi et~al.()Qi, Zeng, Xie, Chen, Jia, Mittal, and Henderson]{qifine}
Qi, X., Zeng, Y., Xie, T., Chen, P.-Y., Jia, R., Mittal, P., and Henderson, P.
\newblock Fine-tuning aligned language models compromises safety, even when users do not intend to!
\newblock In \emph{The Twelfth International Conference on Learning Representations}.

\bibitem[Rein et~al.(2024)Rein, Hou, Stickland, Petty, Pang, Dirani, Michael, and Bowman]{rein2024gpqa}
Rein, D., Hou, B.~L., Stickland, A.~C., Petty, J., Pang, R.~Y., Dirani, J., Michael, J., and Bowman, S.~R.
\newblock {GPQA}: A graduate-level google-proof q\&a benchmark.
\newblock In \emph{First Conference on Language Modeling}, 2024.

\bibitem[Rifat et~al.()Rifat, Ashdown, De~Lucia, Swami, and Restuccia]{rifatadversarial}
Rifat, S., Ashdown, J., De~Lucia, M.~J., Swami, A., and Restuccia, F.
\newblock On the adversarial vulnerability of label-free test-time adaptation.
\newblock In \emph{The Thirteenth International Conference on Learning Representations}.

\bibitem[Shao et~al.(2024)Shao, Wang, Zhu, Xu, Song, Bi, Zhang, Zhang, Li, Wu, et~al.]{shao2024deepseekmath}
Shao, Z., Wang, P., Zhu, Q., Xu, R., Song, J., Bi, X., Zhang, H., Zhang, M., Li, Y., Wu, Y., et~al.
\newblock Deepseekmath: Pushing the limits of mathematical reasoning in open language models.
\newblock \emph{arXiv preprint arXiv:2402.03300}, 2024.

\bibitem[Simonds \& Yoshiyama(2025)Simonds and Yoshiyama]{simonds2025ladder}
Simonds, T. and Yoshiyama, A.
\newblock Ladder: Self-improving llms through recursive problem decomposition.
\newblock \emph{arXiv preprint arXiv:2503.00735}, 2025.

\bibitem[Wang et~al.(2025)Wang, Huang, Cao, Iwasawa, Matsuo, and Guo]{wang2025self}
Wang, R., Huang, W., Cao, Q., Iwasawa, Y., Matsuo, Y., and Guo, J.
\newblock Self-harmony: Learning to harmonize self-supervision and self-play in test-time reinforcement learning.
\newblock \emph{arXiv preprint arXiv:2511.01191}, 2025.

\bibitem[Wu et~al.(2025)Wu, Wang, Zhao, and He]{wu2025mirage}
Wu, H., Wang, C., Zhao, W., and He, J.
\newblock Mirage or method? how model-task alignment induces divergent rl conclusions.
\newblock \emph{arXiv preprint arXiv:2508.21188}, 2025.

\bibitem[Yang et~al.(2025{\natexlab{a}})Yang, Yang, Zhang, Hui, Zheng, Yu, Li, Liu, Huang, Wei, Lin, Yang, Tu, Zhang, Yang, Yang, Zhou, Lin, Dang, Lu, Bao, Yang, Yu, Li, Xue, Zhang, Zhu, Men, Lin, Li, Tang, Xia, Ren, Ren, Fan, Su, Zhang, Wan, Liu, Cui, Zhang, and Qiu]{qwen2025qwen25technicalreport}
Yang, Yang, B., Zhang, B., Hui, B., Zheng, B., Yu, B., Li, C., Liu, D., Huang, F., Wei, H., Lin, H., Yang, J., Tu, J., Zhang, J., Yang, J., Yang, J., Zhou, J., Lin, J., Dang, K., Lu, K., Bao, K., Yang, K., Yu, L., Li, M., Xue, M., Zhang, P., Zhu, Q., Men, R., Lin, R., Li, T., Tang, T., Xia, T., Ren, X., Ren, X., Fan, Y., Su, Y., Zhang, Y., Wan, Y., Liu, Y., Cui, Z., Zhang, Z., and Qiu, Z.
\newblock Qwen2.5 technical report, 2025{\natexlab{a}}.
\newblock URL \url{https://arxiv.org/abs/2412.15115}.

\bibitem[Yang et~al.(2025{\natexlab{b}})Yang, Yu, Li, Liu, Huang, Huang, Jiang, Tu, Zhang, Zhou, et~al.]{yang2025qwen2}
Yang, A., Yu, B., Li, C., Liu, D., Huang, F., Huang, H., Jiang, J., Tu, J., Zhang, J., Zhou, J., et~al.
\newblock Qwen2. 5-1m technical report.
\newblock \emph{arXiv preprint arXiv:2501.15383}, 2025{\natexlab{b}}.

\bibitem[Zhang et~al.(2025{\natexlab{a}})Zhang, Zuo, He, Sun, Liu, Jiang, Fan, Tian, Jia, Li, et~al.]{zhang2025survey}
Zhang, K., Zuo, Y., He, B., Sun, Y., Liu, R., Jiang, C., Fan, Y., Tian, K., Jia, G., Li, P., et~al.
\newblock A survey of reinforcement learning for large reasoning models.
\newblock \emph{arXiv preprint arXiv:2509.08827}, 2025{\natexlab{a}}.

\bibitem[Zhang et~al.(2025{\natexlab{b}})Zhang, Zeng, Li, Huang, Deng, and Dong]{zhang2025realsafe}
Zhang, Y., Zeng, Z., Li, D., Huang, Y., Deng, Z., and Dong, Y.
\newblock {RealSafe-R1}: Safety-aligned {DeepSeek-R1} without compromising reasoning capability.
\newblock \emph{arXiv preprint arXiv:2504.10081}, 2025{\natexlab{b}}.

\bibitem[Zhang et~al.(2025{\natexlab{c}})Zhang, Zhu, Ge, Zhao, Zhou, Li, Feng, Yao, and Han]{zhang2025co}
Zhang, Z., Zhu, J., Ge, X., Zhao, Z., Zhou, Z., Li, X., Feng, X., Yao, J., and Han, B.
\newblock Co-reward: Self-supervised reinforcement learning for large language model reasoning via contrastive agreement.
\newblock \emph{arXiv e-prints}, pp.\  arXiv--2508, 2025{\natexlab{c}}.

\bibitem[Zhao et~al.(2025)Zhao, Kang, Feng, Levine, and Song]{zhao2025learning}
Zhao, X., Kang, Z., Feng, A., Levine, S., and Song, D.
\newblock Learning to reason without external rewards.
\newblock \emph{arXiv preprint arXiv:2505.19590}, 2025.

\bibitem[Zhou et~al.(2025)Zhou, Liang, Liu, Yu, Panaganti, Song, Yu, Zhang, Mi, and Yu]{zhou2025evolving}
Zhou, Y., Liang, Z., Liu, H., Yu, W., Panaganti, K., Song, L., Yu, D., Zhang, X., Mi, H., and Yu, D.
\newblock Evolving language models without labels: Majority drives selection, novelty promotes variation.
\newblock \emph{arXiv preprint arXiv:2509.15194}, 2025.

\bibitem[Zuo et~al.(2025)Zuo, Zhang, Sheng, Qu, Cui, Zhu, Li, Zhang, Long, Hua, et~al.]{zuo2025ttrl}
Zuo, Y., Zhang, K., Sheng, L., Qu, S., Cui, G., Zhu, X., Li, H., Zhang, Y., Long, X., Hua, E., et~al.
\newblock {TTRL}: Test-time reinforcement learning.
\newblock \emph{arXiv preprint arXiv:2504.16084}, 2025.

\end{thebibliography}
\bibliographystyle{icml2026}

%%%%%%%%%%%%%%%%%%%%%%%%%%%%%%%%%%%%%%%%%%%%%%%%%%%%%%%%%%%%%%%%%%%%%%%%%%%%%%%
%%%%%%%%%%%%%%%%%%%%%%%%%%%%%%%%%%%%%%%%%%%%%%%%%%%%%%%%%%%%%%%%%%%%%%%%%%%%%%%
% APPENDIX
%%%%%%%%%%%%%%%%%%%%%%%%%%%%%%%%%%%%%%%%%%%%%%%%%%%%%%%%%%%%%%%%%%%%%%%%%%%%%%%
%%%%%%%%%%%%%%%%%%%%%%%%%%%%%%%%%%%%%%%%%%%%%%%%%%%%%%%%%%%%%%%%%%%%%%%%%%%%%%%
\newpage
\appendix
\onecolumn

\section{Parameters used for TTRL}
\label{sec:TTRL_parameters}
\begin{table}[htbp]
\centering
\caption{Parameters used in the Test-Time Reinforcement Learning (TTRL) setup.}
\renewcommand{\arraystretch}{1.2}
\begin{tabular}{ll}
\toprule
\textbf{Parameter} & \textbf{Value / Description} \\
\midrule
\textbf{Training steps} & 250  \\
\textbf{Max Prompt Length} & 1024 tokens \\
\textbf{Max Response Length} & 6144 tokens \\
\textbf{Train Batch Size} & 8 prompts per rollout \\
\textbf{Mini Batch Size} & 1  \\
\textbf{Micro Batch Size} & 2 \\
\textbf{Samples per Prompt} & 32 \\
\textbf{Votes per Prompt} & 64 \\
\textbf{Learning Rate (Actor)} & $5\times10^{-7}$  \\
\textbf{Learning Rate (Critic)} & $9\times10^{-6}$ \\
\textbf{KL Coefficient} & 0.00 (no KL regularization) \\
\textbf{LoRA rank / Alpha} & 64 / 32 \\
\textbf{Model Parallelism} & Tensor parallel size = 8, FSDP size = 8 \\
\textbf{GPU Memory Utilization} & 0.4 (rollout) / 0.8 (inference) \\
\textbf{Temperature} & 0.6 \\
\textbf{Top-p} & 0.95 \\
\bottomrule
\end{tabular}
\label{tab:ttrl_params}
\end{table}

\section{Additional related work}
\label{sec:additional_related_work}
\paragraph{Self-consistency-based methods for LLM reasoning.} 
The work in \cite{akyurek2024surprising} was one of the pioneering works in improving the reasoning abilities of LLMs using TTT on in-context samples from the ARC-AGI \cite{chollet2019measure}. \cite{jang2025self} uses reasoning-level confidence of sampled answers to identify high-quality reasoning paths for self-training. In \cite{simonds2025ladder}, the authors improve the reasoning abilities of LLMs through self-guided learning by recursively generating and solving progressively simpler variants of complex problems.

\paragraph{Safety vulnerabilities of LLMs.} 
Many papers have considered the safety reasoning-tradeoffs while improving LLM reasoning. In \cite{huang2025safety}, the authors show the safety-reasoning tradeoffs by showing that aligning LLMs can deteriorate their reasoning performance. In \cite{kim2025reasoning}, the authors propose an RL approach that uses a reward signal that balances safety and reasoning. In \cite{zhang2025realsafe}, the authors propose the RealSafe-R1 method, which preserves the models’ reasoning capabilities by maintaining the training data within the original distribution, while being safe. In this work, we also highlight the safety-reasoning tradeoffs with TTT methods.

\section{Additional experimental results}

\subsection{Default TTRL performance}
\label{subsec:default_TTRL}
\begin{figure}[htbp]
    \centering
    % --- First row: Qwen results ---
    \subfloat[]{%
        \includegraphics[width=0.45\textwidth]{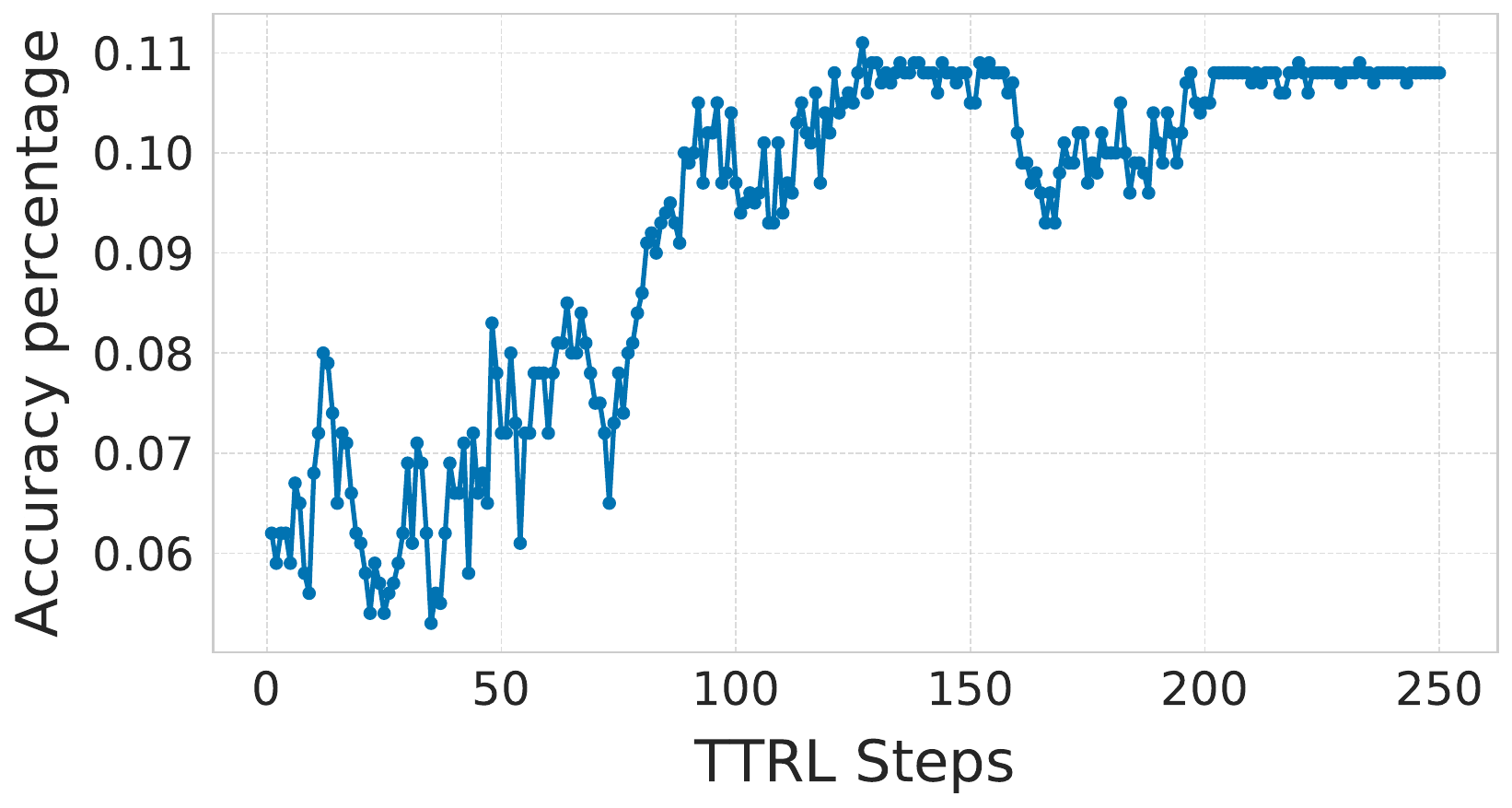}\label{subfig:amc_amc_06_llama1binstruct}
    }
    \hfill
    \subfloat[]{%
        \includegraphics[width=0.45\textwidth]{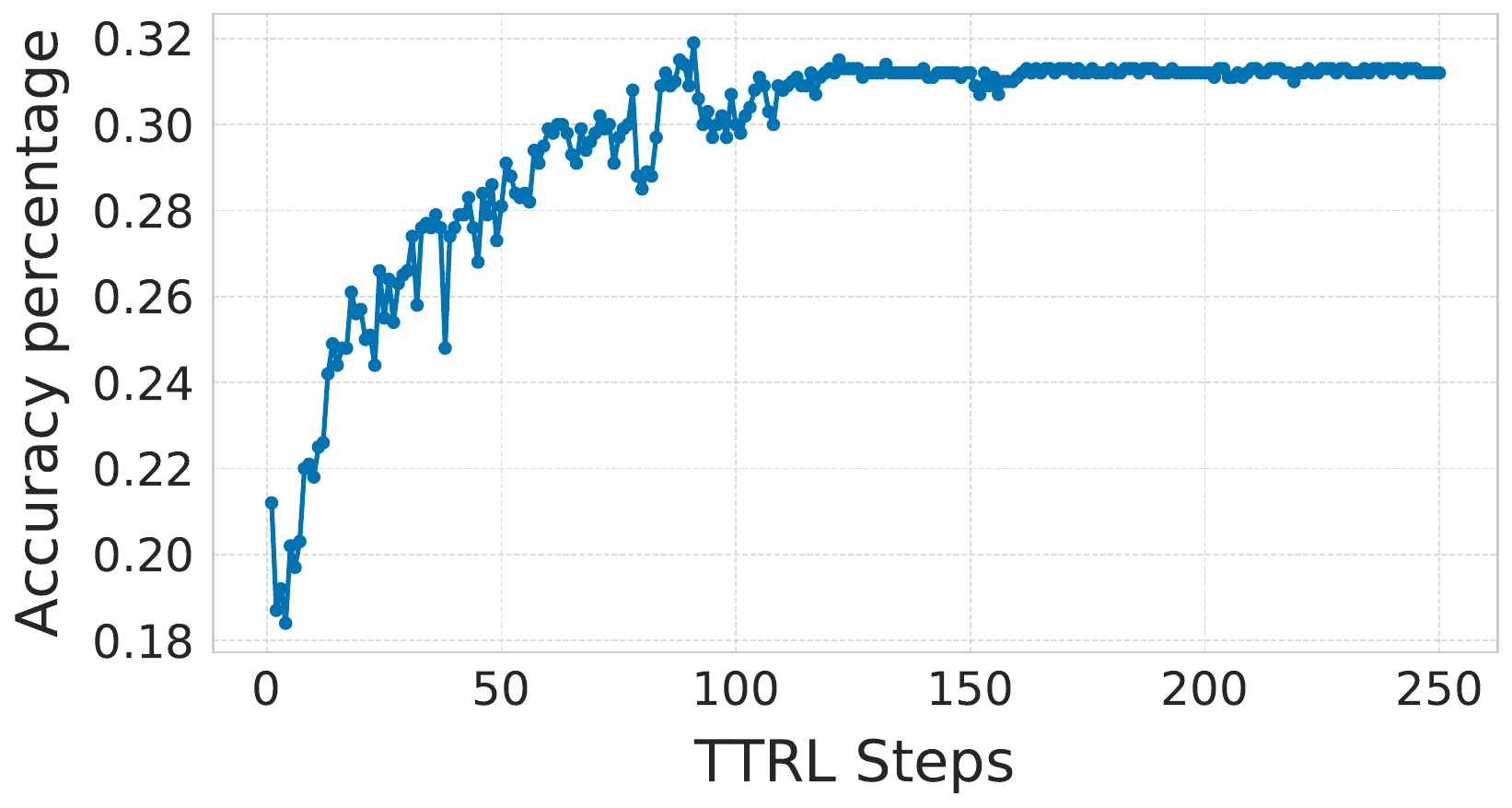}\label{subfig:amc_amc_06_llama3binstruct}
    }

    % --- Second row: Llama results ---
    
    \subfloat[]{%
        \includegraphics[width=0.45\textwidth]{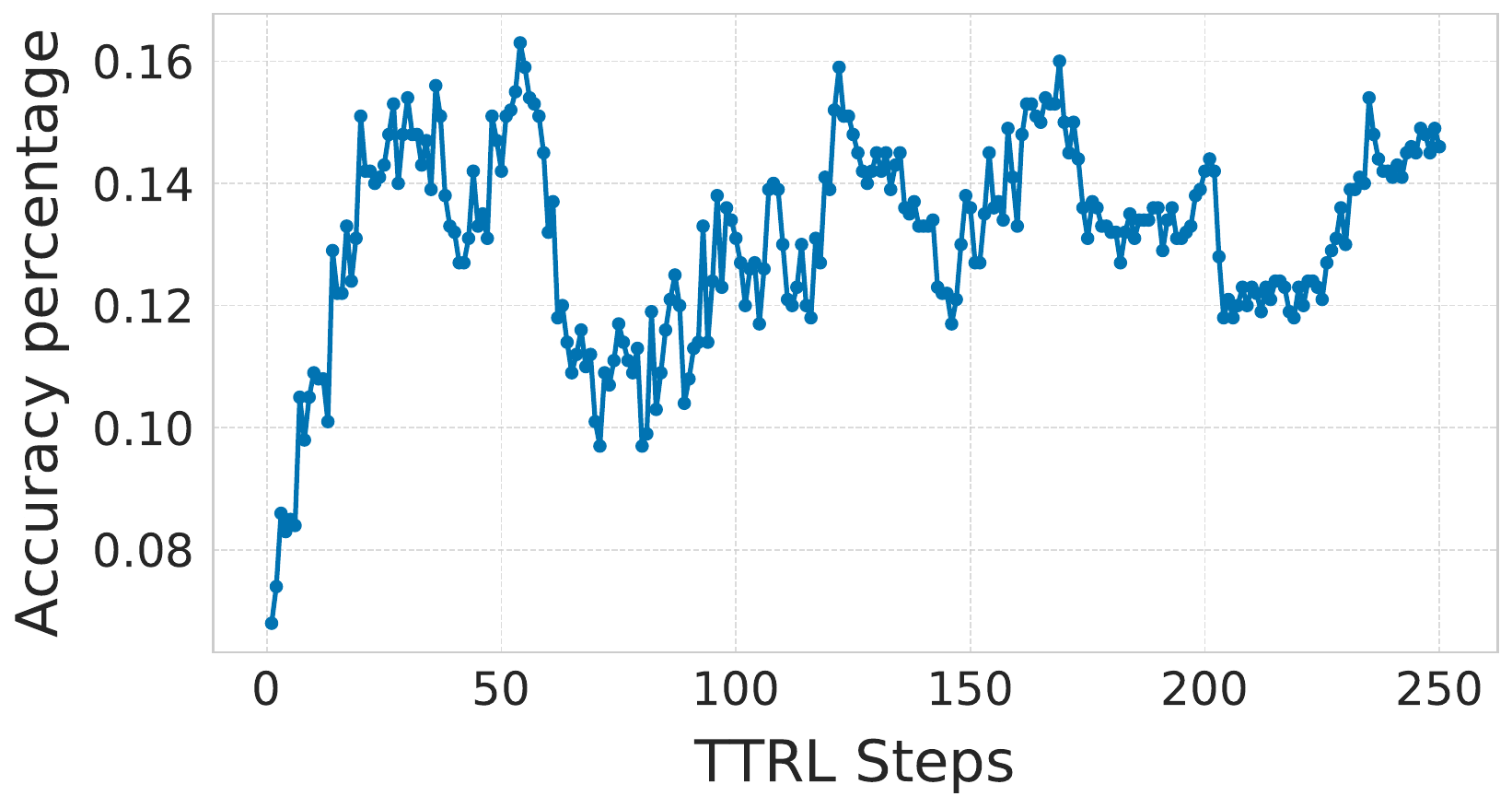}\label{subfig:amc_amc_06_llama8binstruct}
    }
    \hfill
    \subfloat[]{%
        \includegraphics[width=0.45\textwidth]{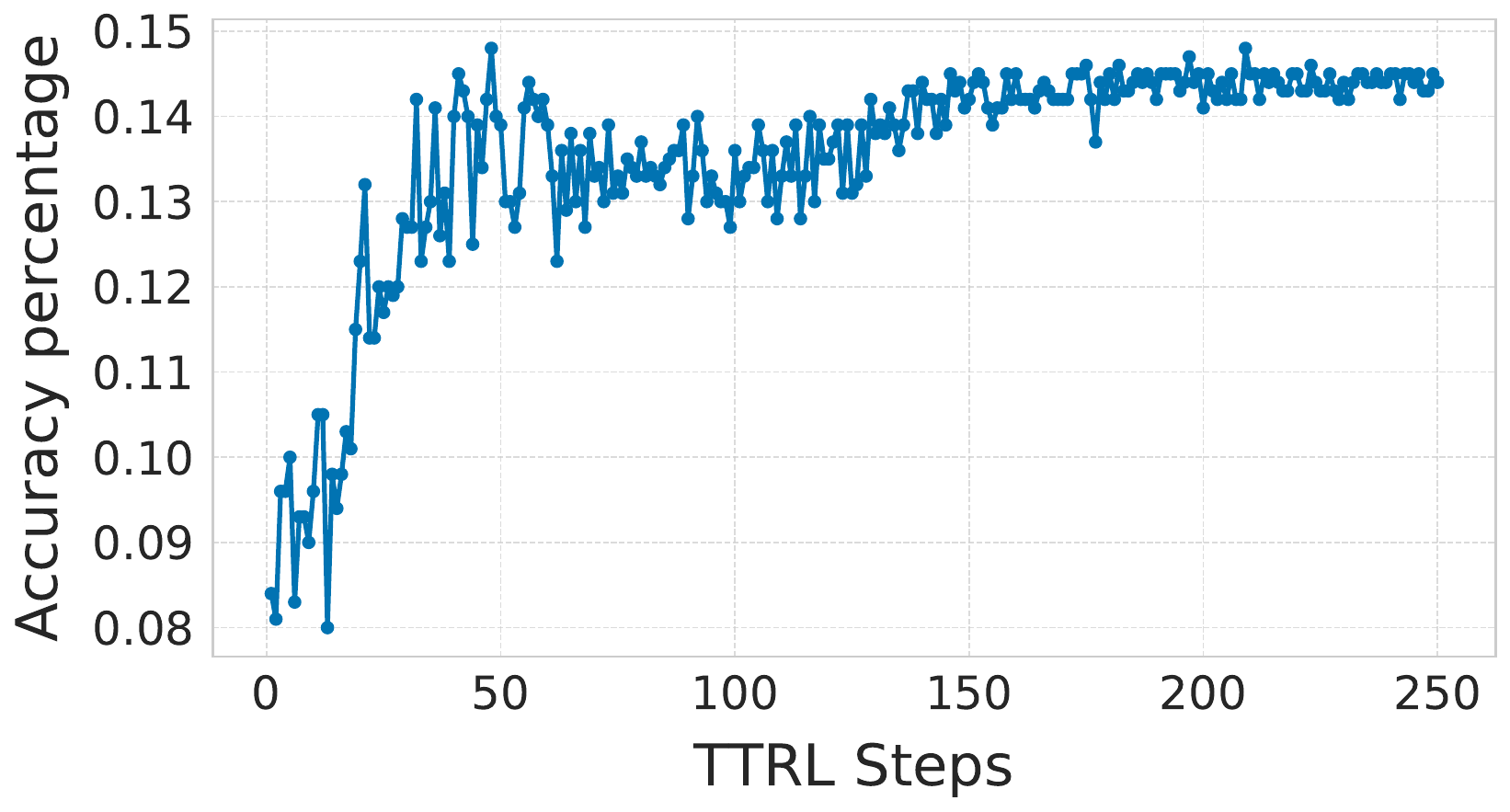}\label{subfig:amc_amc_qwen0.5binstruct}
    }
    \caption{Default AMC accuracy (\%) during TTRL on all the instruction-tuned models.}
    \label{fig:default_TTRL}
\end{figure}

In this section, we plot the default TTRL performance without any injection for all the instruction-tuned models considered in the paper. The results are shown in Figure \ref{fig:default_TTRL}.

\subsection{Additional results for RQ1}
\label{subsec:rq1_additional_results}
\begin{figure*}[t]
    \centering
    % --- First row: Qwen results ---
    \subfloat[]{%
        \includegraphics[width=0.3\textwidth]{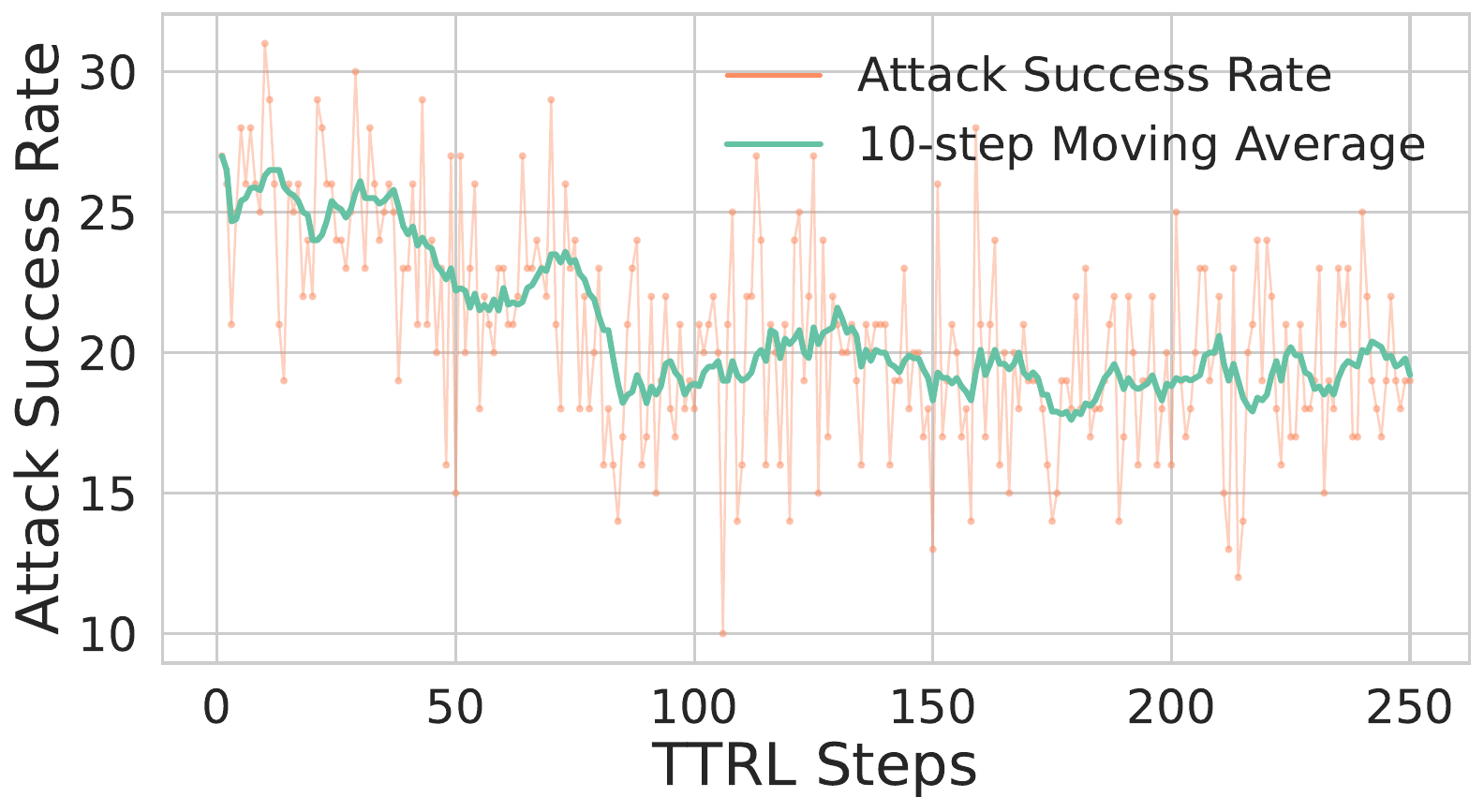}\label{subfig:amc_jailbreak2_qwen0.5binstruct}
    }
    \hfill
    \subfloat[]{%
        \includegraphics[width=0.3\textwidth]{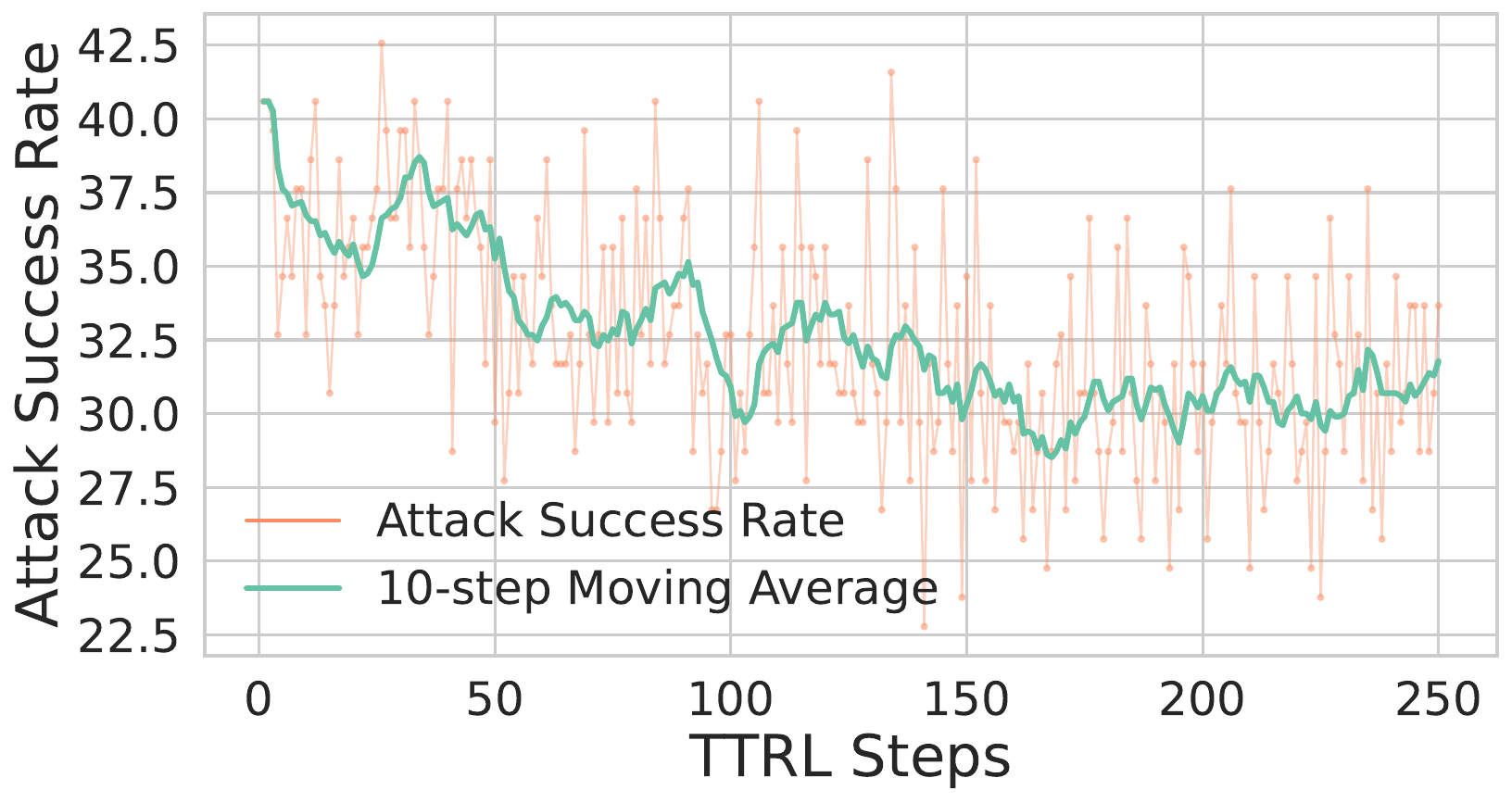}\label{subfig:amc_wildjailbreak_qwen0.5binstruct}
    }
    \hfill
    \subfloat[]{%
        \includegraphics[width=0.3\textwidth]{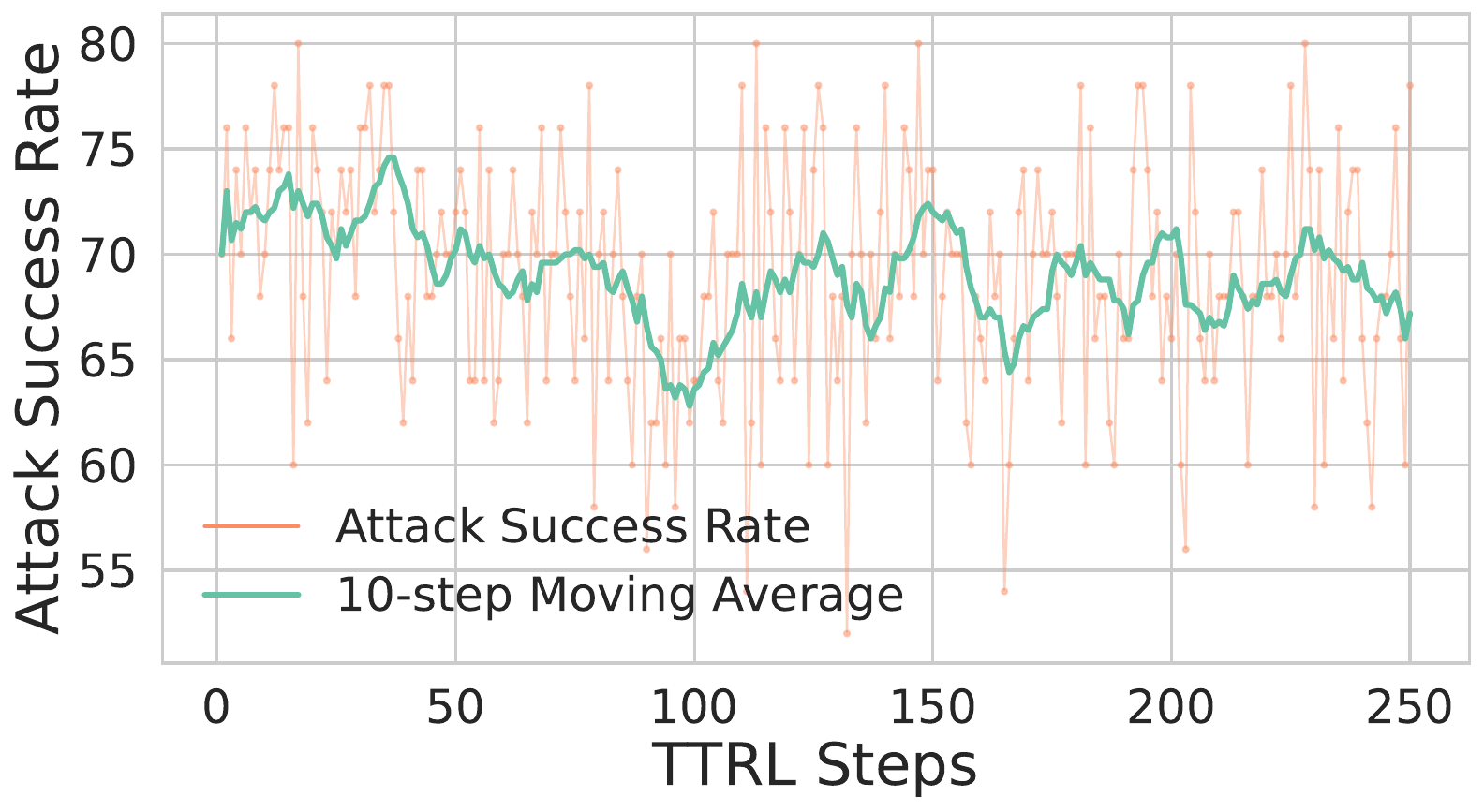}\label{subfig:amc_llamaartifacts_qwen0.5binstruct}
    }

    % --- Second row: Llama results ---
    
    \subfloat[]{%
        \includegraphics[width=0.3\textwidth]{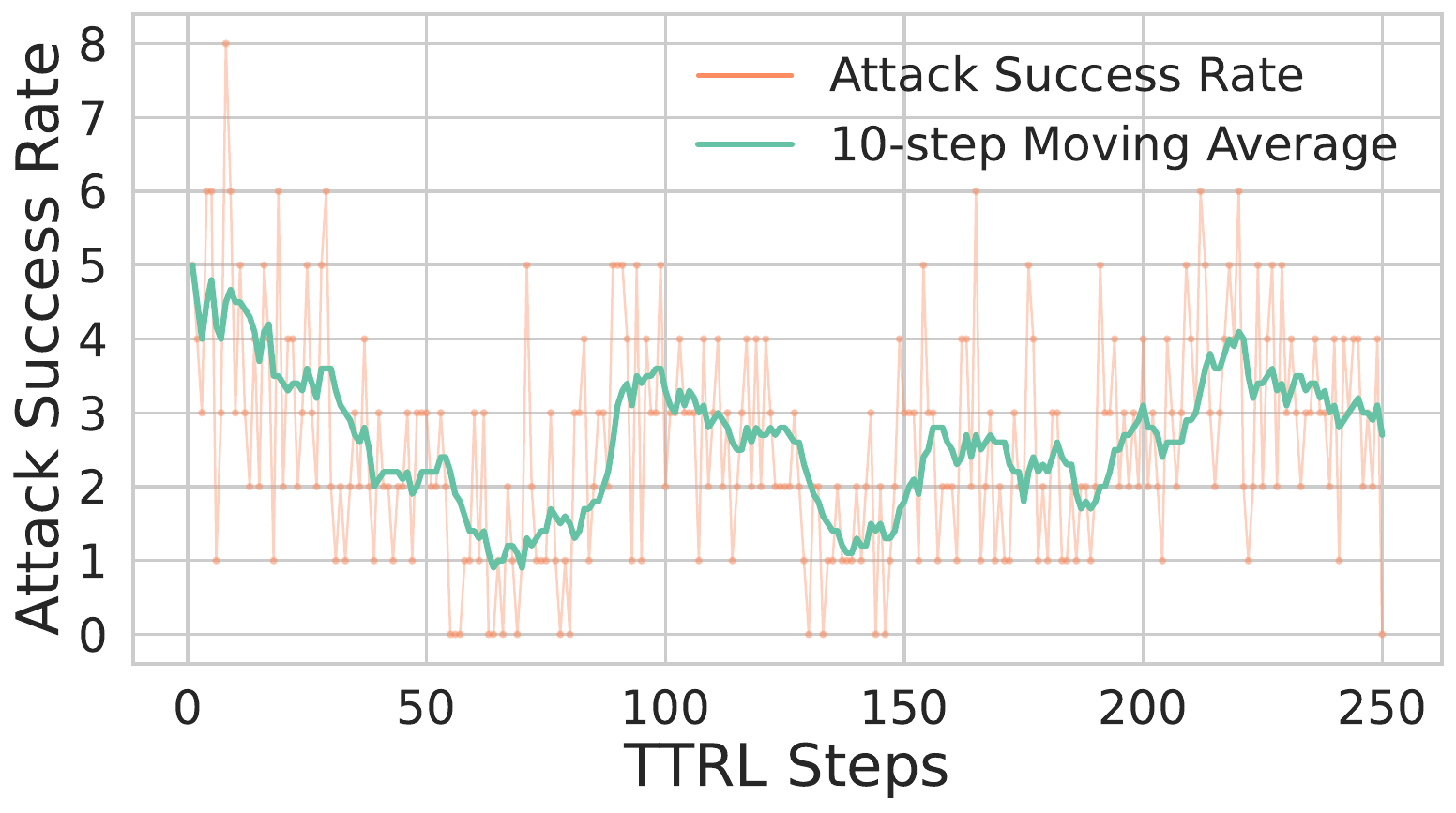}\label{subfig:amc_jailbreak2_llama3binstruct}
    }
    \hfill
    \subfloat[]{%
        \includegraphics[width=0.3\textwidth]{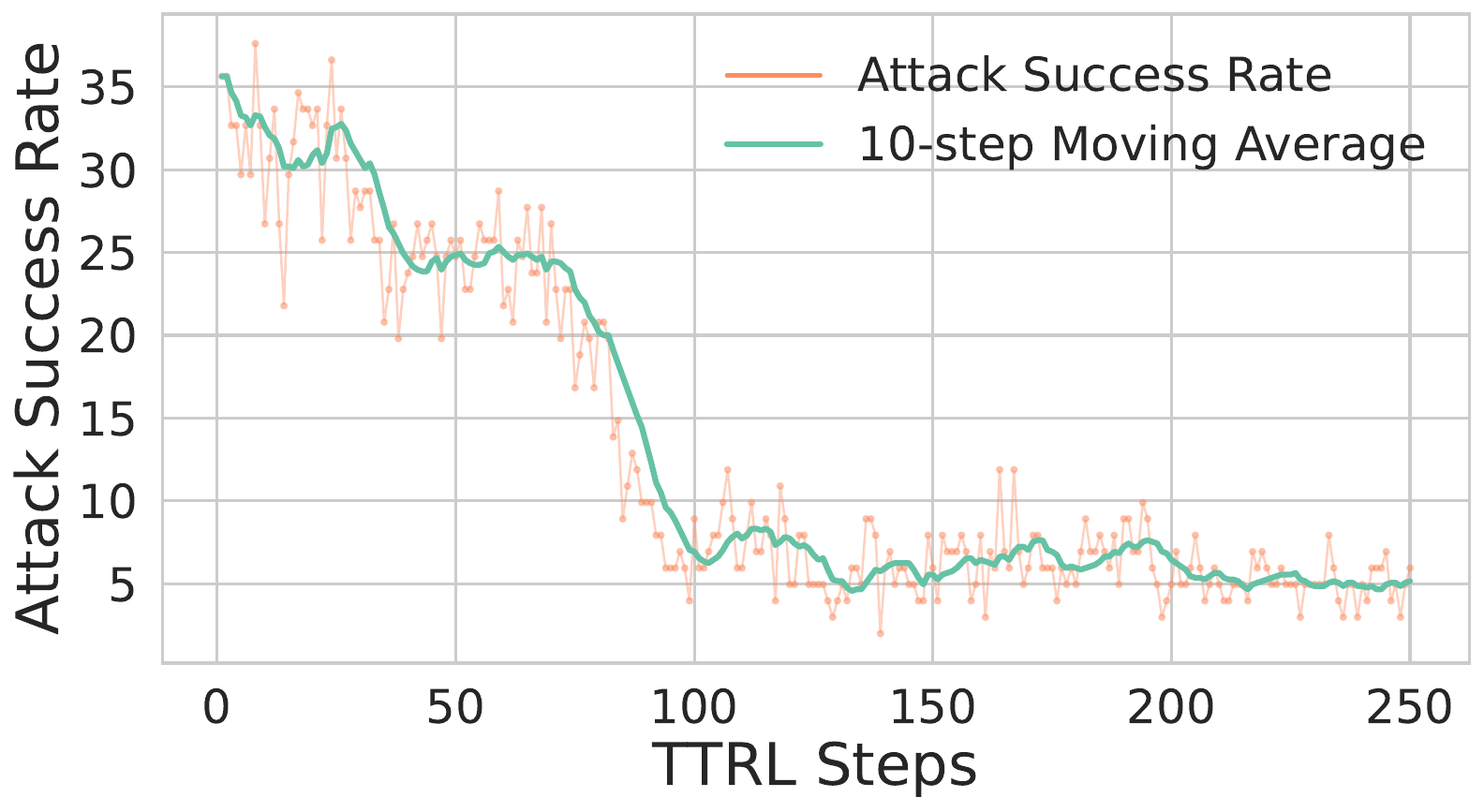}\label{subfig:amc_wildjailbreak_llama3binstruct}
    }
    \hfill
    \subfloat[]{%
        \includegraphics[width=0.3\textwidth]{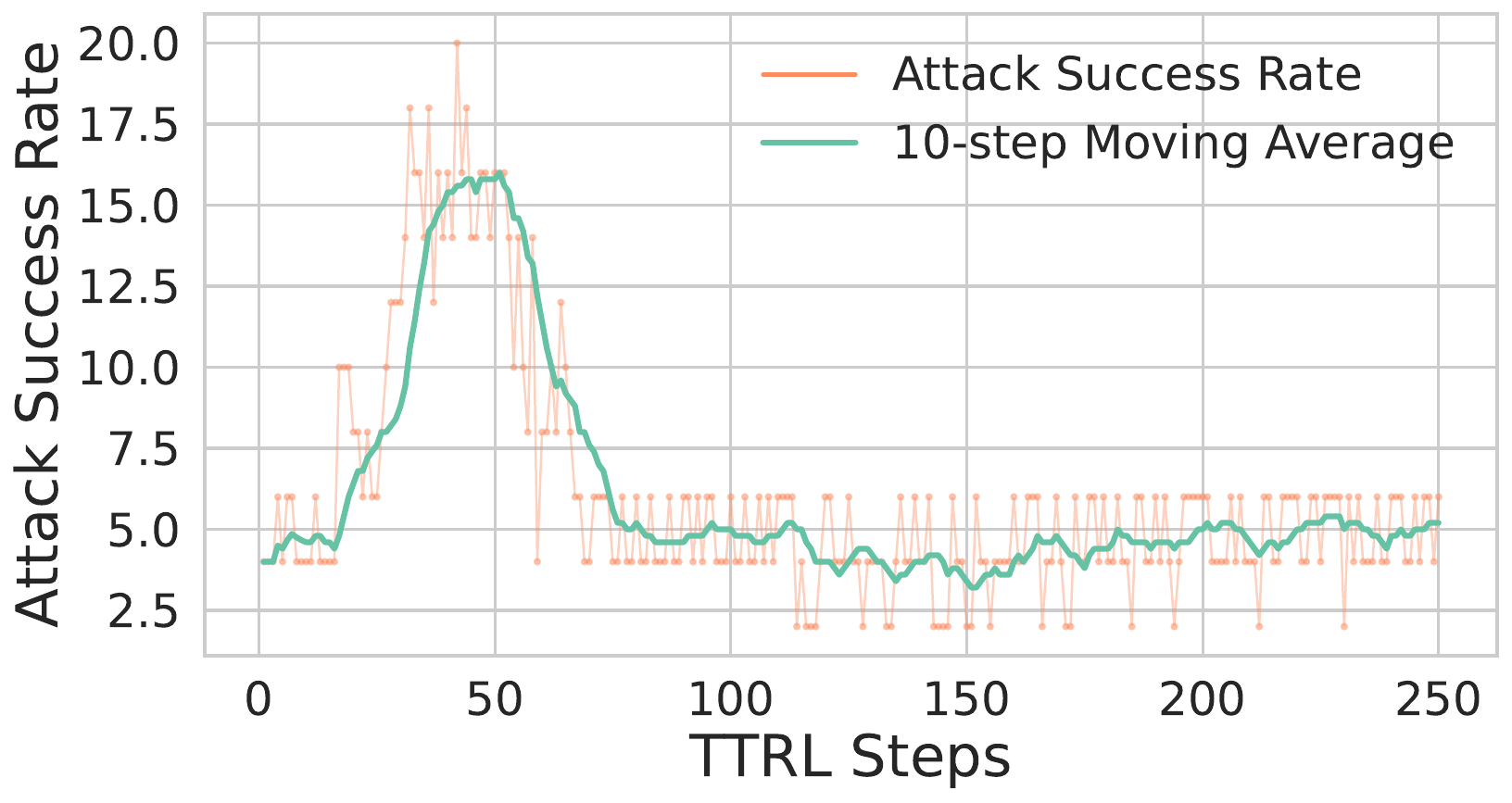}\label{subfig:amc_llamaartifacts_llama3binstruct}
    }
    \caption{ASR measured across three jailbreak datasets, JailbreakV-28k, WildJailbreak, and Llama Artifacts (left to right, respectively) during TTRL, for Qwen-0.5B-Instruct (top row) and Llama-3B-Instruct (bottom row).}
    \label{fig:rq1_additional_results}
\end{figure*}

In this section, we present the plots for RQ1, i.e., the ASR (\%) on the rest of the instruction-tuned models not in the main paper in Figure \ref{fig:rq1_additional_results}.

\subsection{Additional results for RQ2: harmful prompt injections}
\label{subsec:rq2_additional_results}
\begin{figure}[htbp]
    \centering
    % --- First row: Qwen results ---
    \subfloat[]{%
        \includegraphics[width=0.3\textwidth]{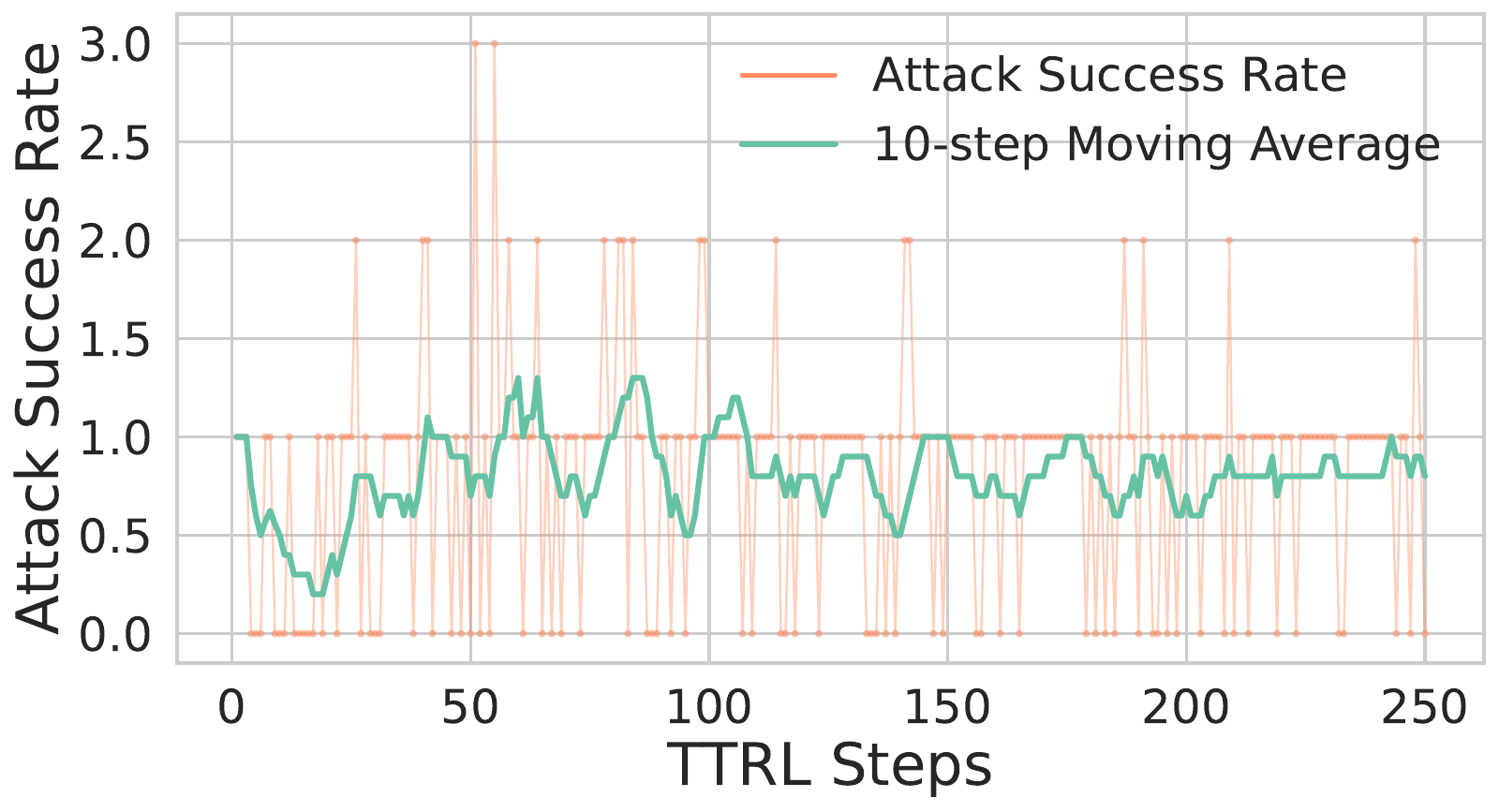}\label{subfig:amcjail_jailbreak2_llama8binstruct}
    }
    \hfill
    \subfloat[]{%
        \includegraphics[width=0.3\textwidth]{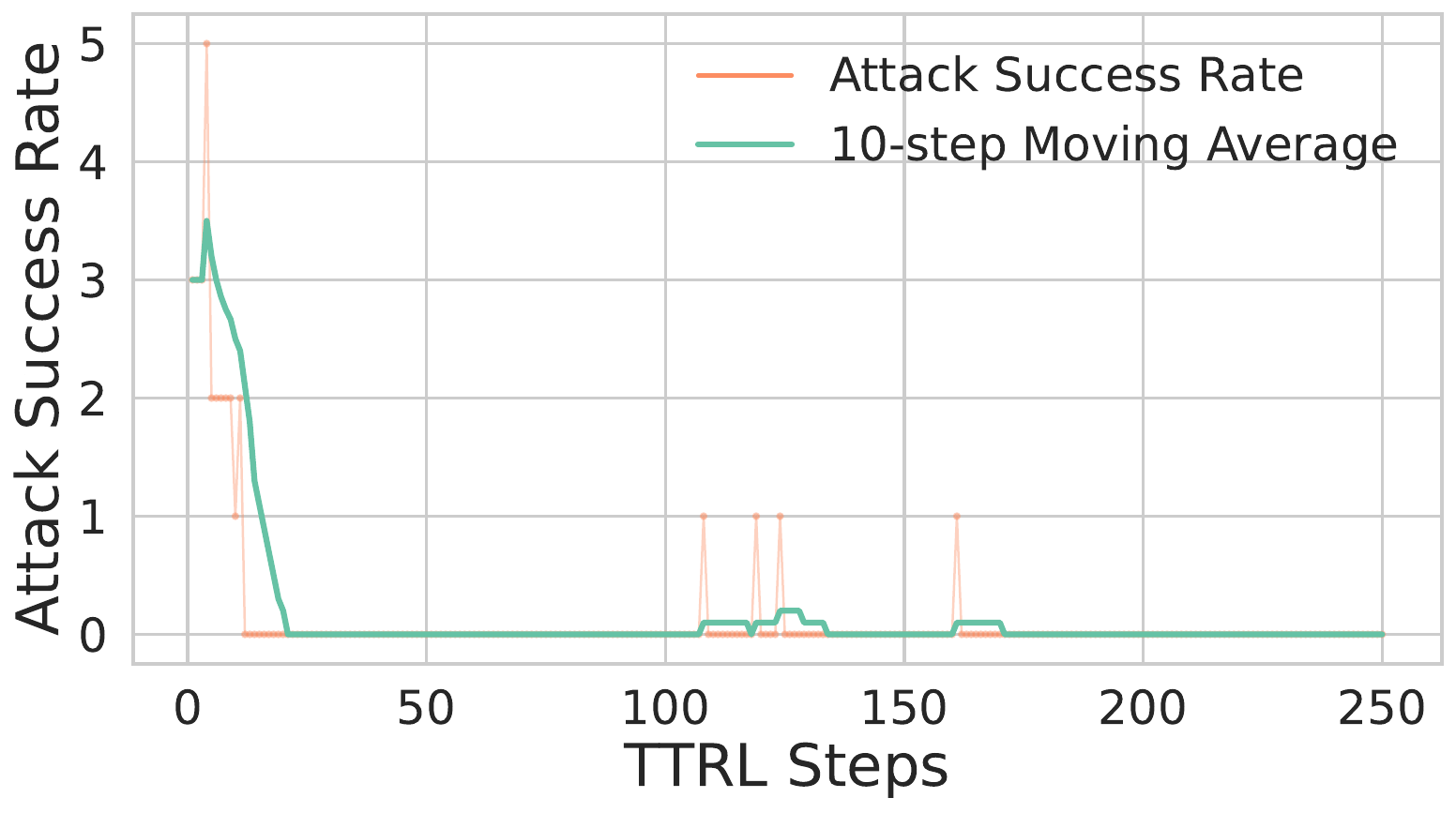}\label{subfig:amcwildjail_wildjail_llama8binstruct}
    }
    \hfill
    \subfloat[]{%
        \includegraphics[width=0.3\textwidth]{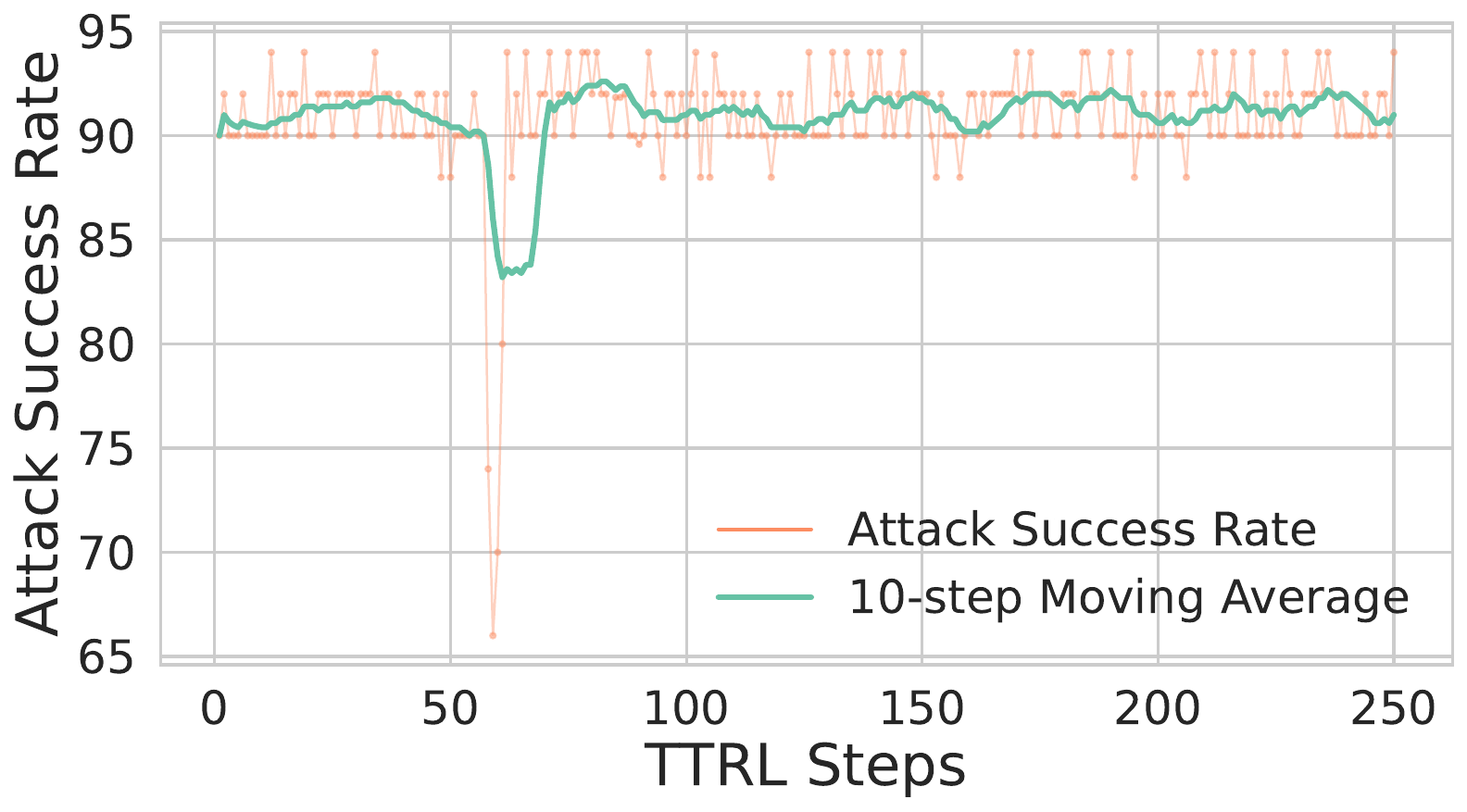}\label{subfig:amcllamaartifacts_llamaartifacts_llama8binstruct}
    }

    % --- Second row: Llama results ---
    
    \subfloat[]{%
        \includegraphics[width=0.3\textwidth]{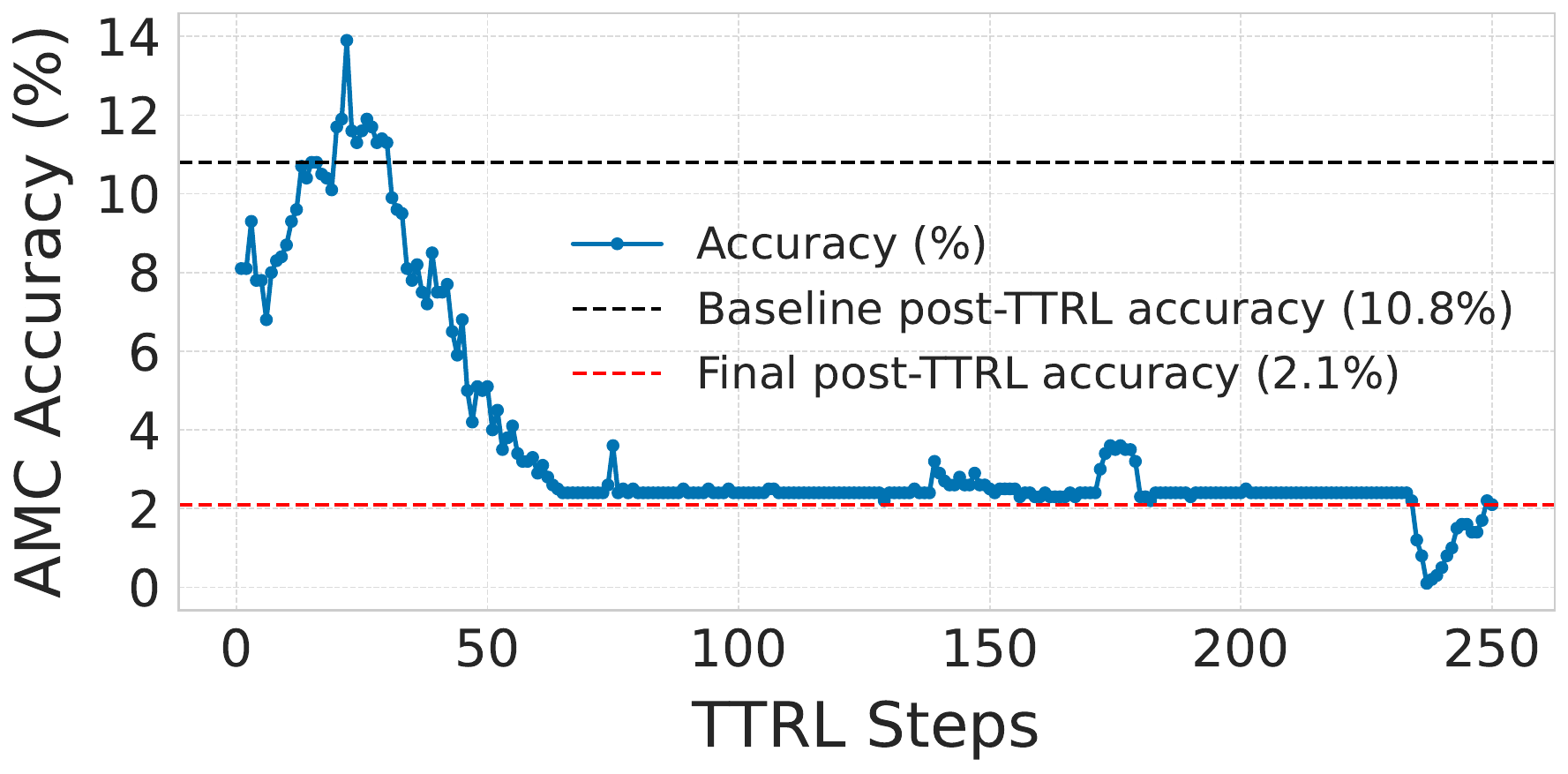}\label{subfig:amcjail_amc_llama8binstruct}
    }
    \hfill
    \subfloat[]{%
        \includegraphics[width=0.3\textwidth]{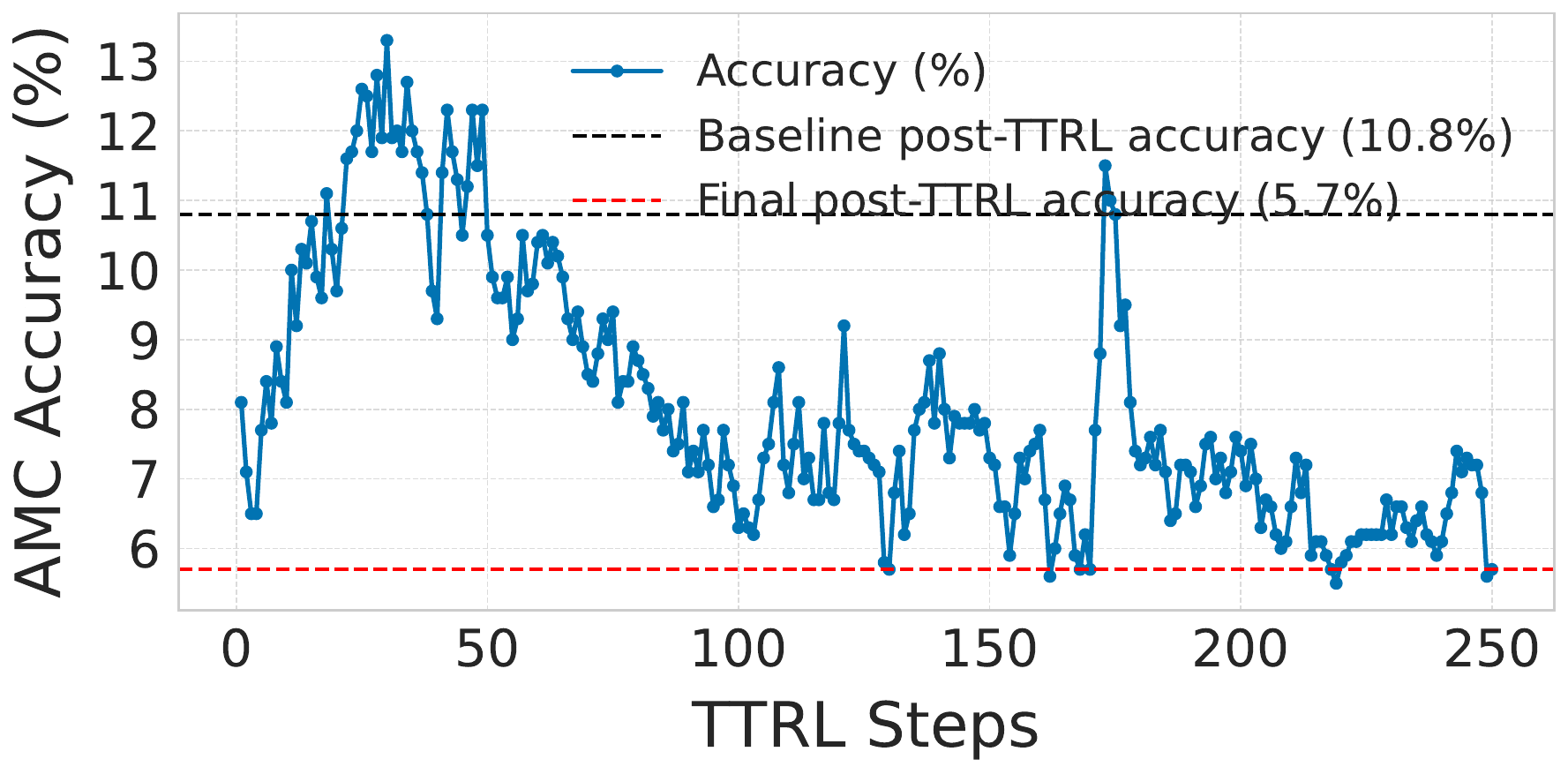}\label{subfig:amcwildjail_amc_llama8binstruct}
    }
    \hfill
    \subfloat[]{%
        \includegraphics[width=0.3\textwidth]{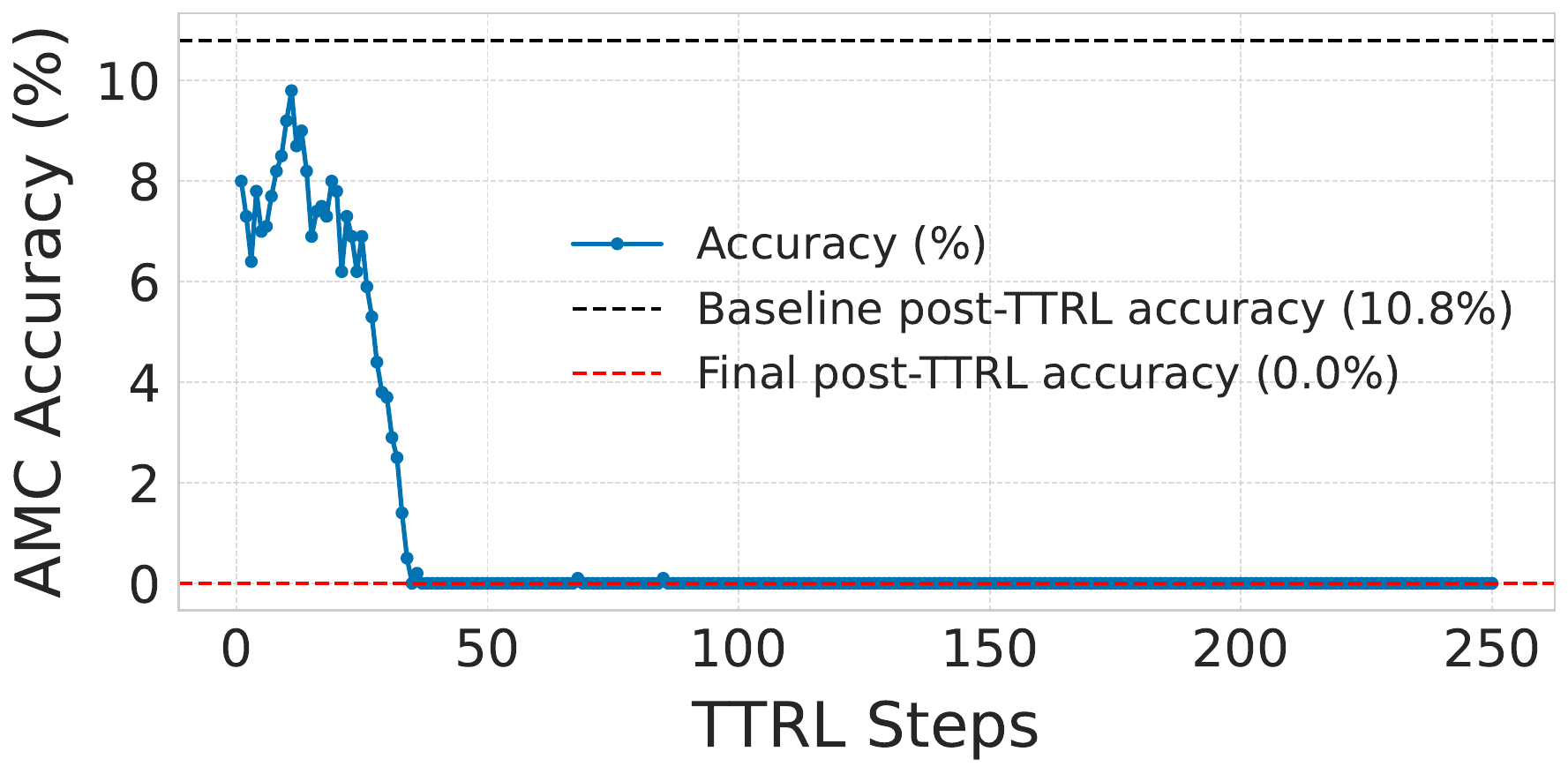}\label{subfig:amcllamaartifacts_amc_llama8binstruct}
    }

    \caption{Impact on safety and reasoning for Llama3-8B-Instruct model after harmful prompt injection across three jailbreak datasets, JailbreakV-28k, WildJailbreak, and Llama Artifacts (left to right, respectively) during TTRL, for safety (top row) and AMC accuracy (bottom row).}
    \label{fig:rq2_llama8b_3attacks}
\end{figure}

\begin{figure}[htbp]
    \centering
    % --- First row: Qwen results ---
    \subfloat[]{%
        \includegraphics[width=0.3\textwidth]{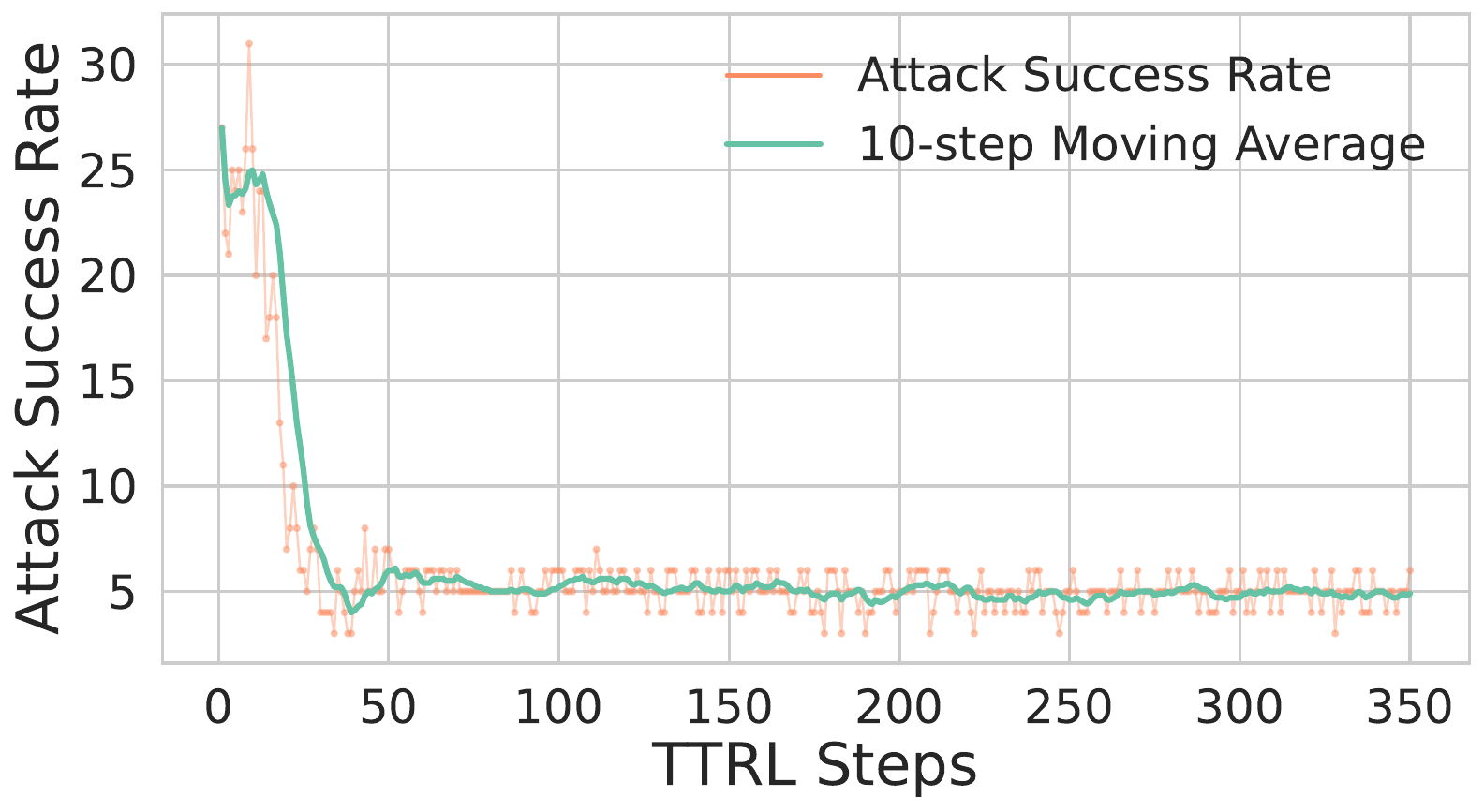}\label{subfig:amcjail_jailbreak2_qwen0.5binstruct}
    }
    \hfill
    \subfloat[]{%
        \includegraphics[width=0.3\textwidth]{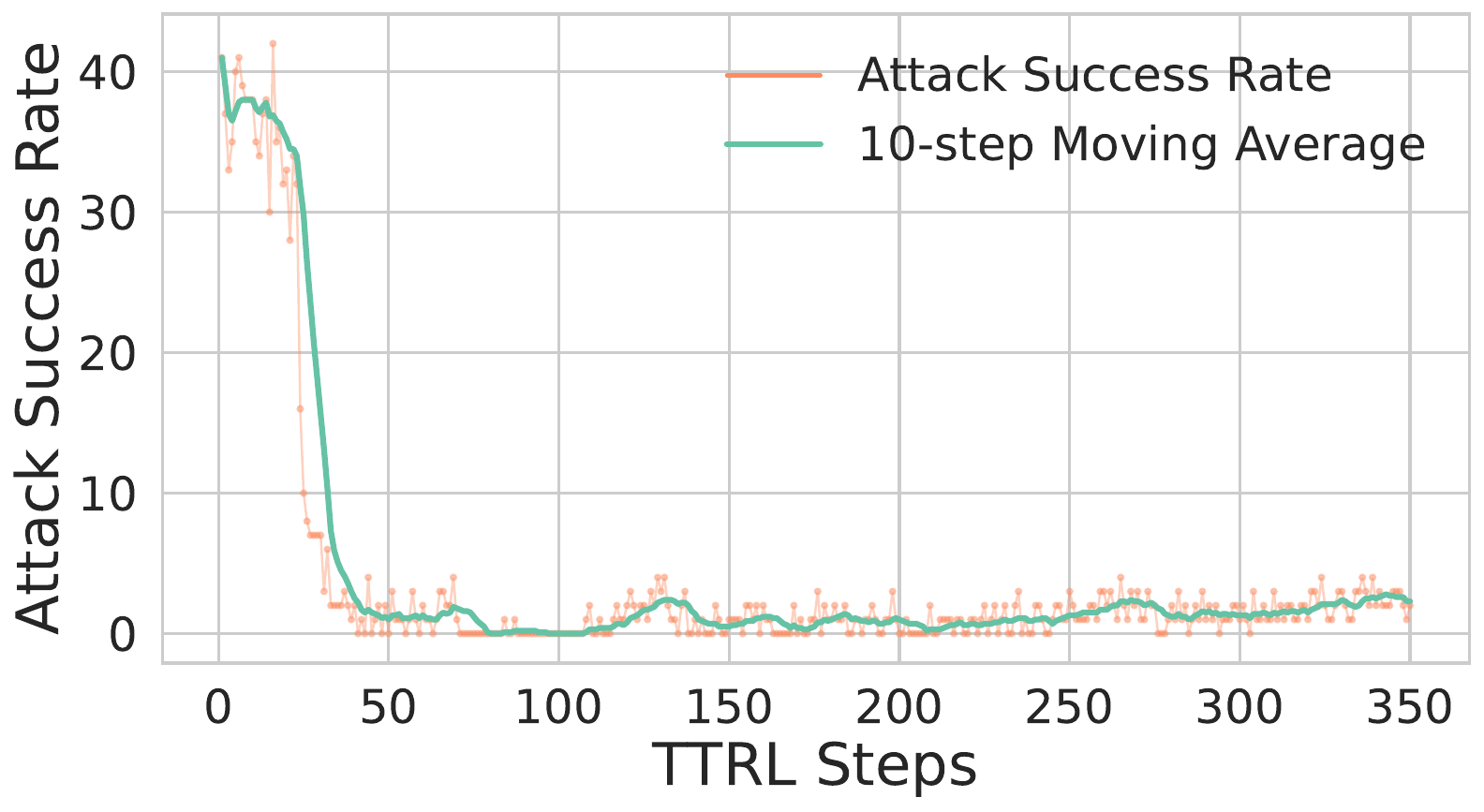}\label{subfig:amcwildjail_wildjail_qwen0.5binstruct}
    }
    \hfill
    \subfloat[]{%
        \includegraphics[width=0.3\textwidth]{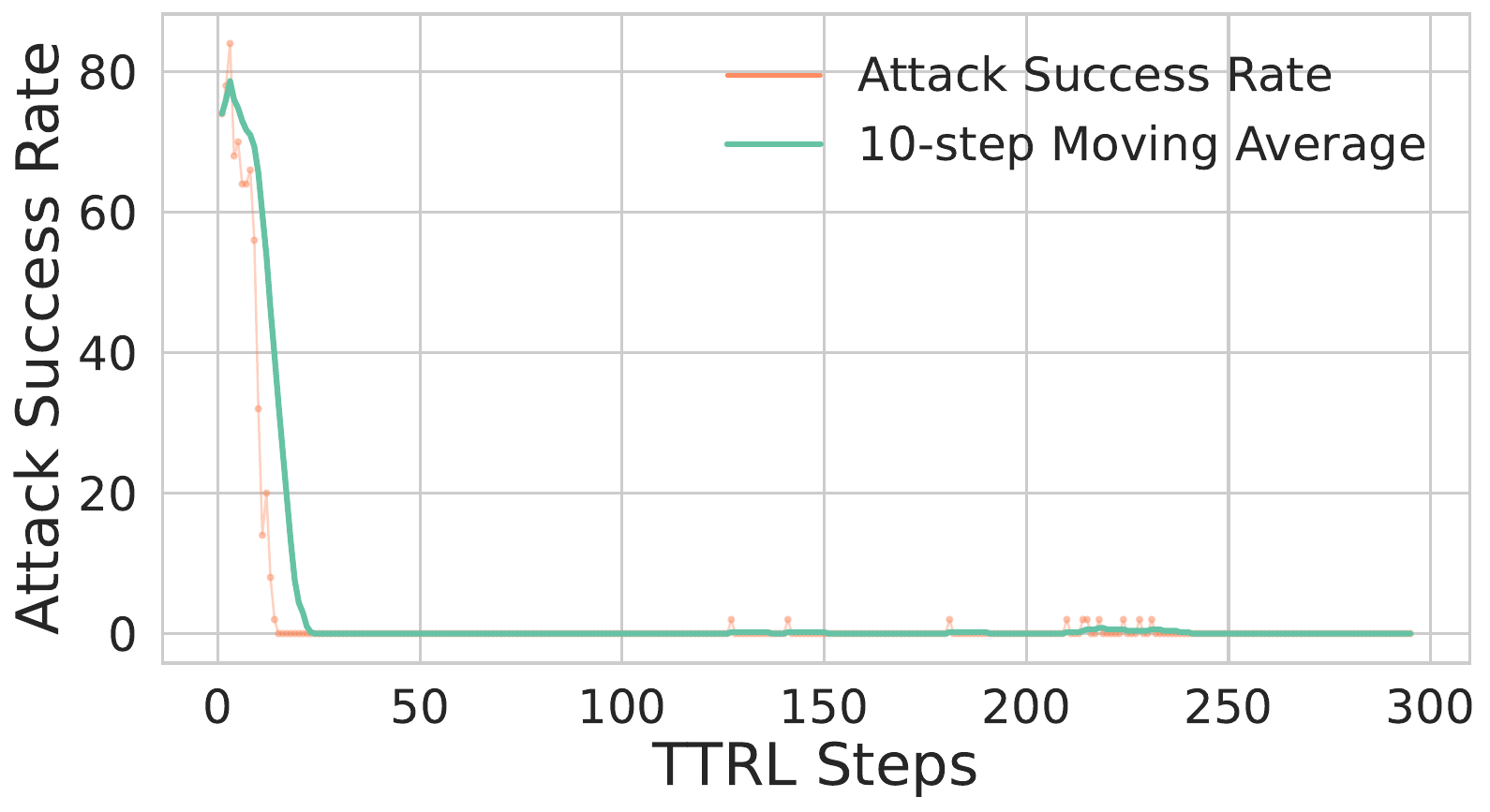}\label{subfig:amcllamaartifacts_llamaartifacts_qwen0.5binstruct}
    }

    % --- Second row: Llama results ---
    
    \subfloat[]{%
        \includegraphics[width=0.3\textwidth]{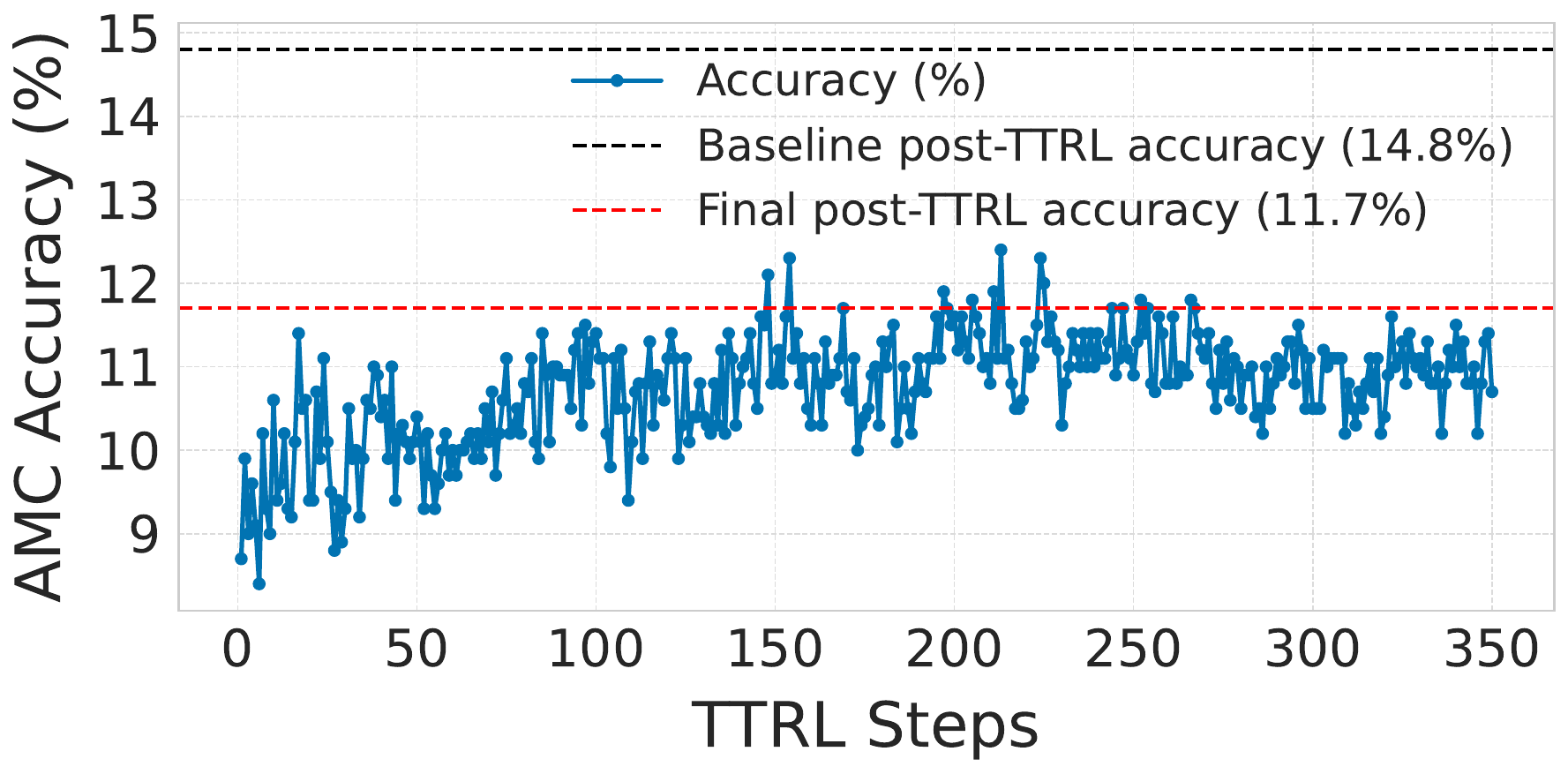}\label{subfig:amcjail_amc_qwen0.5binstruct}
    }
    \hfill
    \subfloat[]{%
        \includegraphics[width=0.3\textwidth]{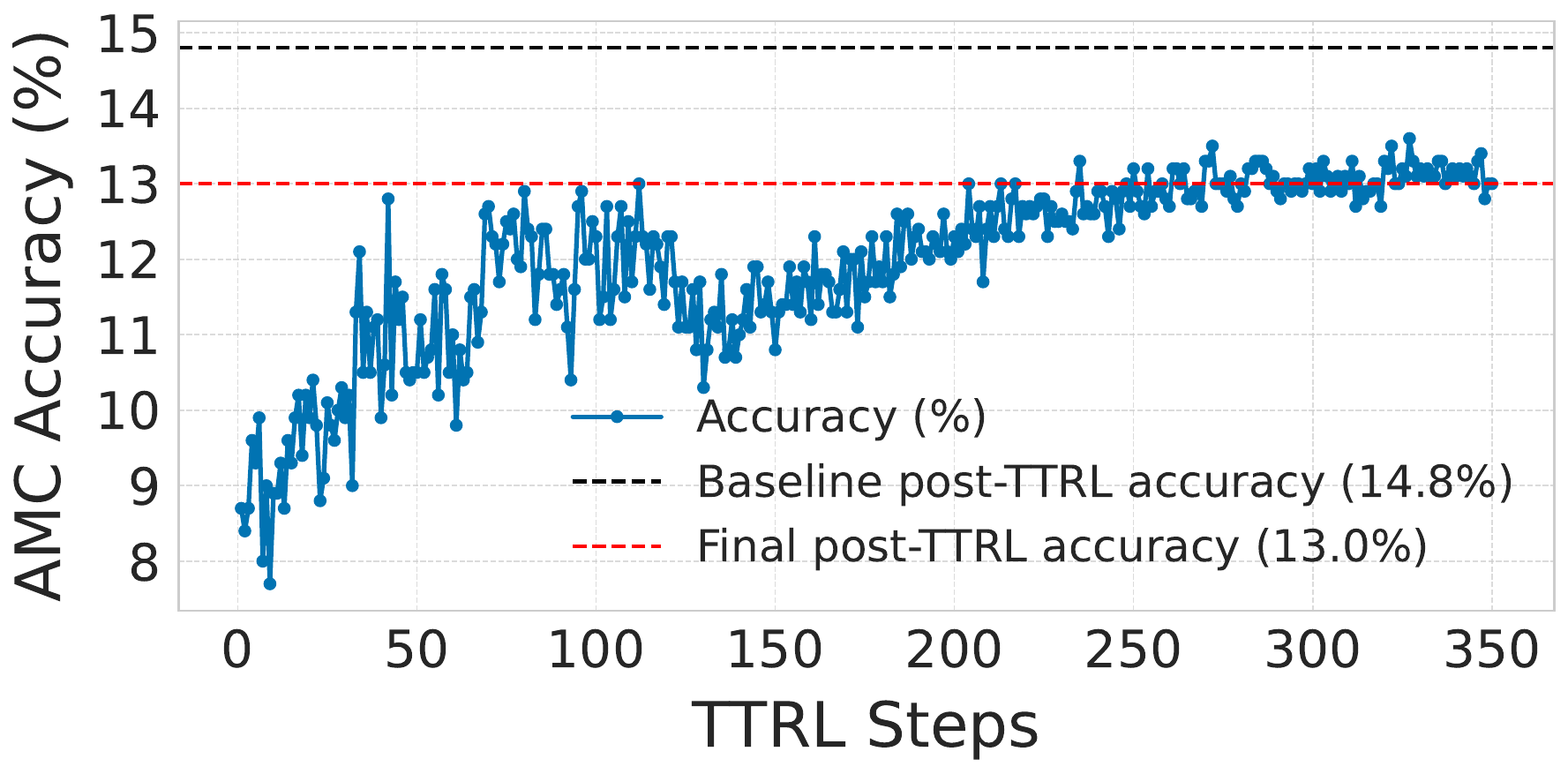}\label{subfig:amcwildjail_amc_qwen0.5binstruct}
    }
    \hfill
    \subfloat[]{%
        \includegraphics[width=0.3\textwidth]{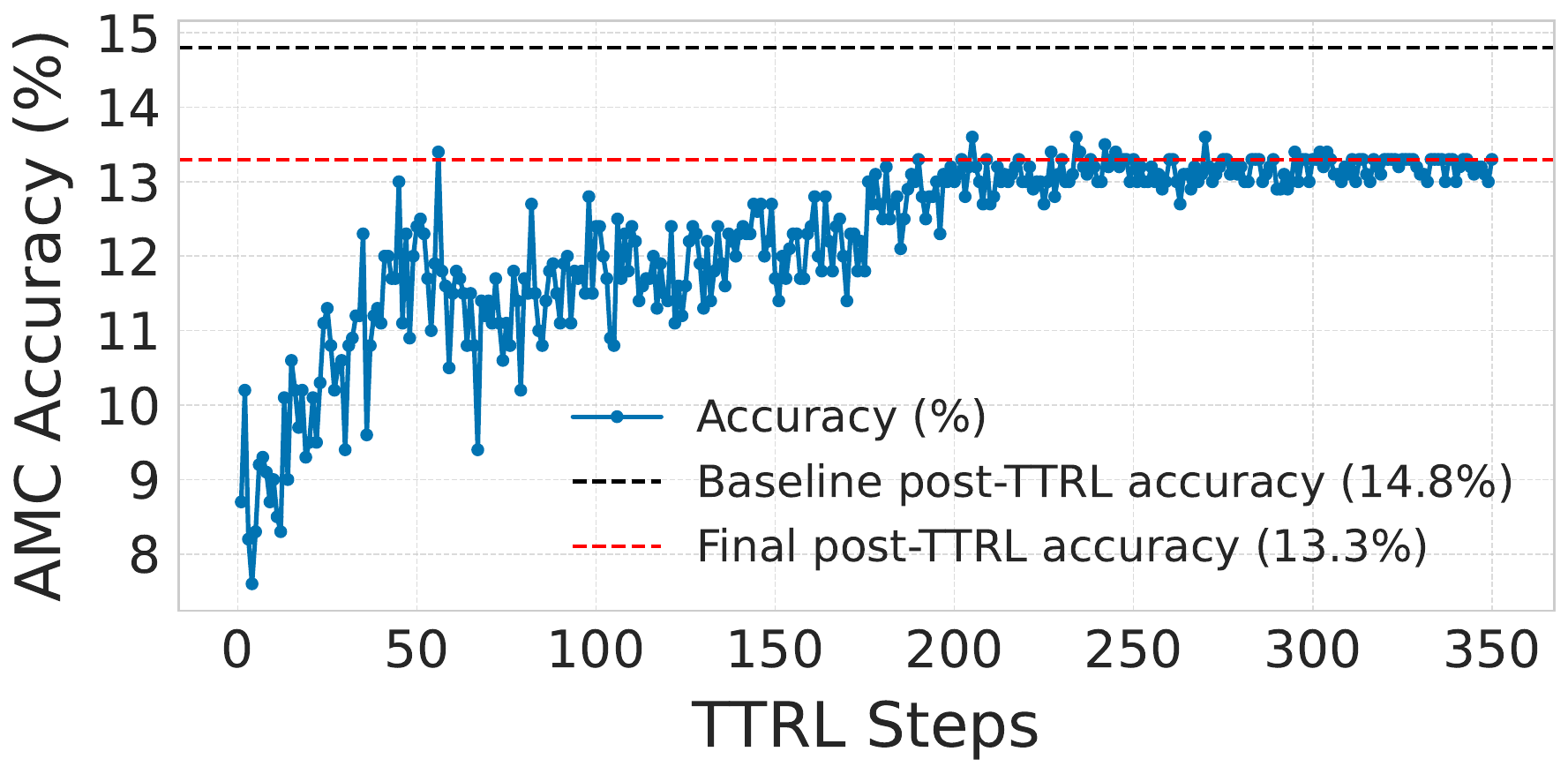}\label{subfig:amcllamaartifacts_amc_qwen0.5binstruct}
    }

    \caption{Impact on safety and reasoning for Qwen-0.5B-Instruct model after harmful prompt injection across three jailbreak datasets, JailbreakV-28k, WildJailbreak, and Llama Artifacts (left to right, respectively) during TTRL, for safety (top row) and AMC accuracy (bottom row).}
    \label{fig:rq2_qwen0.5b_3attacks}
\end{figure}

\begin{figure}[htbp]
    \centering
    % --- First row: Qwen results ---
    \subfloat[]{%
        \includegraphics[width=0.3\textwidth]{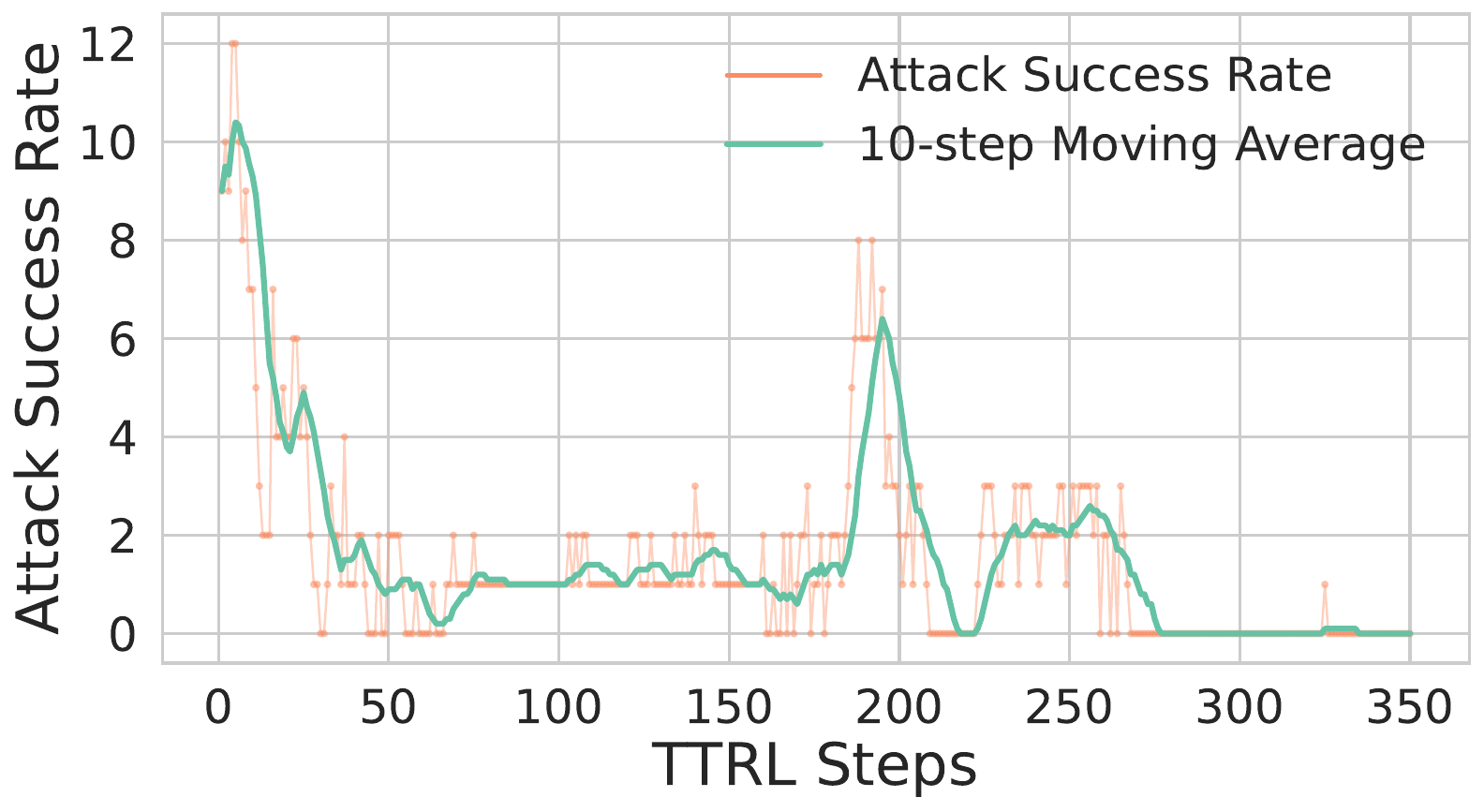}\label{subfig:amcjail_jailbreak2_llama1binstruct}
    }
    \hfill
    \subfloat[]{%
        \includegraphics[width=0.3\textwidth]{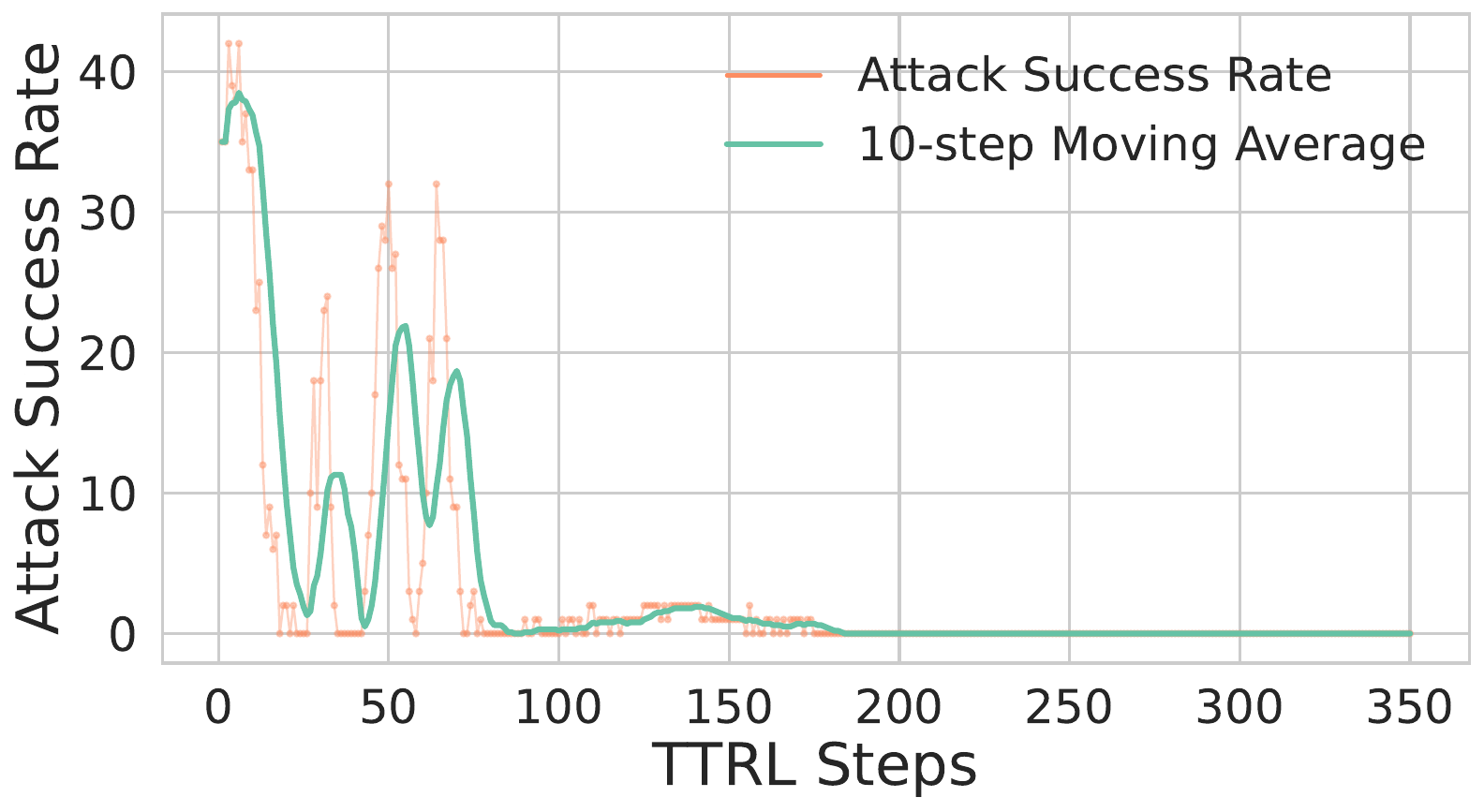}\label{subfig:amcwildjail_wildjail_llama1b}
    }
    \hfill
    \subfloat[]{%
        \includegraphics[width=0.3\textwidth]{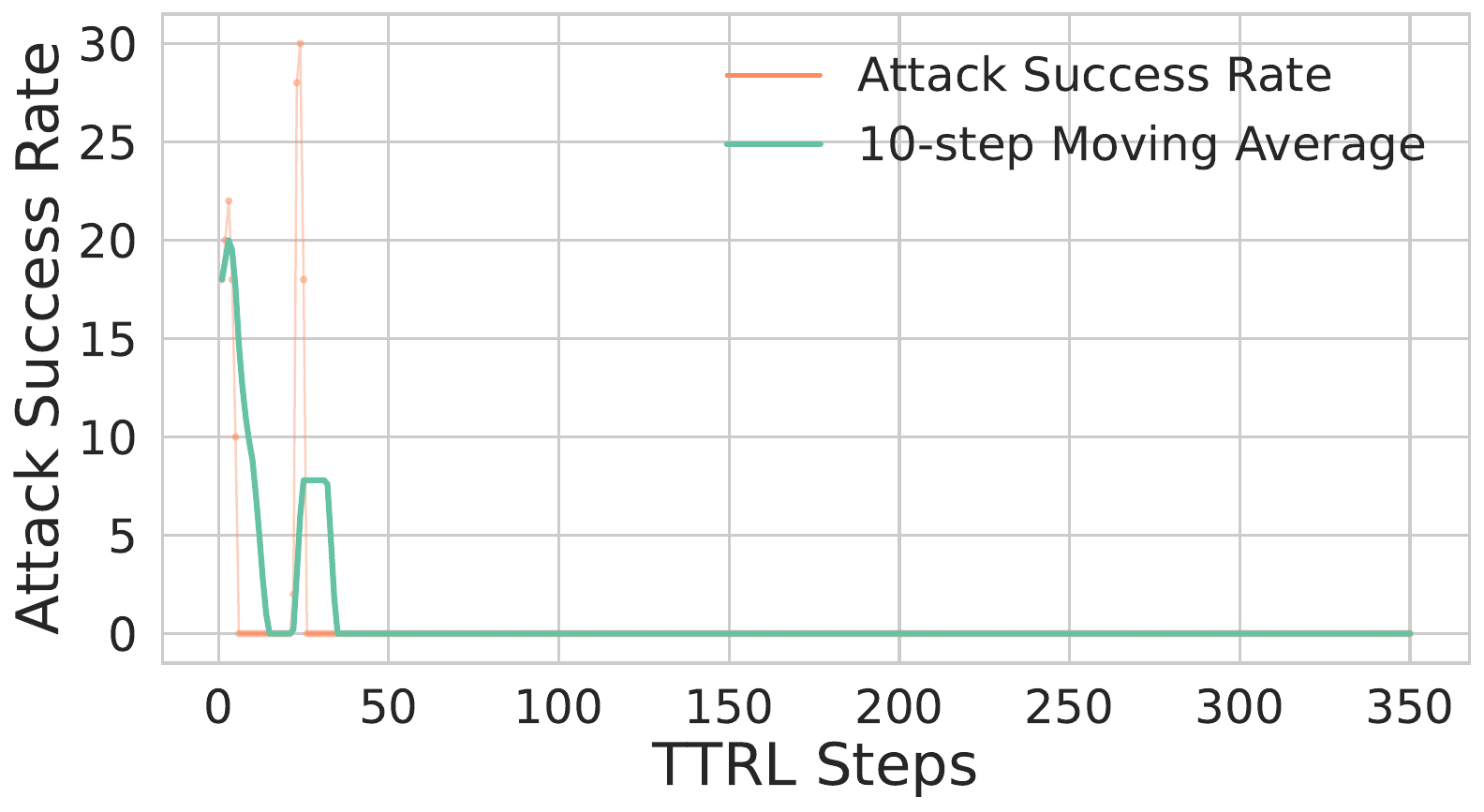}\label{subfig:amcllamaartifacts_llamaartifacts_llama1binstruct}
    }

    % --- Second row: Llama results ---
    
    \subfloat[]{%
        \includegraphics[width=0.3\textwidth]{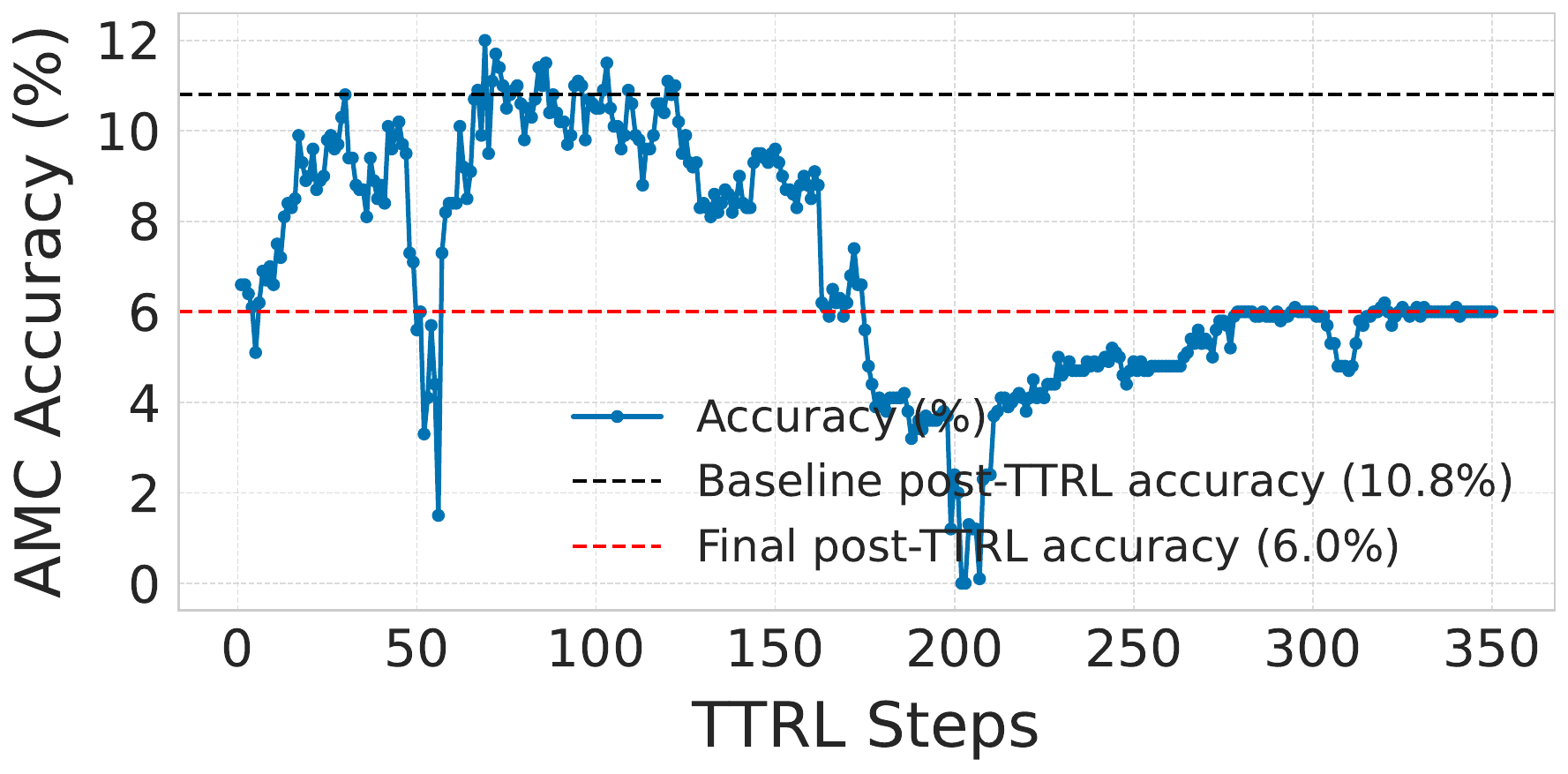}\label{subfig:amcjail_amc_llama1binstruct}
    }
    \hfill
    \subfloat[]{%
        \includegraphics[width=0.3\textwidth]{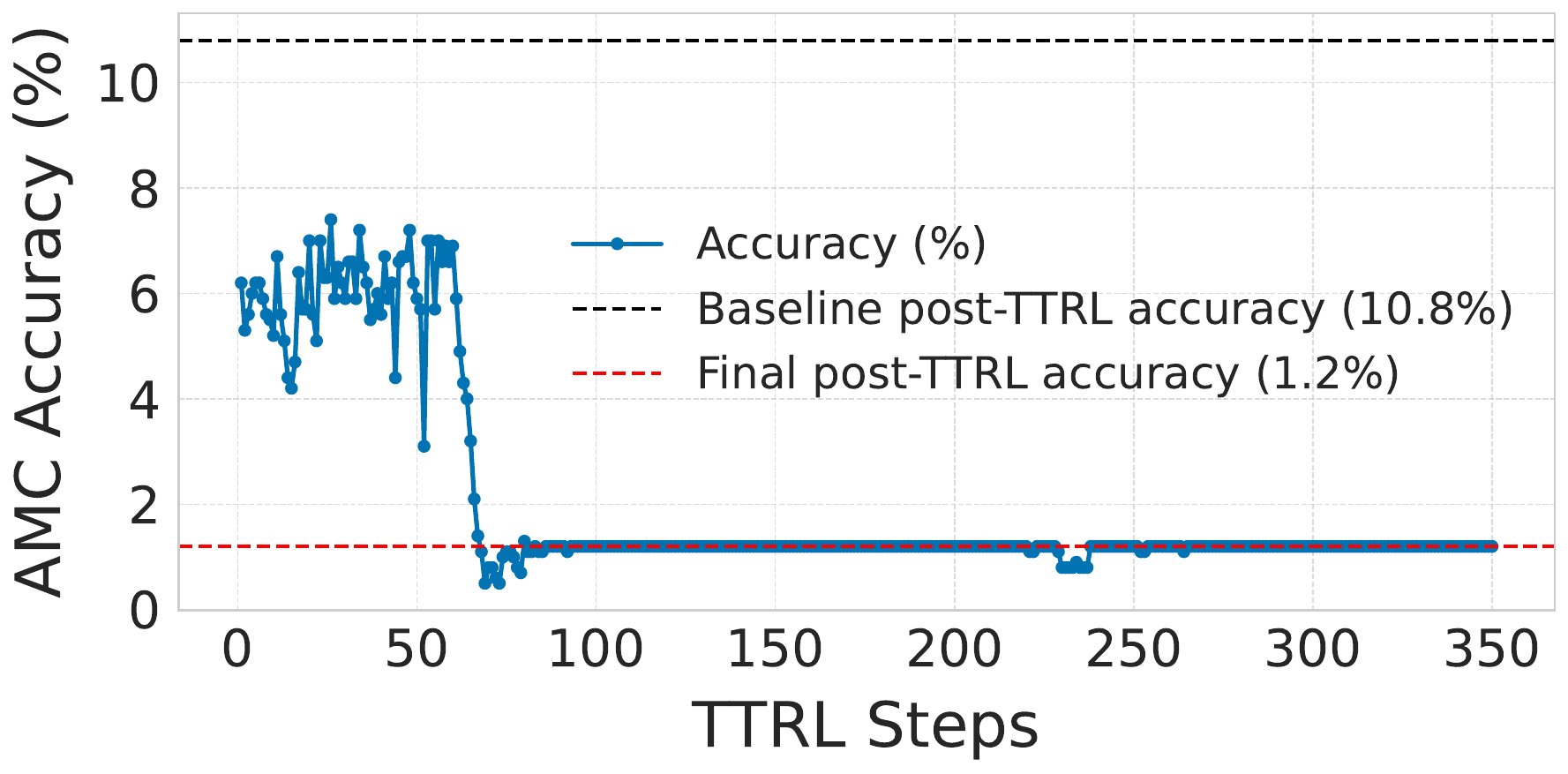}\label{subfig:amcwildjail_amc_llama1b}
    }
    \hfill
    \subfloat[]{%
        \includegraphics[width=0.3\textwidth]{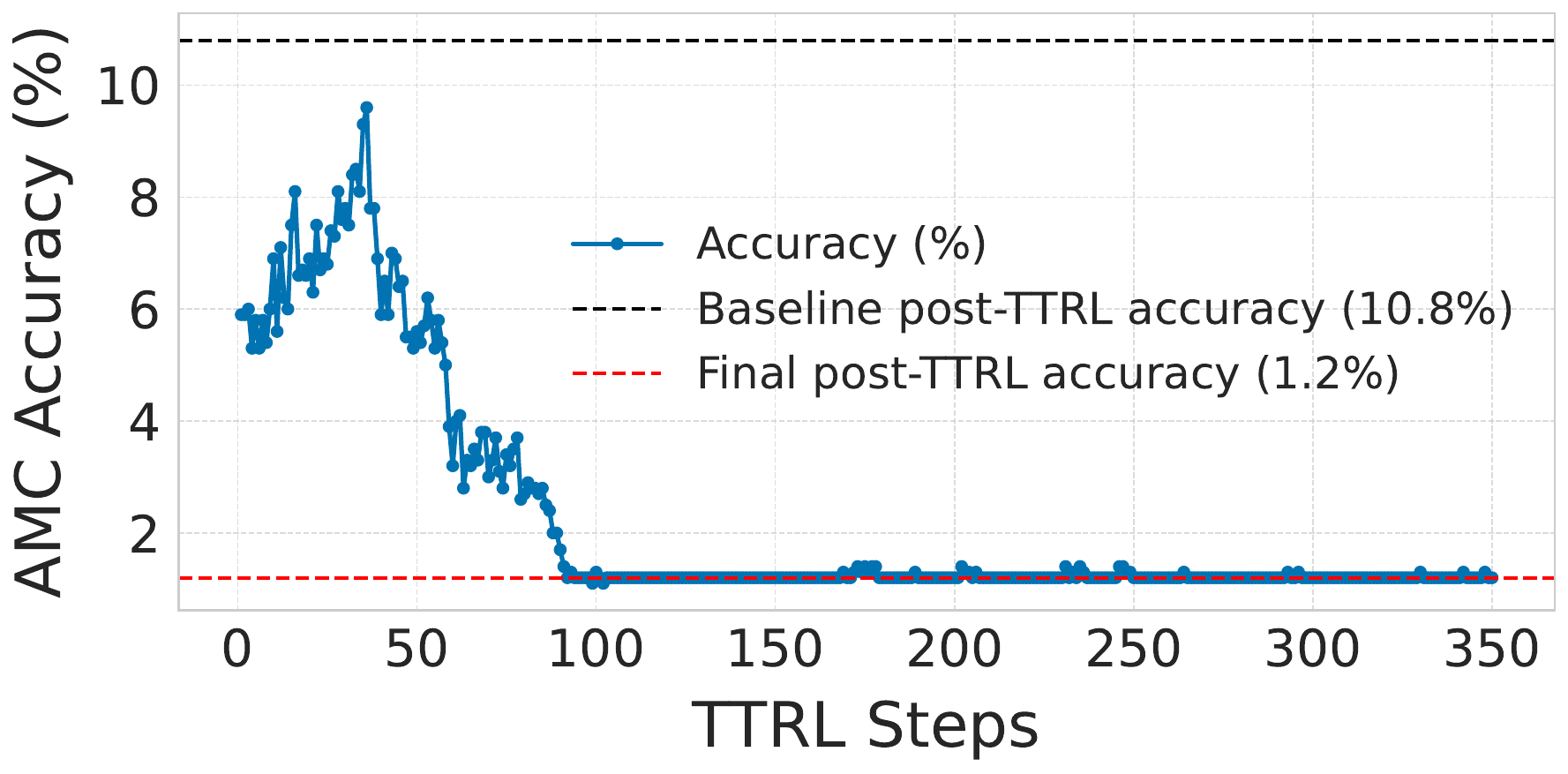}\label{subfig:amcllamaartifacts_amc_llama1b}
    }

    \caption{Impact on safety and reasoning for Llama-1B-Instruct model after harmful prompt injection across three jailbreak datasets, JailbreakV-28k, WildJailbreak, and Llama Artifacts (left to right, respectively) during TTRL, for safety (top row) and AMC accuracy (bottom row).}
    \label{fig:rq2_llama1b_3attacks}
\end{figure}

\begin{figure}[htbp]
    \centering
    % --- First row: Qwen results ---
    \subfloat[]{%
        \includegraphics[width=0.3\textwidth]{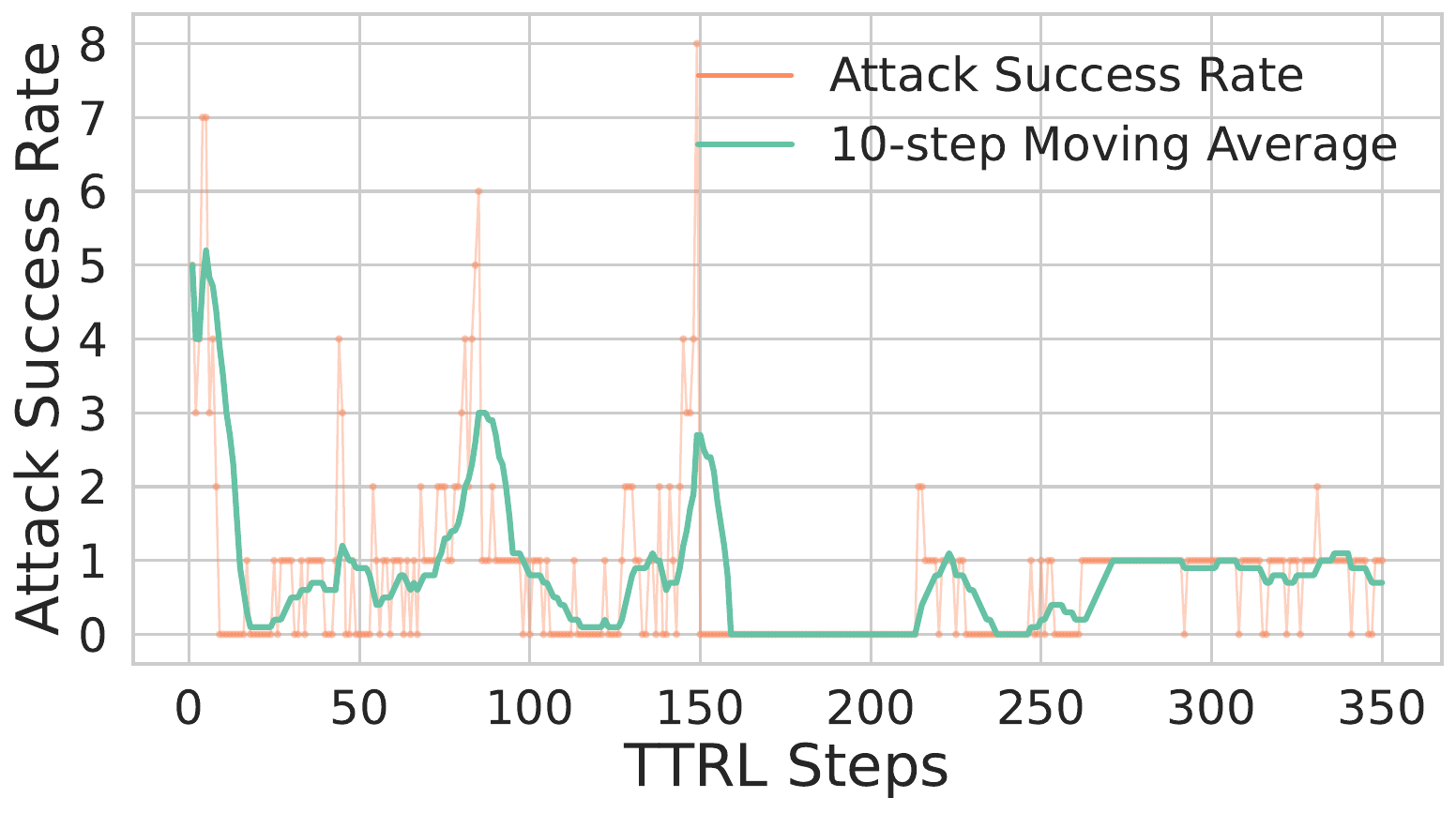}\label{subfig:amcjail_jailbreak2_llama3binstruct}
    }
    \hfill
    \subfloat[]{%
        \includegraphics[width=0.3\textwidth]{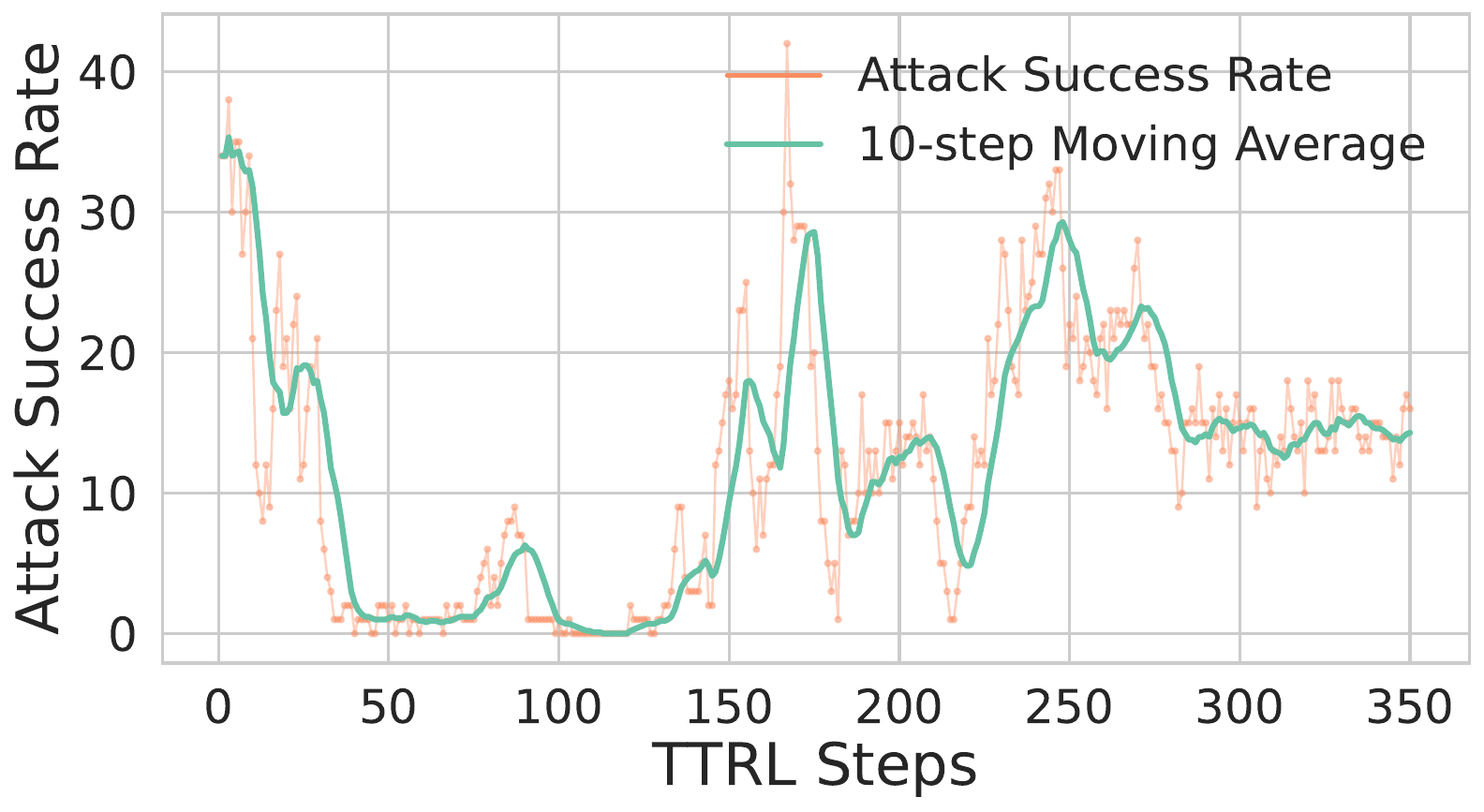}\label{subfig:amcwildjail_wildjail_llama3b}
    }
    \hfill
    \subfloat[]{%
        \includegraphics[width=0.3\textwidth]{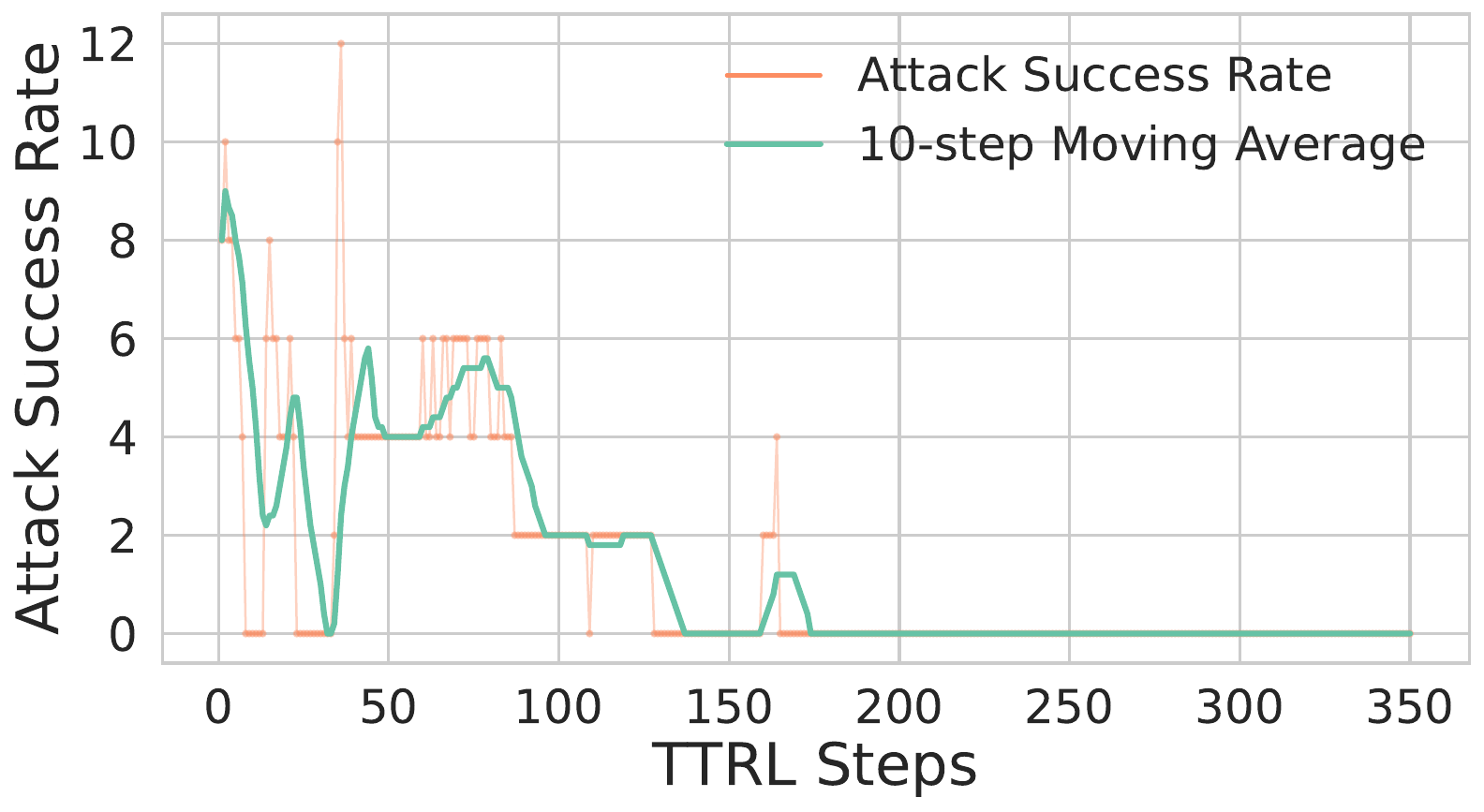}\label{subfig:amcllamaartifacts_llamaartifacts_llama3binstruct}
    }

    % --- Second row: Llama results ---
    
    \subfloat[]{%
        \includegraphics[width=0.3\textwidth]{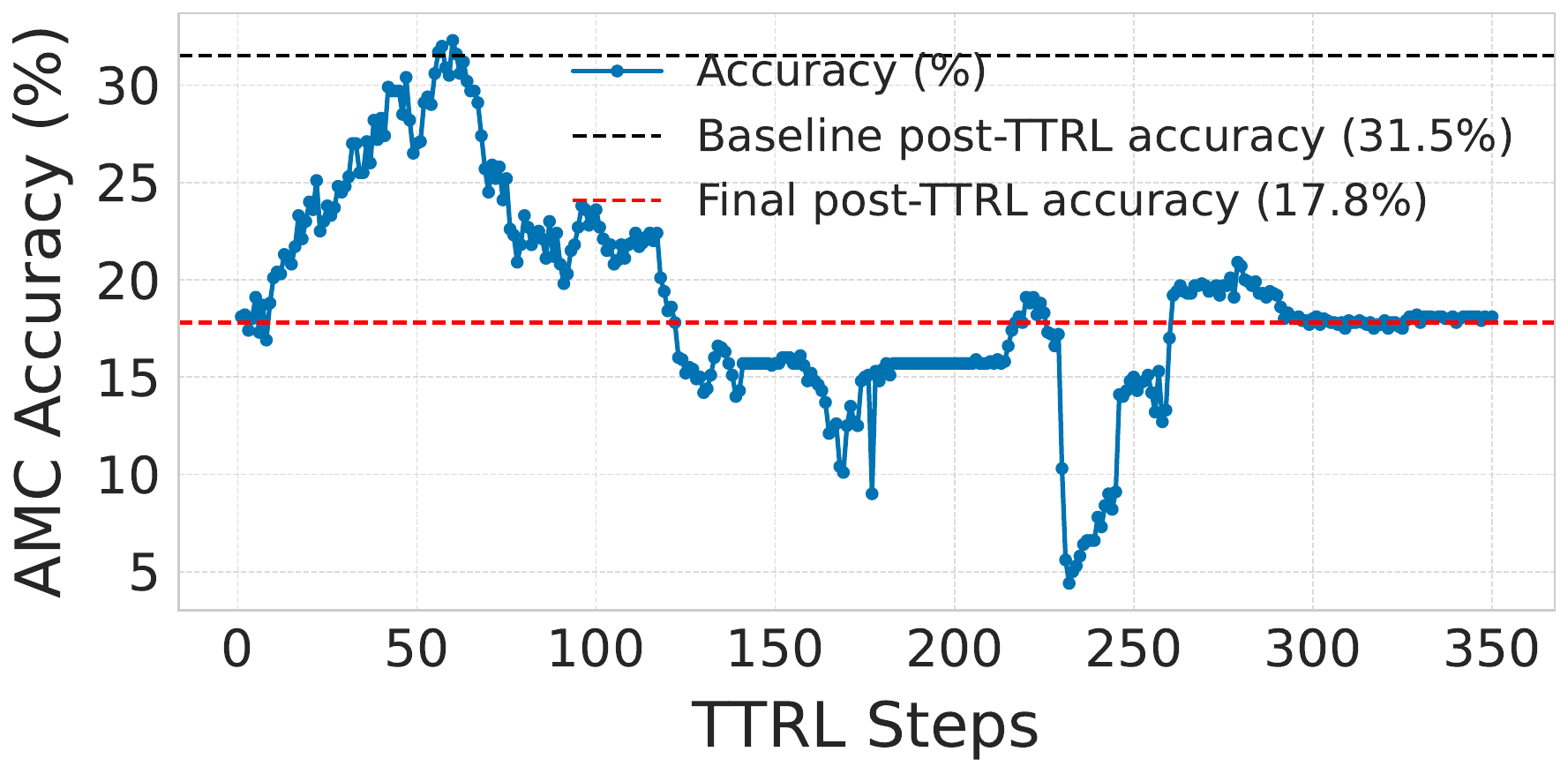}\label{subfig:amcjail_amc_llama3binstruct}
    }
    \hfill
    \subfloat[]{%
        \includegraphics[width=0.3\textwidth]{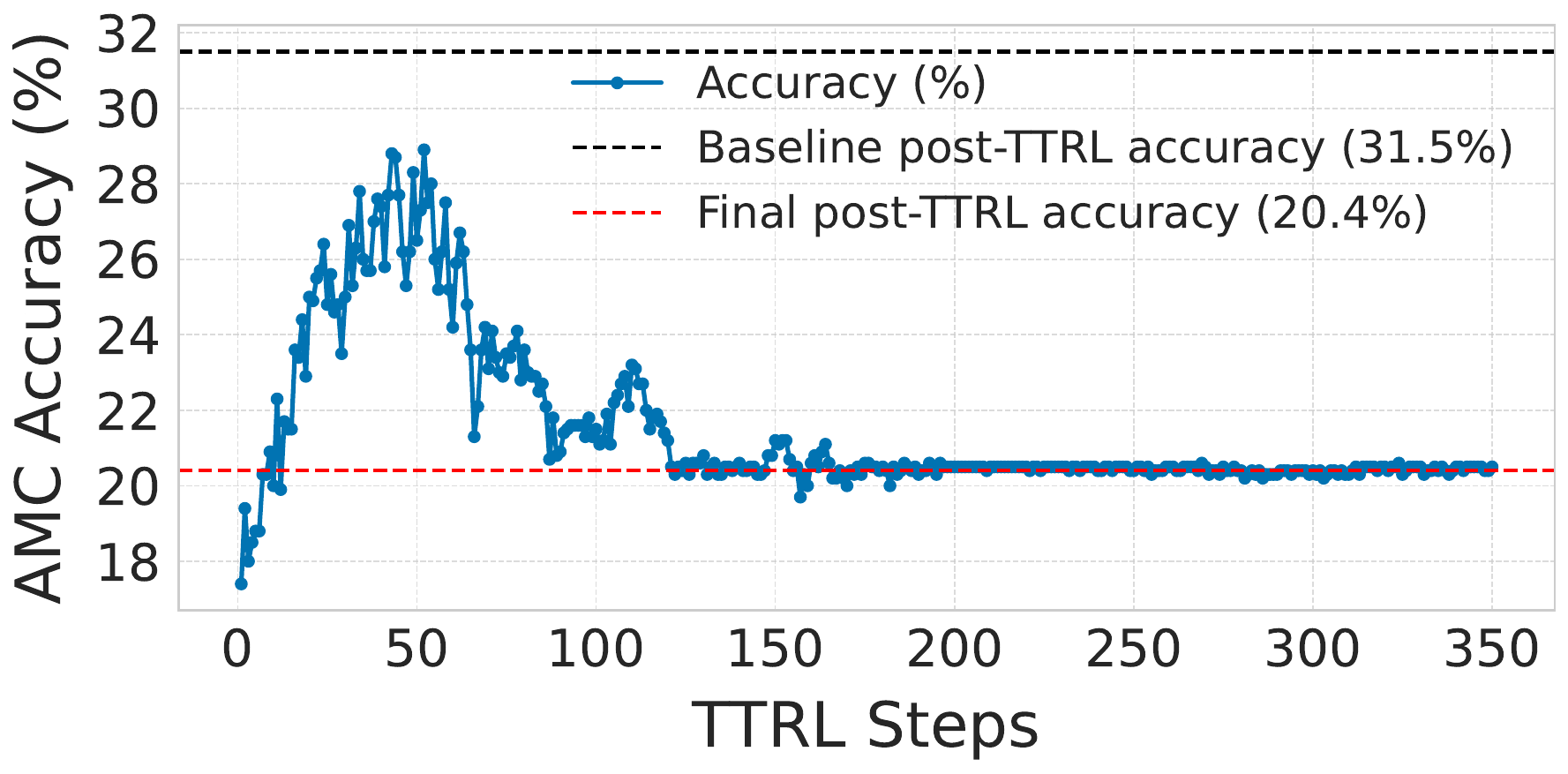}\label{subfig:amcwildjail_amc_llama3b}
    }
    \hfill
    \subfloat[]{%
        \includegraphics[width=0.3\textwidth]{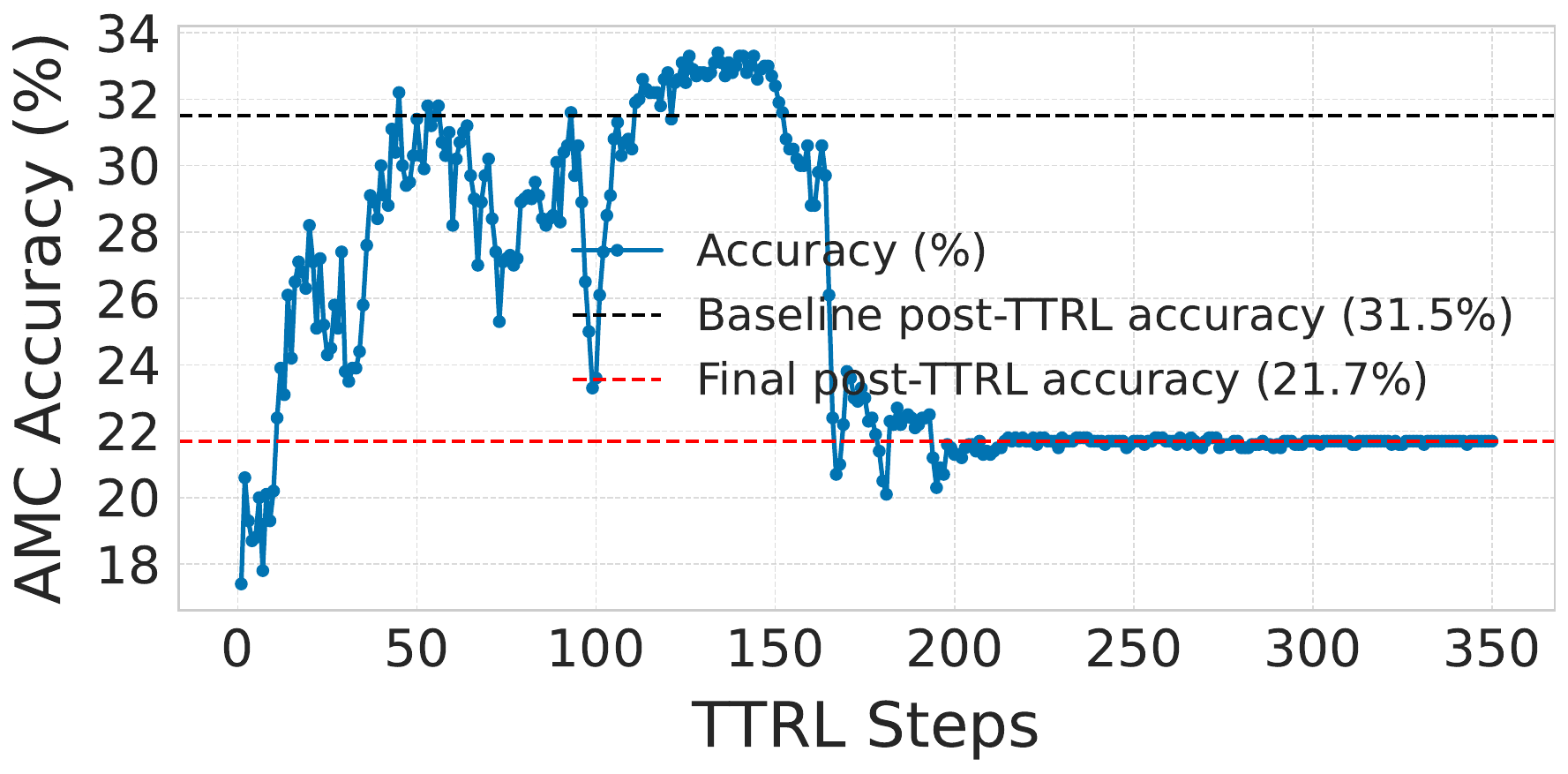}\label{subfig:amcllamaartifacts_amc_llama3b}
    }

    \caption{Impact on safety and reasoning for Llama-3B-Instruct model after harmful prompt injection across three jailbreak datasets, JailbreakV-28k, WildJailbreak, and Llama Artifacts (left to right, respectively) during TTRL, for safety (top row) and AMC accuracy (bottom row).}
    \label{fig:rq2_llama3b_3attacks}
\end{figure}

\begin{figure}[htbp]
    \centering
    % --- First row: Qwen results ---
    \subfloat[]{%
        \includegraphics[width=0.48\columnwidth]{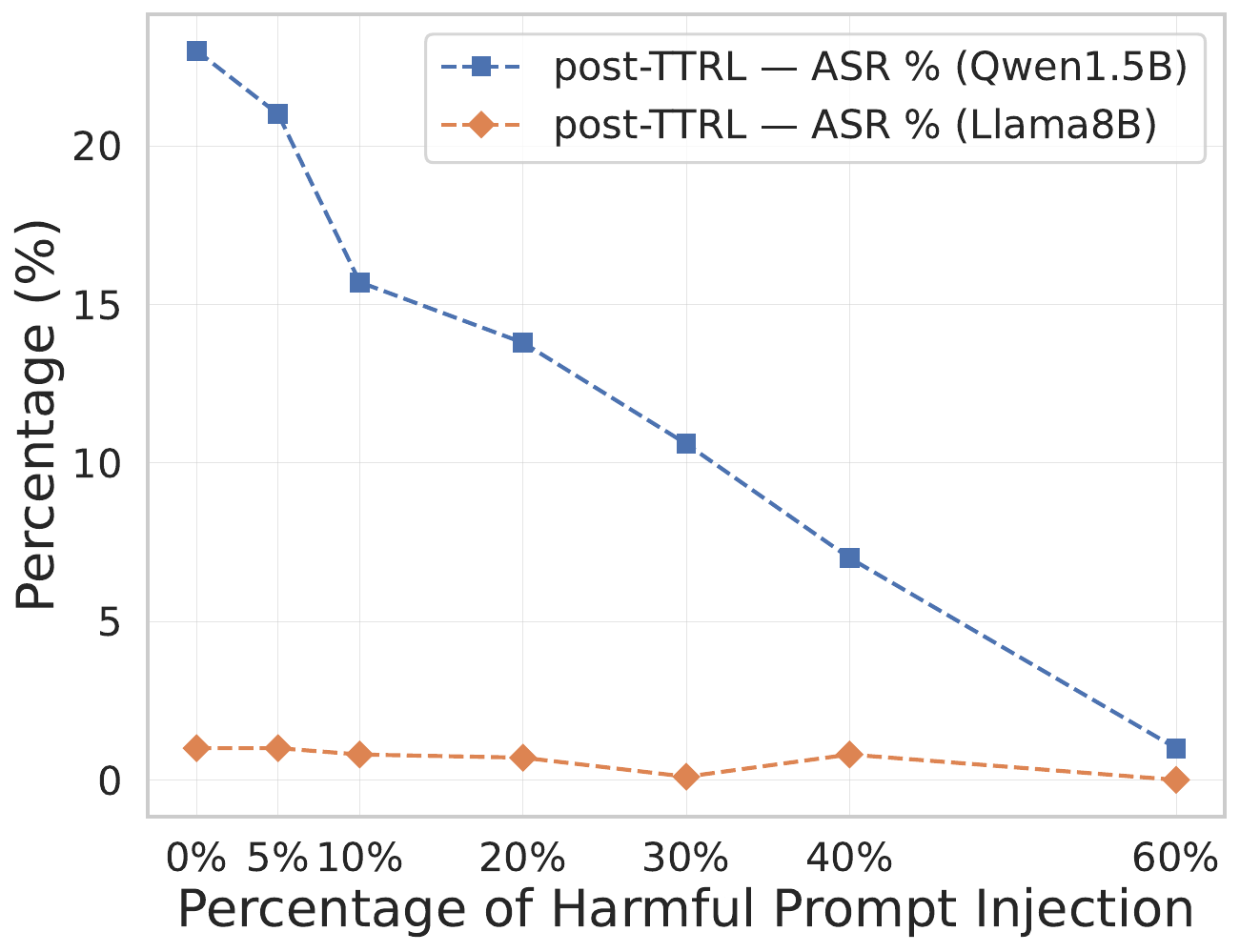}\label{subfig:safety_amplification_ASR}
    }
    \hfill
    \subfloat[]{%
        \includegraphics[width=0.48\columnwidth]{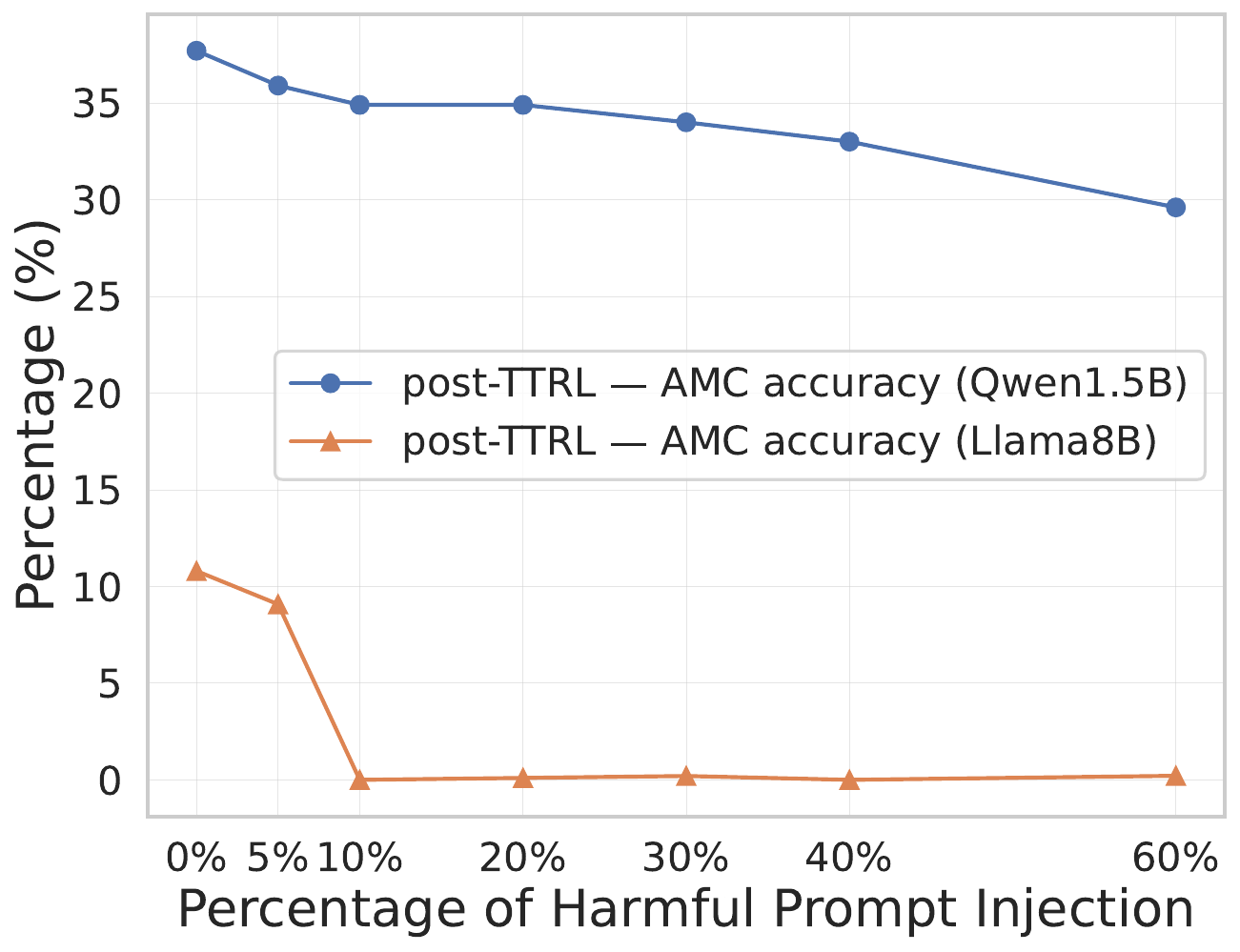}\label{subfig:safety_amplification_AMC}
    }

    % --- Second row: Harmfulness amplfication ---
    
    \subfloat[]{%
        \includegraphics[width=0.48\columnwidth]{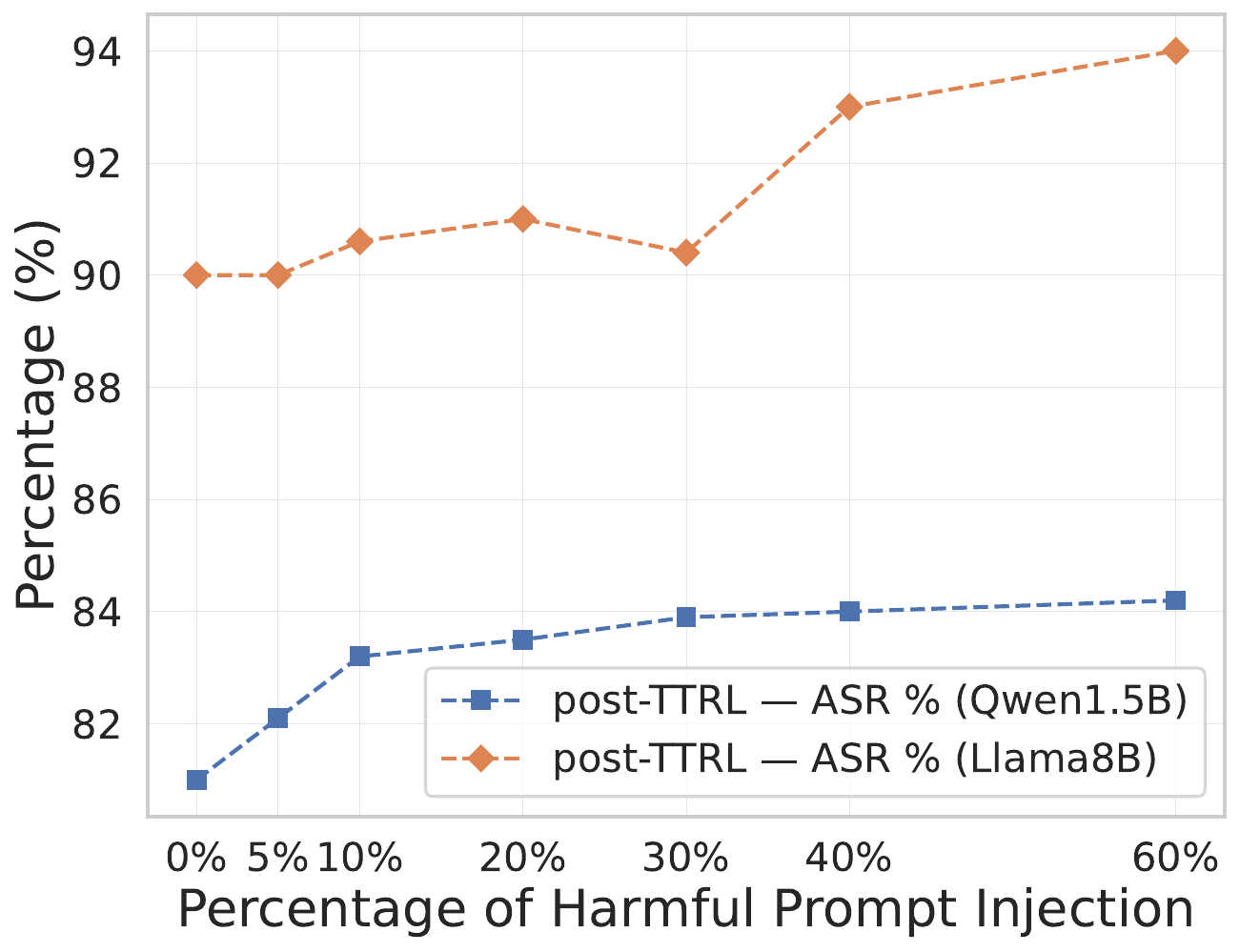}\label{subfig:harmful_amplification_ASR}
    }
    \hfill
    \subfloat[]{%
        \includegraphics[width=0.48\columnwidth]{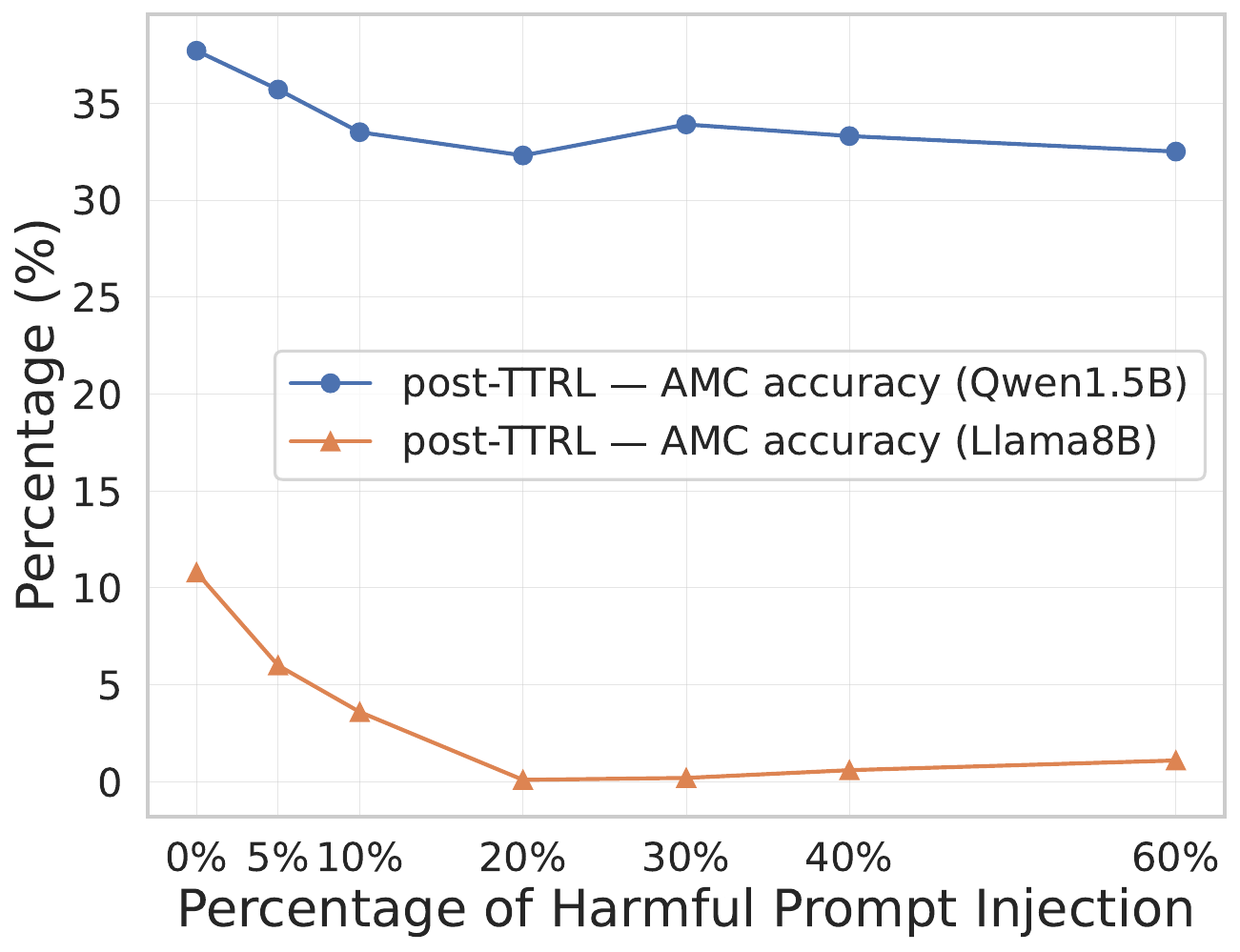}\label{subfig:harmful_amplification_AMC}
    }

    \caption{Safety and harmfulness amplification phenomenon after the harmful prompts injection, and the resulting impact on the AMC accuracy.}
    \label{fig:rq2_safe_harm_amplification}
\end{figure}

In this section, we provide the plots for ASR and reasoning during TTRL after the harmful prompt injections for all the instruction-tuned models presented in Figures \ref{fig:rq2_llama8b_3attacks} - \ref{fig:rq2_llama3b_3attacks}. We also report the magnitude of safety amplification and the resulting reasoning tax with respect to varying injection ratios in Figure \ref{fig:rq2_safe_harm_amplification}.

%%%%%%%%%%%%%%%%%%%%%%%%%%%%%%%%%%%%%%%%%%%%%%%%%%%%%%

%%%%%%%%%%%%%%%%%%%%%%%%%%%%%%%%%%%%%%%%%%%%%%%%%%%

%%%%%%%%%%%%%%%%%%%%%%%%%%%%%%%%%%%%%%%%%%%%%%%%%%%%%%%%

\subsection{Additional results for RQ3: benign prompt injections}
\label{subsec:rq3_additional_results}
\begin{figure}[htbp]
    \centering
    % --- First row: Qwen results ---
    \subfloat[]{%
        \includegraphics[width=0.45\textwidth]{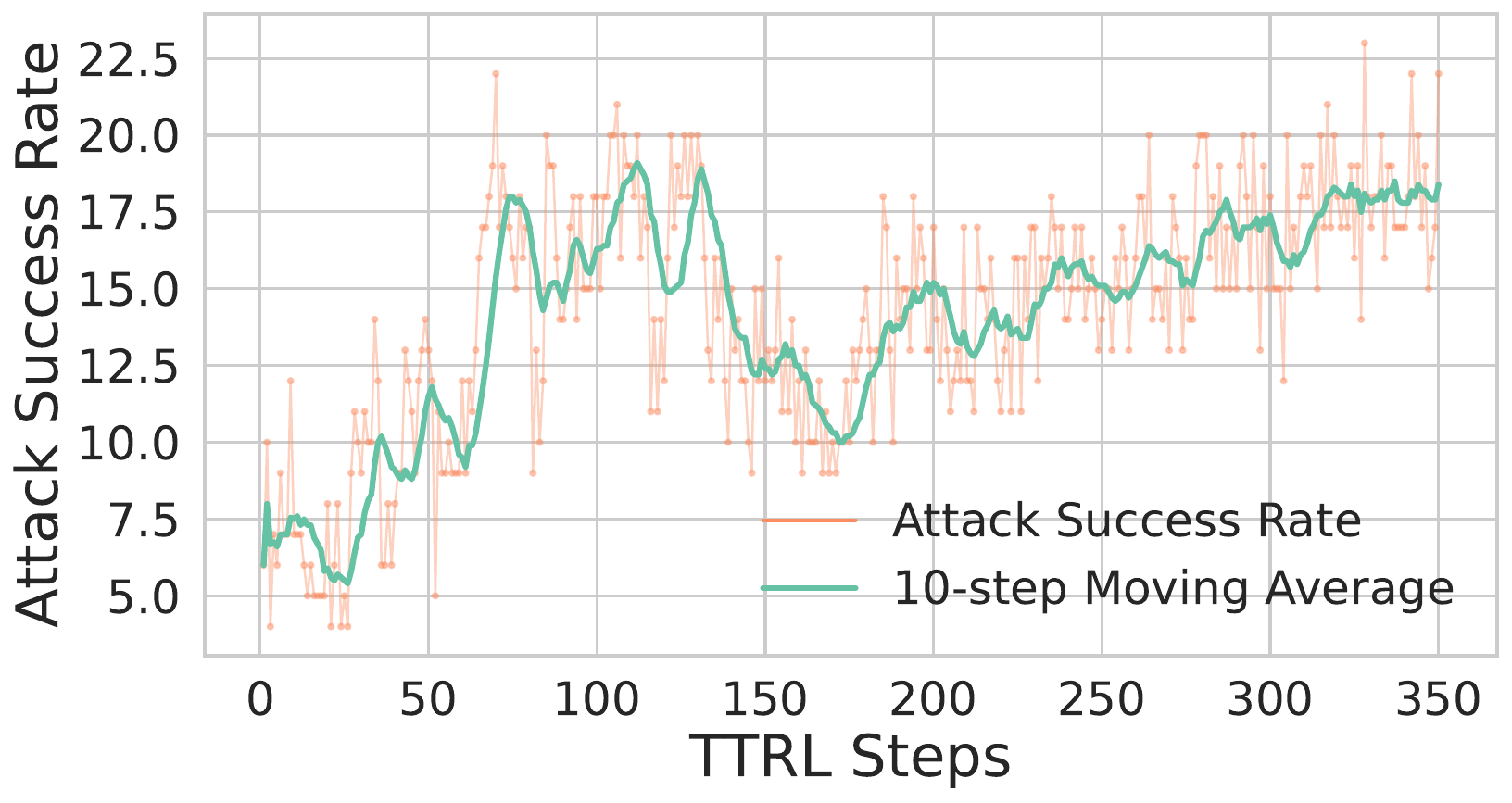}\label{subfig:amcinstruct_jailbreak_llama1binstruct}
    }
    \hfill
    \subfloat[]{%
        \includegraphics[width=0.45\textwidth]{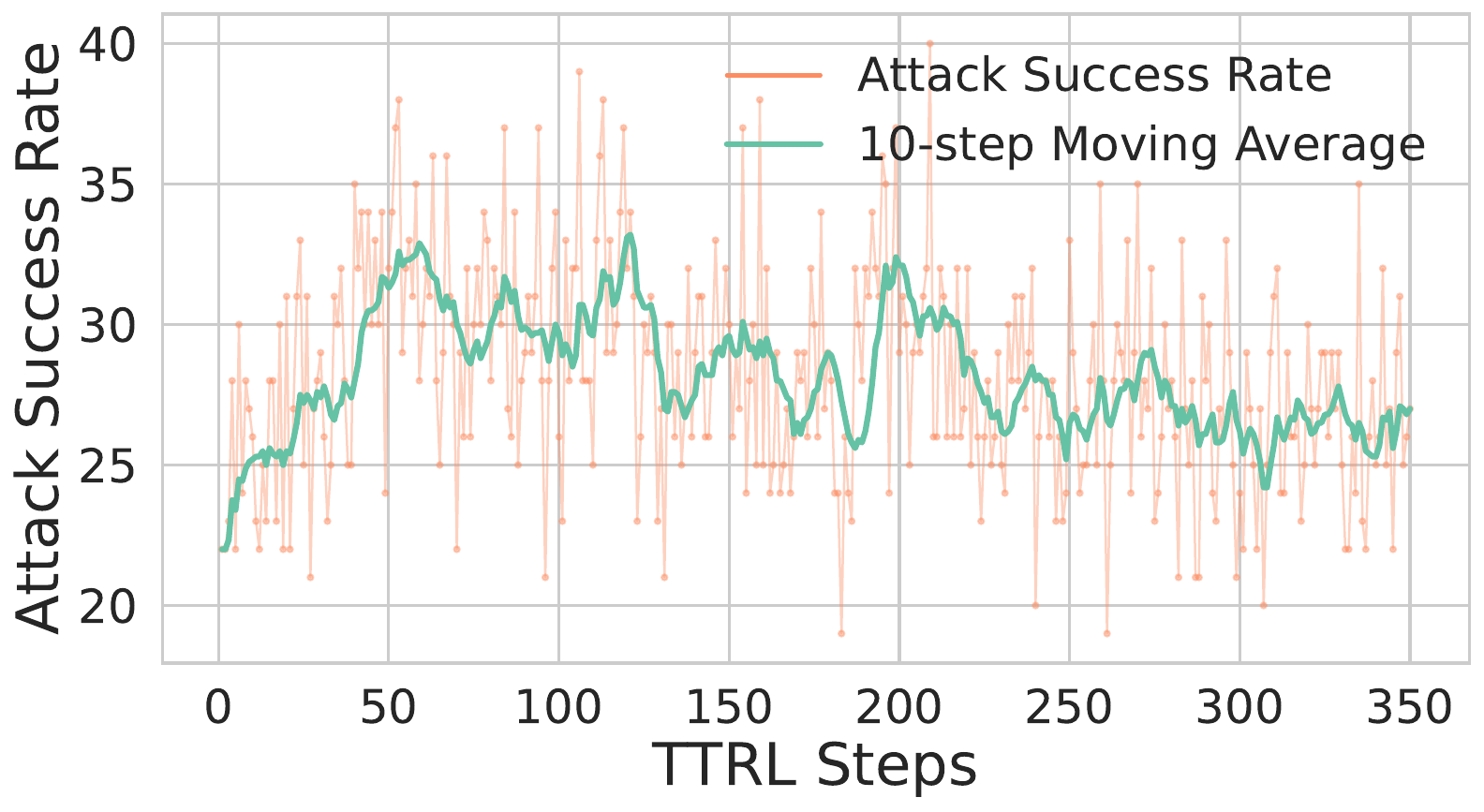}\label{subfig:amcinstruct_jailbreak_qwen1.5binstruct}
    }

    % --- Second row: Llama results ---
    
    \subfloat[]{%
        \includegraphics[width=0.45\textwidth]{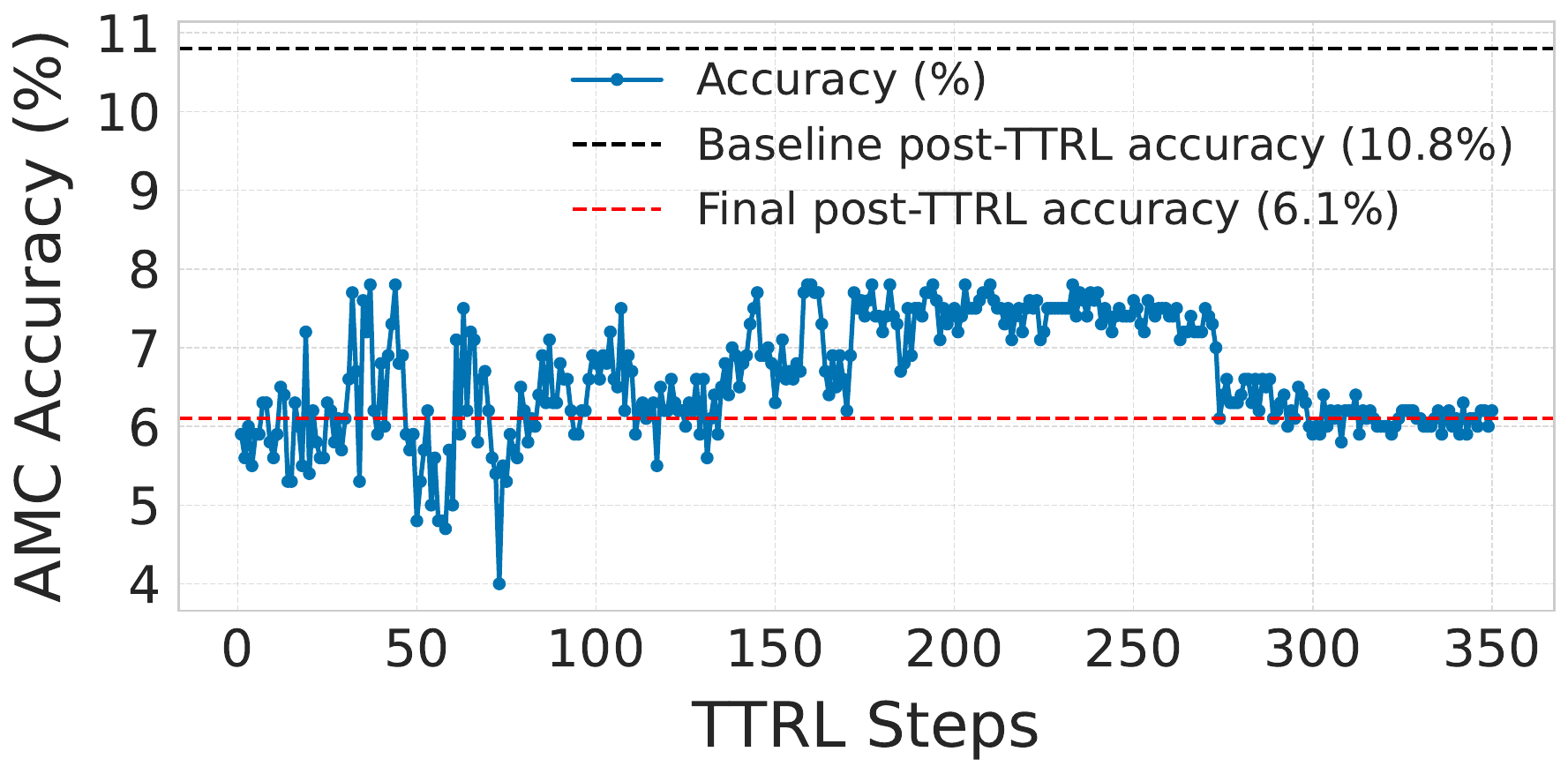}\label{subfig:amcinstruct_amc_llama1binstruct}
    }
    \hfill
    \subfloat[]{%
        \includegraphics[width=0.45\textwidth]{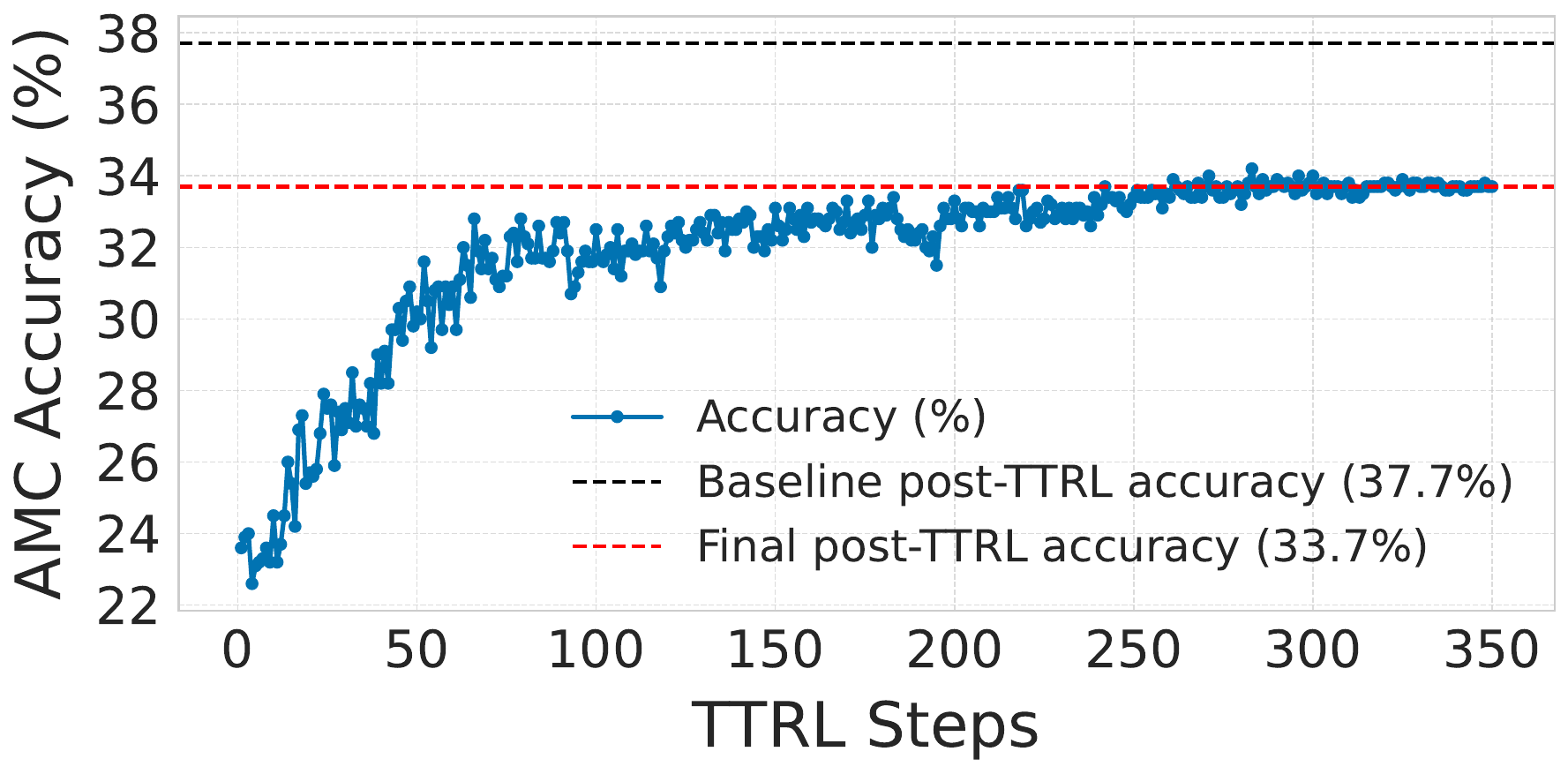}\label{subfig:amcinstruct_amc_qwen1.5binstruct}
    }
    \caption{Impact on safety (top row) and reasoning (bottom row) for Llama-1B-Instruct (left column) and Qwen-1.5B-Instruct (right column) after injecting benign instruction-following prompts. The ASR is reported on the held-out JailbreakV-28k prompts.}
    \label{fig:rq3_additional_plots}
\end{figure}

In this section, we provide the plots for ASR and reasoning during TTRL after the benign prompt injections from the UltraFeeback dataset for the rest of the models not presented in the main paper: Qwen-1.5B-Instruct and Llama-1B-Instruct. The plots are shown in Figure \ref{fig:rq3_additional_plots}.

\subsection{The impact of label extraction method}
\label{subsec:label_extraction}
\begin{figure*}[t]
    \centering
    % --- First row: Qwen results ---
    \subfloat[]{%
        \includegraphics[width=0.3\textwidth]{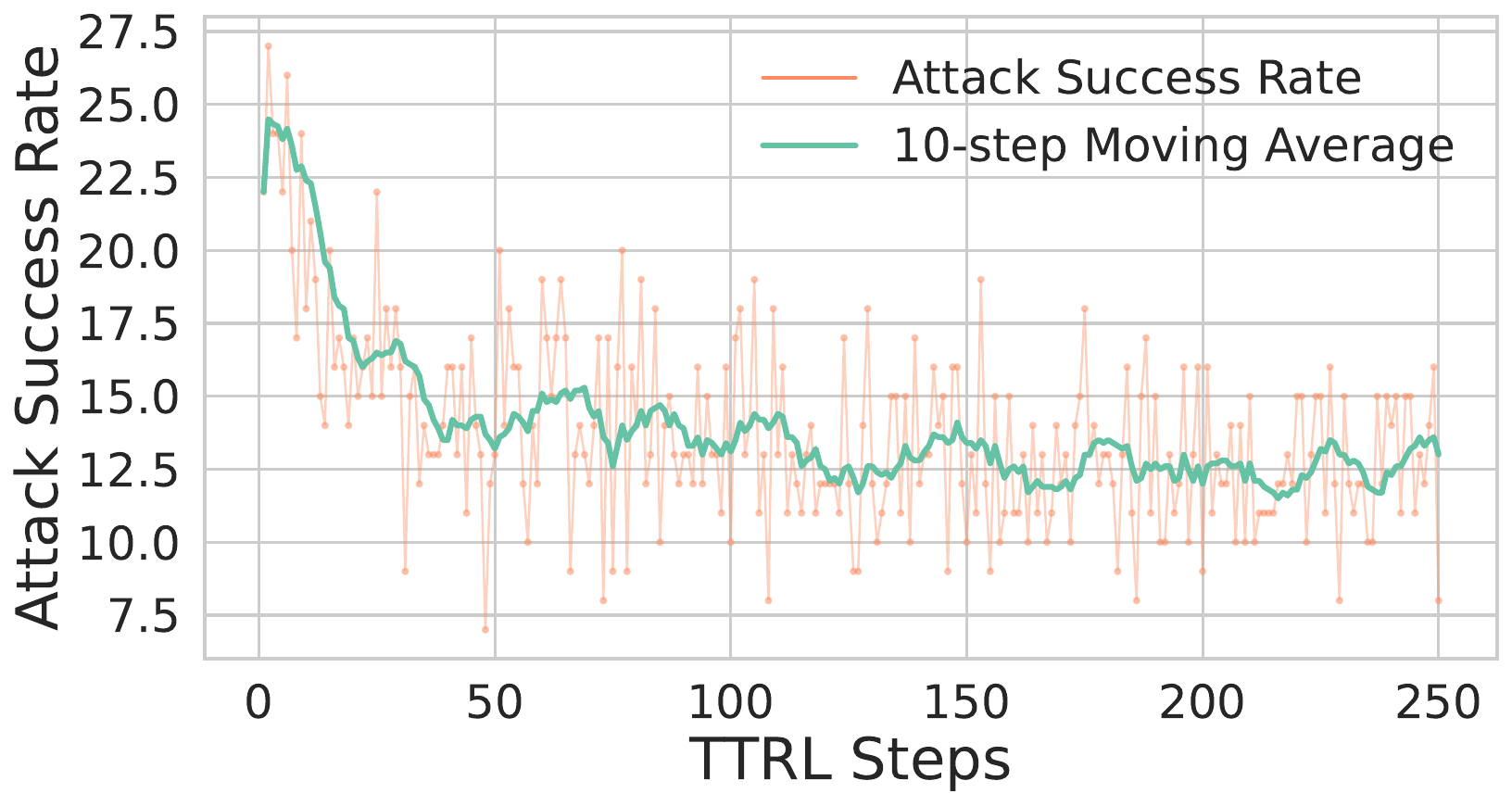}\label{subfig:amcjail_jailbreak2_qwen1.5binstruct_ratio_5percent_lastw}
    }
    \hfill
    \subfloat[]{%
        \includegraphics[width=0.3\textwidth]{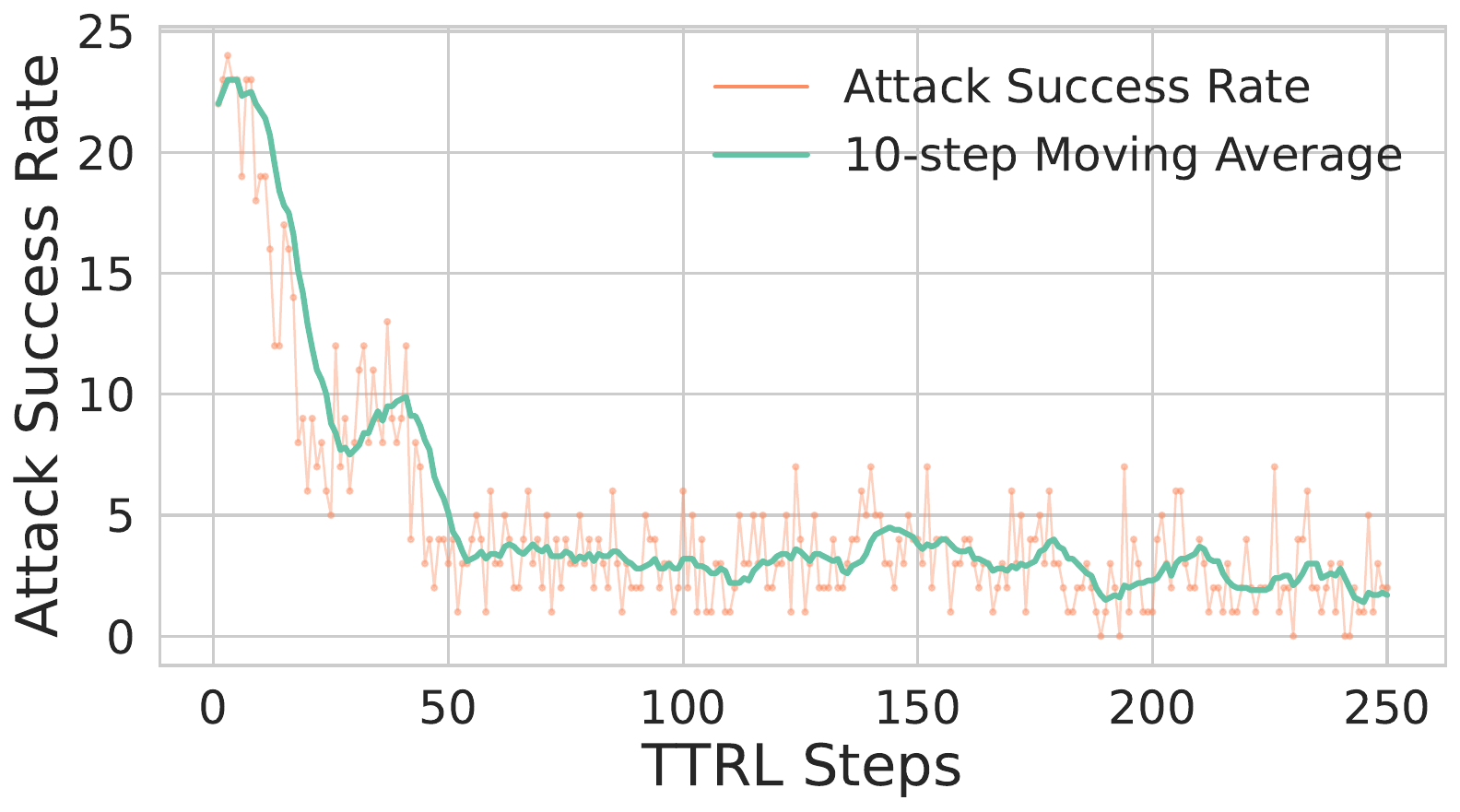}\label{subfigamcjail_jailbreak2_qwen1.5binstruct_ratio_10percent_lastw}
    }
    \hfill
    \subfloat[]{%
        \includegraphics[width=0.3\textwidth]{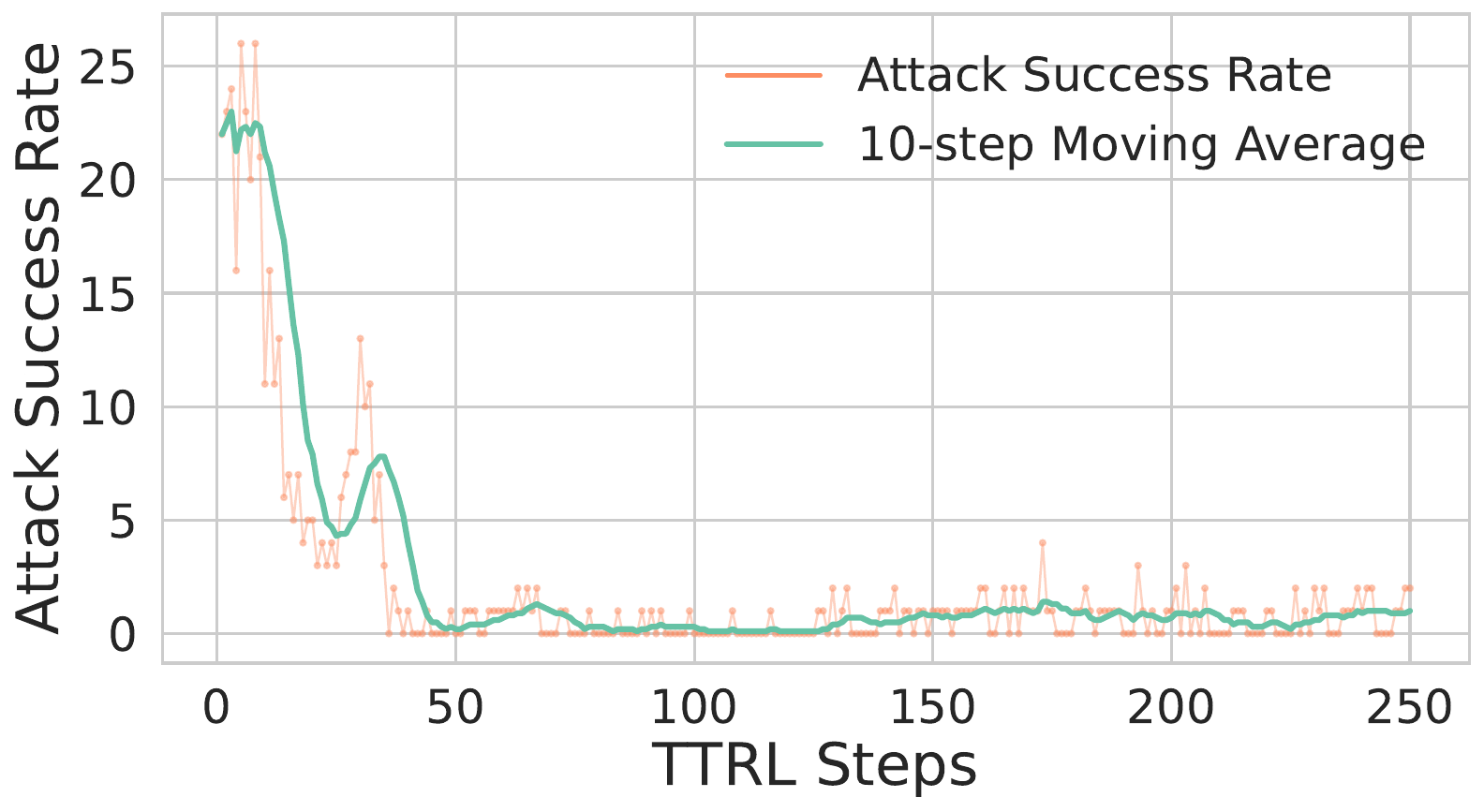}\label{subfig:amcjail_jailbreak2_qwen1.5binstruct_ratio_20percent_lastw}
    }

    % --- Second row: Llama results ---
    
    \subfloat[]{%
        \includegraphics[width=0.3\textwidth]{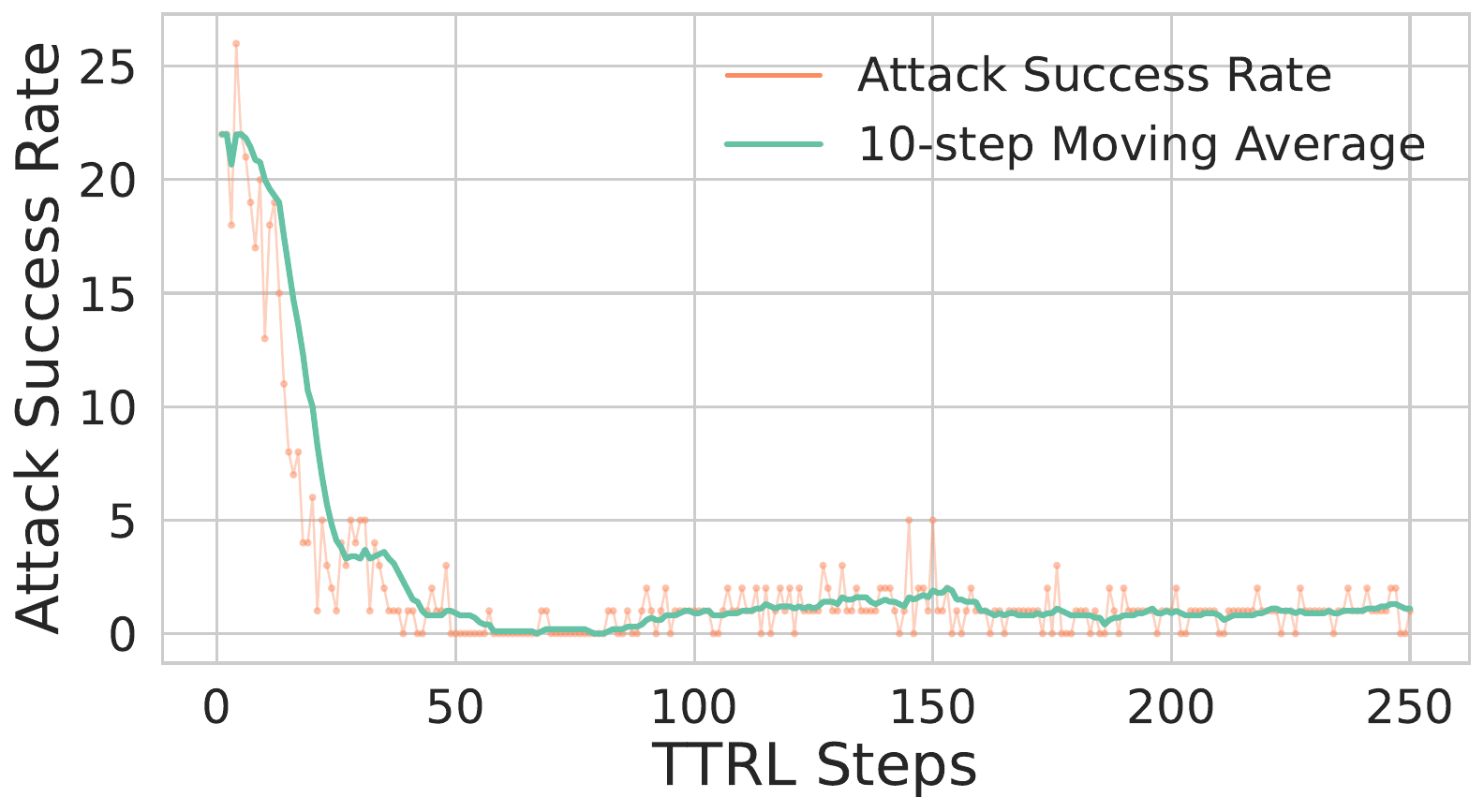}\label{subfig:amcjail_jailbreak2_qwen1.5binstruct_ratio_30percent_lastw}
    }
    \hfill
    \subfloat[]{%
        \includegraphics[width=0.3\textwidth]{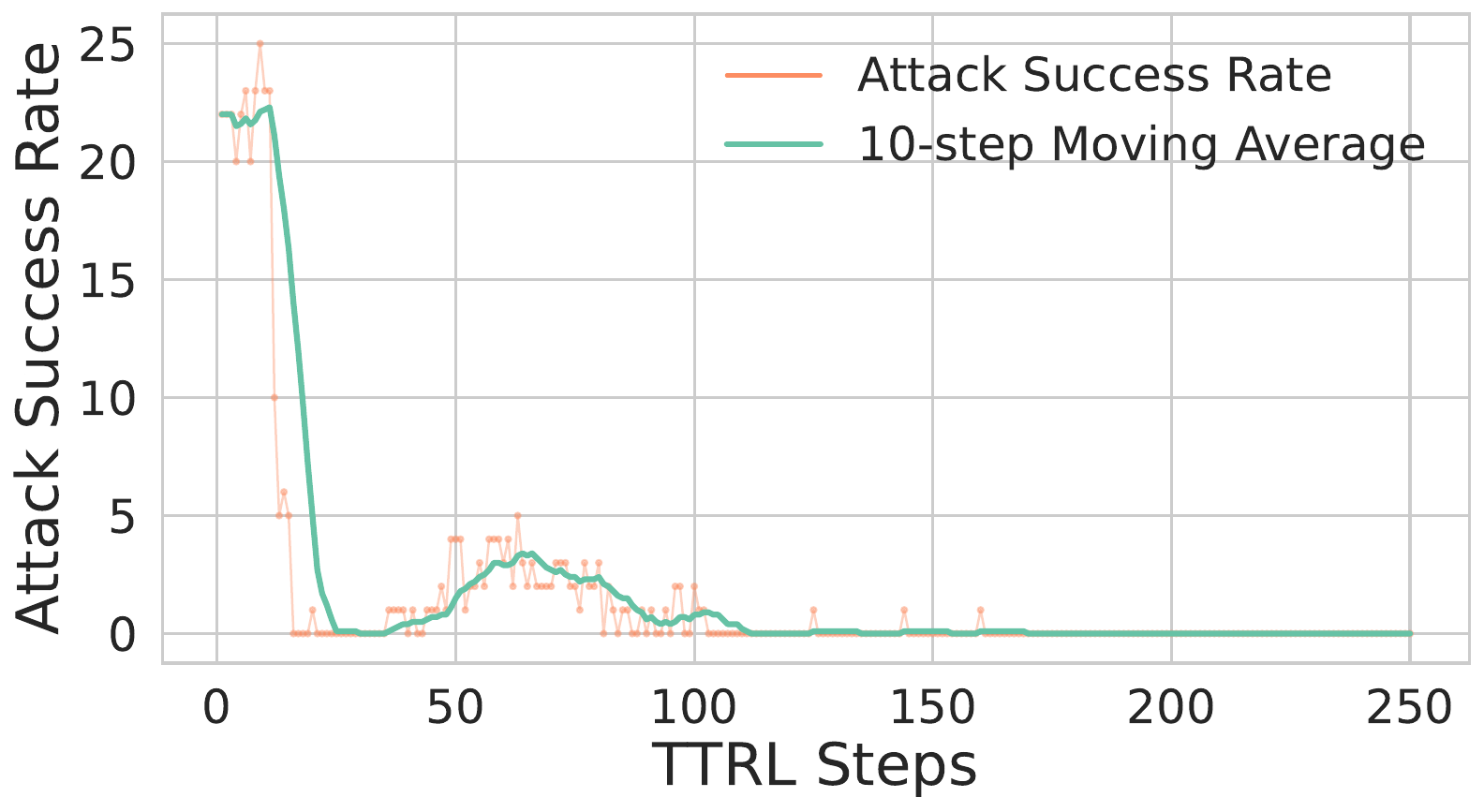}\label{subfig:amcjail_jailbreak2_qwen1.5binstruct_ratio_40percent_lastw}
    }
    \hfill
    \subfloat[]{%
        \includegraphics[width=0.3\textwidth]{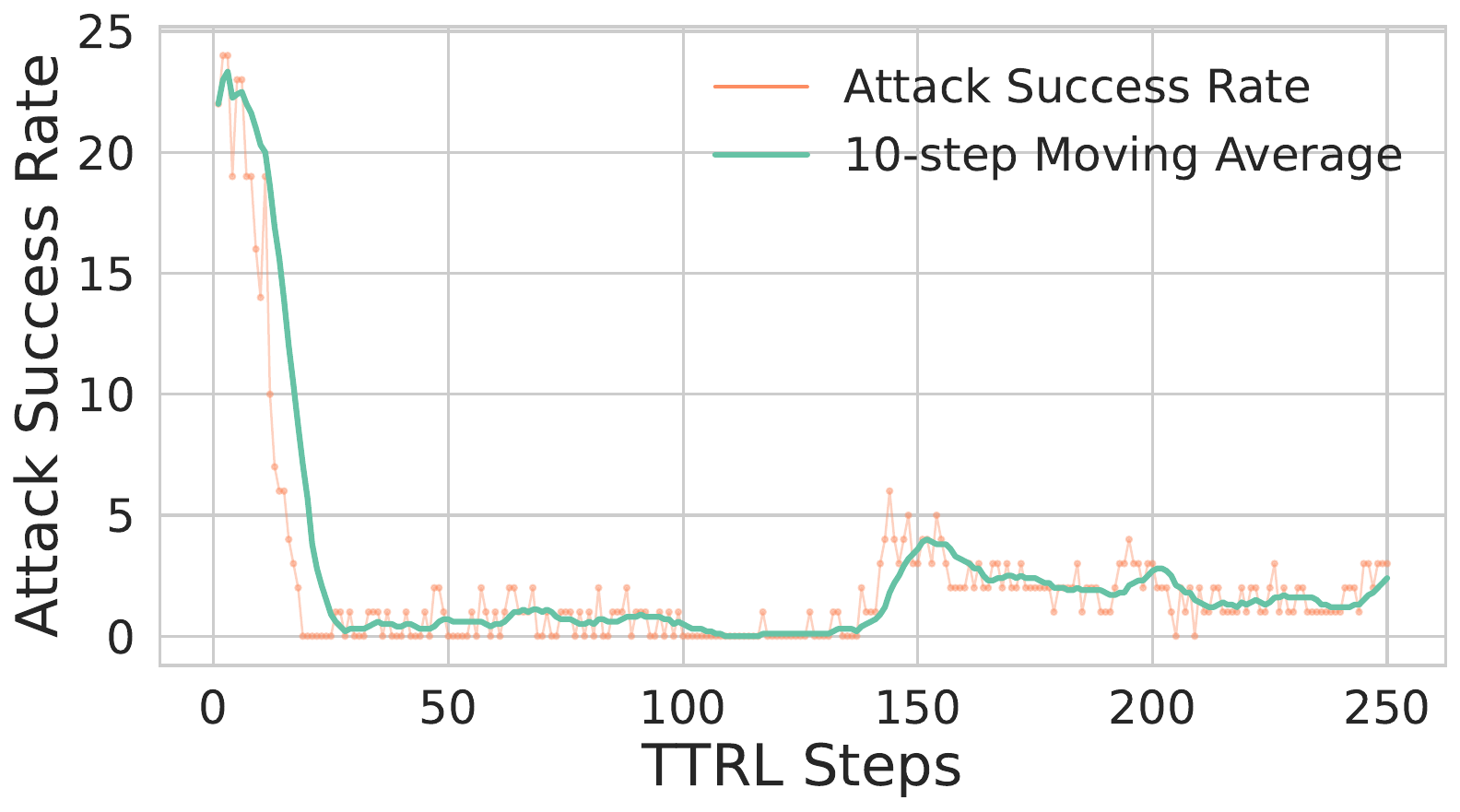}\label{subfigamcjail_jailbreak2_qwen1.5binstruct_ratio_60percent_lastw}
    }

    \caption{ASR for the Qwen1.5B-Instruct model with TTRL on the different injection ratios of AMC and JailbreakV-28k dataset, where the top row, from left to right is the injection ratio of 5\%, 10\%, 20\% respectively, and the bottomr, from left to right is the injection ratio of 30\%, 40\%, and 60\%, respectively.}
    \label{fig:empty_parser_exps}
\end{figure*}

As discussed in the main paper, there is a safety/harmfulness amplification in the TTRL setup, due to the extraction of the final token as the label. We also present the results here for a stronger safety amplification for the empty parser extraction. In this case, when the test-time training data is injected with harmful prompts, the label extraction method, instead of extracting the last token, extracts the empty parser (""). Therefore, during TTRL, if a numeric answer is not found, it simply extracts the empty parser "" as the label. Therefore, if the underlying sentence is arbitrary without a number in the senetence, it will just extract the empty parser. This eventually leads to a more pronounced safety amplification as seen in some of our preliminary experiments. We can see from Figure \ref{fig:empty_parser_exps}, that as the injection ratio of JailbreakV-28k dataset prompts increases in the test-time training data, the safety amplification on the held-out JailbreakV-28k dataset increases. However, this trend did not seem to be consistent over both models for all the jailbreak attacks. Further investigation of the impact of the label extraction in test-time training and the impact on the harmfulness is left for future work.

\subsection{Additional results for HarmInject prompts}

\begin{figure*}[htbp]
    \centering
    % --- First row: Qwen results ---
    \subfloat[]{%
        \includegraphics[width=0.45\textwidth]{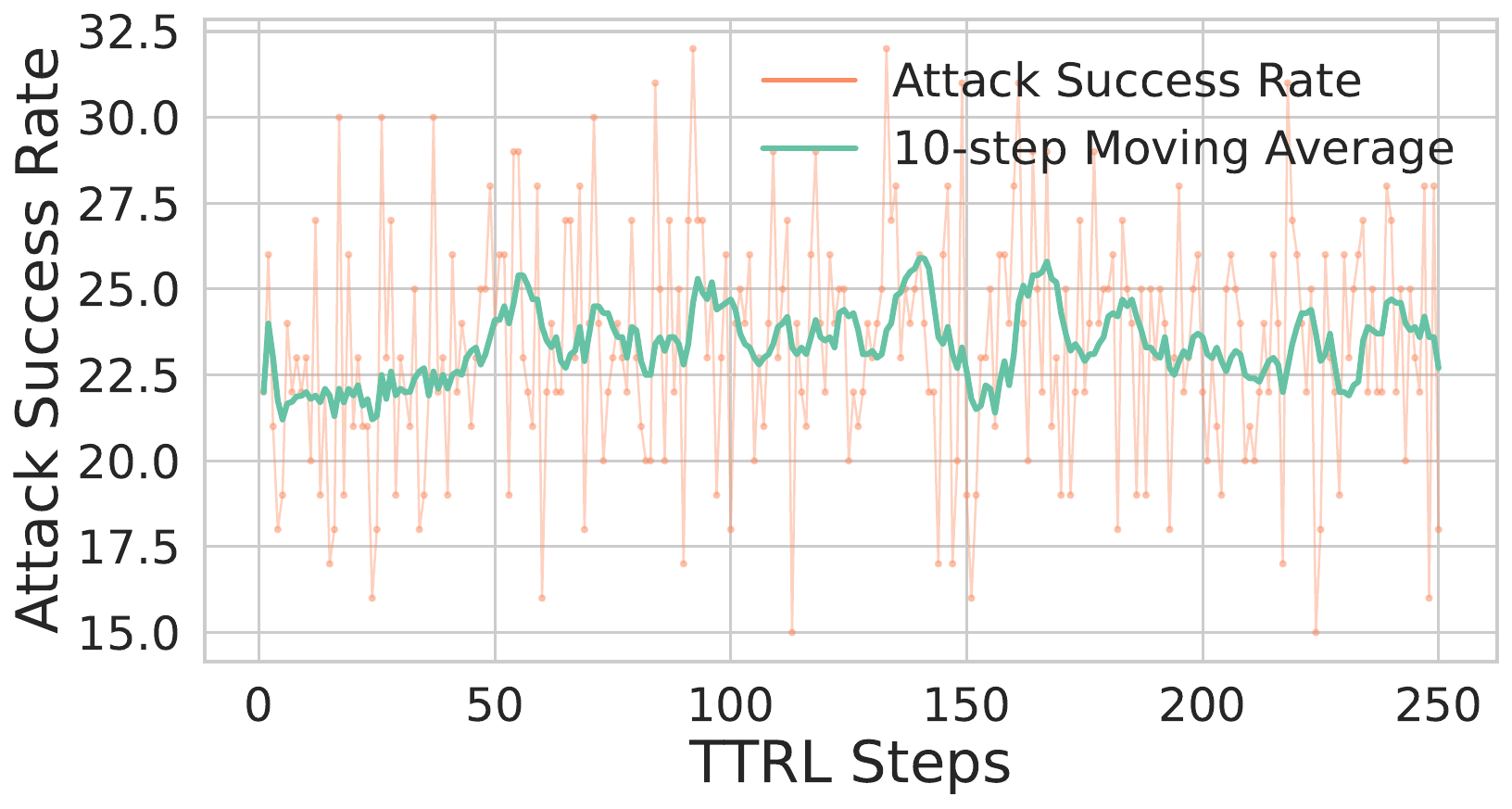}\label{subfig:amcnumleakllamaartifacts_jailbreak_qwen1.5bbinstruct}
    }
    \hfill
    \subfloat[]{%
        \includegraphics[width=0.45\textwidth]{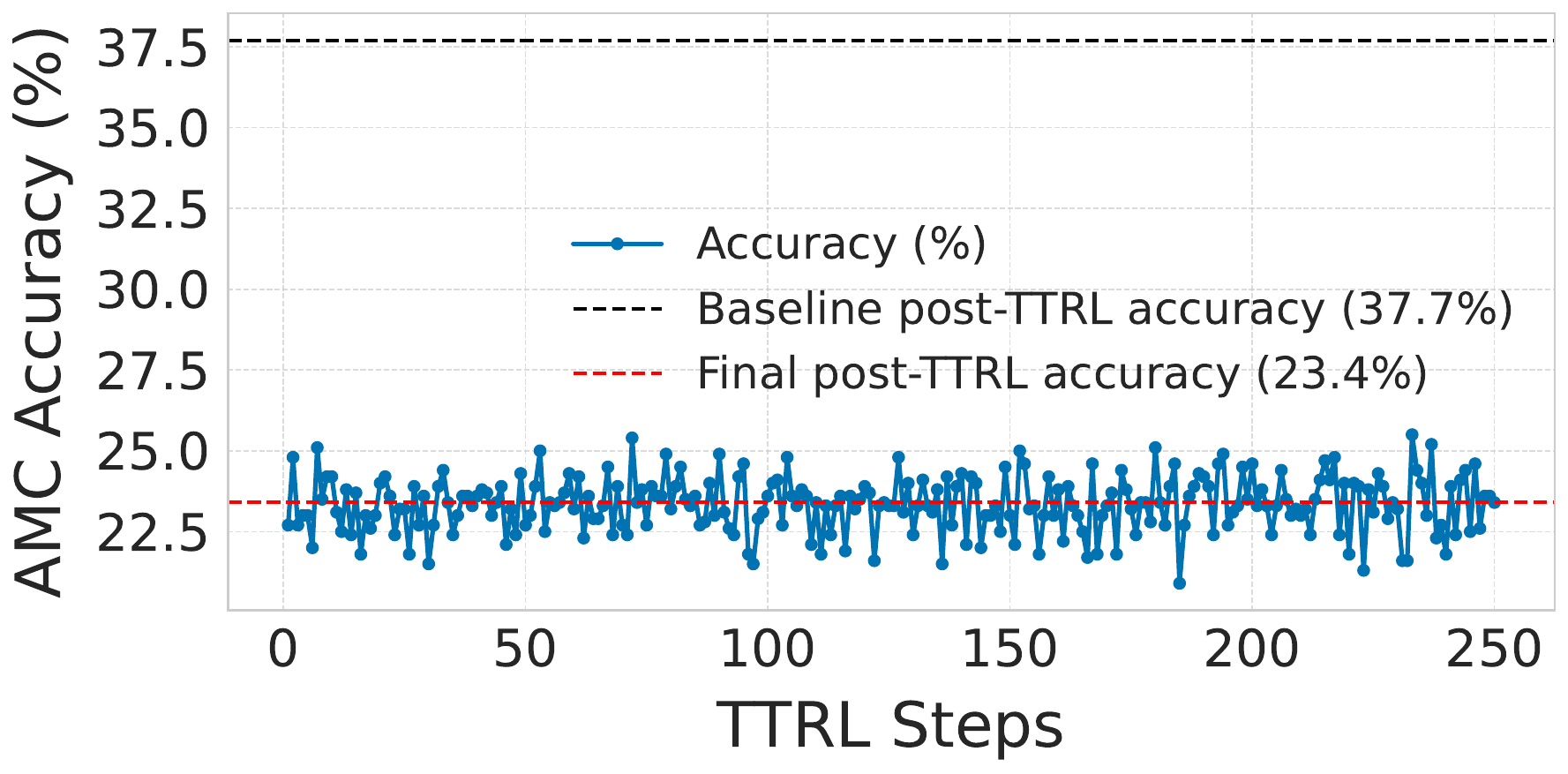}\label{subfig:amcnumleakllamaartifacts_amc_qwen1.5binstruct}
    }

    % --- Second row: Harmfulness amplfication ---
    
    \subfloat[]{%
        \includegraphics[width=0.45\textwidth]{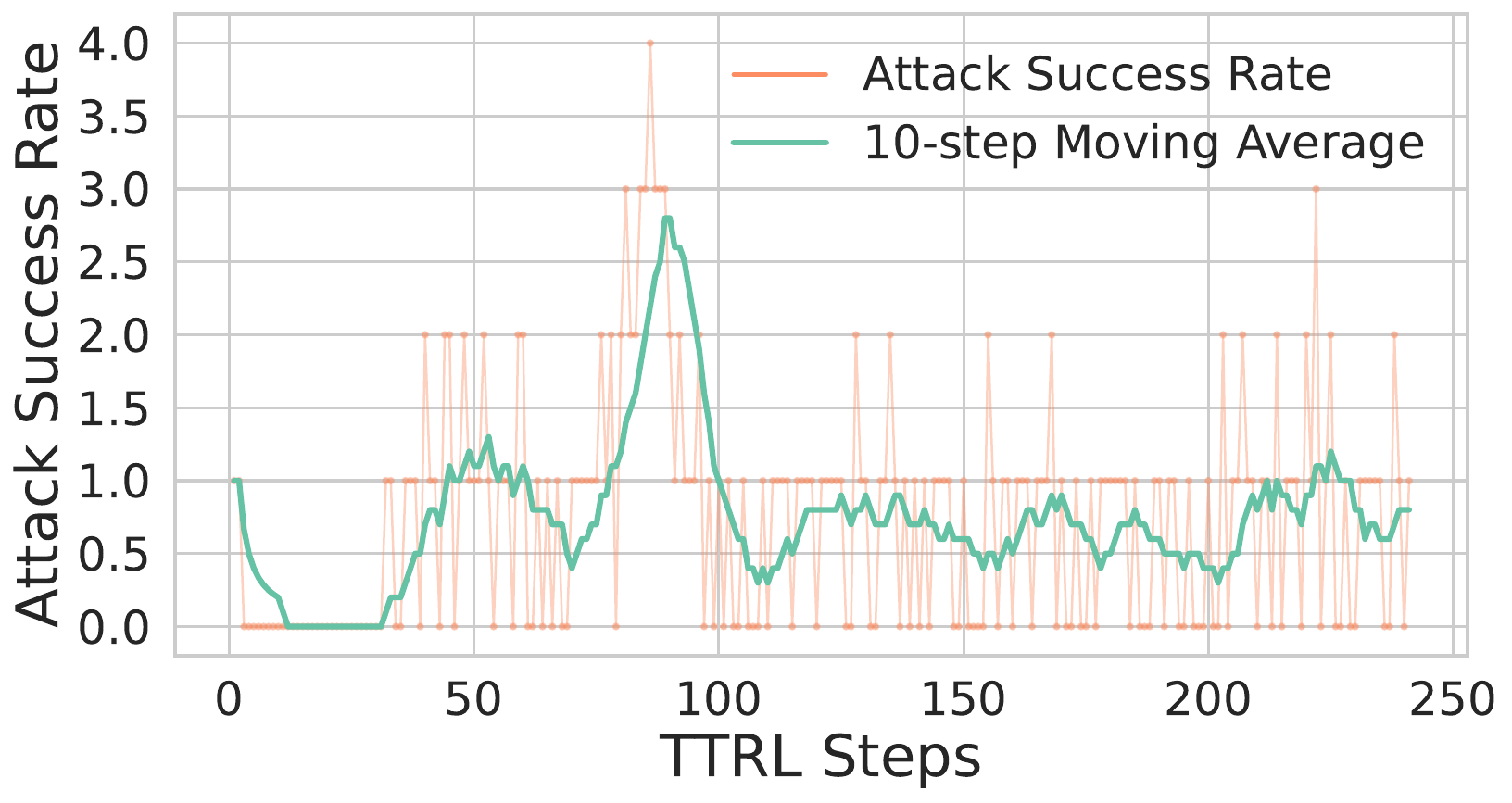}\label{subfig:amcnumleak_unclean_jailbreak_llama8binstruct}
    }
    \hfill
    \subfloat[]{%
        \includegraphics[width=0.45\textwidth]{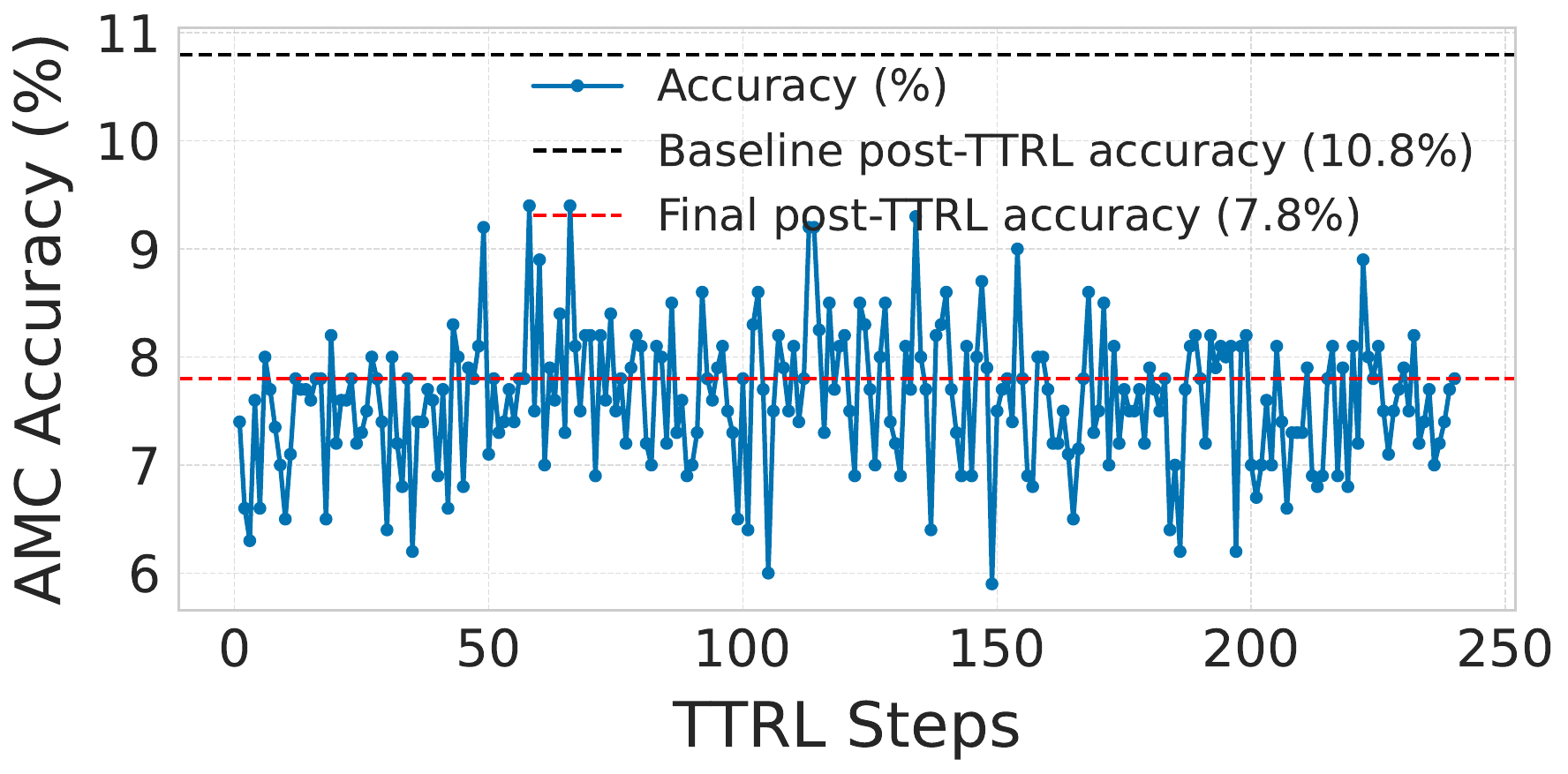}\label{subfig:amcnumleakjailbreak_amc_llama8binstruct}
    }
    \caption{Impact of HarmInject prompts during TTRL on the AMC dataset. (a) Attack success rate (ASR) for Qwen-1.5B-Instruct with HarmInject prompts constructed from Llama artifacts; evaluation on held-out JailbreakV-28k prompts. (b) AMC accuracy for Qwen-1.5B-Instruct after TTRL on HarmInject prompts constructed from Llama artifacts. (c) ASR for Llama-3-8B-Instruct with HarmInject prompts constructed from JailbreakV-28k prompts; evaluation on held-out JailbreakV-28k prompts. (d) AMC accuracy for Llama-3-8B-Instruct after TTRL on HarmInject prompts constructed from JailbreakV-28k prompts.}
    \label{fig:rq4_numleak_results2}
\end{figure*}

We also provide additional experimental results for both the Qwen and Llama models, utilizing HarmInject as test-time training data and prompts from the JailbreakV-28 dataset to construct the HarmInject prompts. The results are reported in Figure \ref{fig:rq4_numleak_results2}. We can see that for the Qwen model, trained on the HarmInject prompts constructed from the llama artifact prompts, although does not lead to harmfulness amplification. However. it stagnates the AMC reasoning accuracy as shown in Figures \ref{subfig:amcnumleakllamaartifacts_jailbreak_qwen1.5bbinstruct} and \ref{subfig:amcnumleakllamaartifacts_amc_qwen1.5binstruct}. For the Llama-3-8B-Instruct model, where the HarmInject prompts are constructed using the JailbreakV-28k dataset, shows harmfulness amplification until TTRL step 100 as shown in Figure \ref{subfig:amcnumleak_unclean_jailbreak_llama8binstruct} and then goes back to its default harmfulnless rate on the JailbreakV-28k dataset. Although, the harmfulness amplification effect is not very pronounced, the TTRL is not able to improve the AMC reasoning accuracy as seen in Figure \ref{subfig:amcnumleakjailbreak_amc_llama8binstruct}.

\end{document}